\documentclass[manuscript,screen]{acmart}

\setcopyright{none}
\renewcommand\footnotetextcopyrightpermission[1]{}

\makeatletter
\let\@authorsaddresses\@empty
\makeatother

\makeatletter
\def\@runningfoot{}
\def\@evenfoot{}
\def\@oddfoot{}
\makeatother
\usepackage{utfsym}
\usepackage[table]{xcolor}
\usepackage[edges]{forest}
\usepackage{multirow}
\usepackage{array} 
\usepackage{makecell} 
\usepackage{tikz}
\usetikzlibrary{shapes.geometric, arrows.meta, positioning, fit, calc, shadows, backgrounds}

\definecolor{lightblue}{RGB}{173, 216, 230}

\usepackage[most]{tcolorbox}
\AtBeginDocument{%
  }
\begin{document}

\title{Rethinking Memory in LLM based Agents: Representations, Operations, and Emerging Topics}

\author{YIMING DU}
\authornote{Both authors contributed equally to this research.}
\affiliation{%
  \institution{The Chinese University of Hong Kong}
  \city{Hong Kong}
  \country{China}}
\affiliation{%
  \institution{The University of Edinburgh}
  \city{Edinburgh}
  \country{UK}
}
\email{ydu@se.cuhk.edu.hk}

\author{WENYU HUANG}
\authornotemark[1]
\affiliation{%
  \institution{The University of Edinburgh}
  \city{Edinburgh}
  \country{UK}}
\email{w.huang@ed.ac.uk}

\author{DANNA ZHENG}
\authornotemark[1]
\affiliation{%
  \institution{The University of Edinburgh}
  \city{Edinburgh}
  \country{UK}}
\email{dzheng@ed.ac.uk}

\author{ZHAOWEI WANG}
\affiliation{%
  \institution{The Hong Kong University of Science and Technology}
  \city{Hong Kong}
  \country{China}
}
\email{zwanggy@cse.ust.hk}

\author{Sebastien Montella}
\affiliation{%
 \institution{Huawei Technologies Research \& Development (UK) Limited}
 \city{Edinburgh}
 \country{UK}}

\author{Mirella Lapata}
\affiliation{%
  \institution{The University of Edinburgh}
  \city{Edinburgh}
  \country{UK}}

\author{Kam-Fai Wong}
\affiliation{%
  \institution{The Chinese University of Hong Kong}
  \city{Hong Kong}
  \country{China}}
\email{kfwong@se.cuhk.edu.hk}

\author{Jeff Z. Pan}
\affiliation{%
  \institution{The University of Edinburgh}
  \city{Edinburgh}
  \country{UK}}
\affiliation{%
  \institution{Huawei Technologies Research \& Development (UK) Limited}
  \city{Edinburgh}
  \country{UK}
}
\email{j.z.pan@ed.ac.uk}

\renewcommand{\shortauthors}{Du et al.}


\begin{abstract}
\textbf{Abstract:} Memory is fundamental to large language model (LLM)-based agents, but existing surveys emphasize application-level use (e.g., personalized dialogue), while overlooking the atomic operations governing memory dynamics. This work categorizes memory into parametric (implicit in model weights) and contextual (explicit external data, structured/unstructured) forms, and defines six core operations: Consolidation, Updating, Indexing, Forgetting, Retrieval, and Condensation. Mapping these dimensions reveals four key research topics: long-term, long-context, parametric modification, and multi-source memory. The taxonomy provides a structured view of memory-related research, benchmarks, and tools, clarifying functional interactions in LLM-based agents and guiding future advancements. The datasets, papers, and tools are publicly available at \url{https://github.com/Elvin-Yiming-Du/Survey_Memory_in_AI}.
\end{abstract}

\maketitle

\section{Introduction}
\label{sec:introduction}
Memory is a core component of Large Language Model (LLM) based agents \citep{wang2024towards} and a critical step towards AGI \citep{park2023generative}, 
enabling persistent 
interactions \citep{maharana2024evaluating, zhong2024memorybank}, 
reasoning \citep{ge2025tremu}, 
multi-modal understanding \citep{long2025seeinglisteningrememberingreasoning}, 
personalization~\citep{li2024hello},
and 
multi-agent collaboration \citep{wang2025mirix}. While recent studies explore 
memory 
sources \citep{miao2024episodic, du2024perltqa}, operations \citep{yu2025memagent, cao2025memory, wang2025new, zhong2024memorybank}, and application~\citep{hong2023cogagent, mem0, long2025seeinglisteningrememberingreasoning, memobase2025}. A unified and systematic framework for organizing and evolving agent memory remains lacking.  

Existing surveys on agent memory adopt type-based and cognitive-inspired perspectives, offering valuable overviews but a limited unified 
but lacks operational formalization; most focus on subtopics, such as long-context modeling \citep{huang2023advancing}, long-term memory \citep{he2024human, jiang2024long}, personalization \citep{liu2025survey}, or knowledge editing \citep{wang2024knowledge}, without unifying core operations. \citet{zhang2024survey} covers only high-level operations such as writing, management, and reading, and misses some operations like indexing. More broadly, few surveys define the scope of memory research, analyze technical implementations, or provide practical foundations such as benchmarks and tools.

To address these gaps, we categorize memory into \textit{parametric} and \textit{contextual} types. Parametric memory encodes knowledge implicitly in model parameters \citep{wang2024wise}, while contextual memory stores explicit external information, either structured (e.g., graphs, tables, trajectories \citep{rasmussen2025zep}), or unstructured (e.g., text \citep{zhong2024memorybank}, vectors, audio, video \citep{long2025seeinglisteningrememberingreasoning}). Temporally, memory spans both long-term (e.g., multi-turn dialogue, external observations \citep{li2024hello}) and short-term contexts (e.g., kv-cache, current dialogue history \citep{packer2023memgpt}). Based on these types, we define six memory operations, which can be further classified into three categories: \textit{Encoding}, \textit{Evolving}, and \textit{Adapting}. Memory encoding encompasses consolidation (integrating new knowledge into persistent memories \citep{feng2024tasl}) and indexing (organizing memory for retrieval \citep{wu2024longmemeval}). Memory evolving includes updating (modifying existing memory to incorporate recent updates~\citep{chen2024compress}) and forgetting (removing outdated or incorrect content \citep{tian2024forget}). Memory Adapting covers retrieval (accessing relevant memory \citep{gutierrez2024hipporag}) and condensation (reducing size while preserving key information \citep{chen2024compress}).

Beyond this structural taxonomy, functional perspectives of memory provide a complementary lens for understanding LLM systems. Episodic memory, rooted in cognitive psychology \citep{tulving1972episodic}, stores temporally anchored experiences—such as dialogue histories and event sequences—and supports reasoning and adapting in dynamic environments \citep{fountas2024human, miao2024episodic}. Semantic memory encodes structured and generalizable knowledge, often formalized as queryable knowledge graphs or tables, complementing parametric memory to enhance reasoning and retrieval-augmented generation (RAG). Procedural memory captures task execution patterns and learned trajectories, typically formed through large-scale training or reinforcement learning with chain-of-thought data, and drives efficient tool use and problem-solving in task-oriented agents. Working memory acts as a dynamic control mechanism that integrates short-term caches and activated long-term knowledge, enabling real-time reasoning, planning, and decision-making. These functional types highlight the diverse roles memory can play in supporting LLM capabilities and inform the operational framework we propose.

To ground our operational framework, we conduct a pilot study and define four core topics. These topics span complementary dimensions of memory research and represent critical frontiers in developing capable AI agents:

\begin{itemize}
    \item \textbf{Long-Term Memory} (temporal), focusing on memory management, utilization, and personalization in multi-session dialogue systems \citep{xu2021beyond, maharana2024evaluating}, retrieval-augmented generation (RAG), personalized agents \citep{li2024hello}, and question answering   \citep{wu2024longmemeval, zhong2024memorybank}.
    
    \item \textbf{Long-Context Memory} (contextual), addressing both parametric efficiency (e.g. "KV cache eviction" \citep{zhang2023ho}) and context utilization effectiveness (e.g., long-context compression \citep{cheng2024xrag, jiang-etal-2024-longllmlingua}) in handling extended sequences.
    
    \item \textbf{Parametric Memory Modification} (model-internal), covering model editing \citep{fang2025alphaedit, meng2022mass, wang2024wise}, unlearning \citep{maini2024tofu}, and continual learning \citep{wang2024towards} for adapting internal knowledge representations.
    
    \item \textbf{Multi-Source Memory} (cross-source), emphasizing integration across heterogeneous textual sources \citep{hu2023chatdb} but also multi-modal inputs \citep{wang2025new} to further support robust and scene-awareness reasoning.
\end{itemize}

\subsection{Research Methodology}

To provide a systematic and comprehensive view of memory-related research, we first analyzed 37 seed papers that are widely recognized as foundational or representative in the memory-for-LLM literature. Through expert annotation and iterative discussion, these papers were used to define our taxonomy of memory types and core operations and to manually identify four primary research topics: long-term memory, long-context memory, parametric memory modification, and multi-source memory. These topics were selected because they 1) represent the areas most closely related to and actively studied within memory-centric LLM systems, 2) reflect distinct operational challenges across four complementary dimensions including temporal (such as persistence and personalization in long-term usage), 3) contextual (such as efficient handling and compression of long sequences), 4) model-internal (such as updating or editing knowledge within parametric representations), and 5) modality and integration (such as aligning and reasoning across heterogeneous or multi-modal sources), and collectively capture the breadth of recent developments from dialogue agents to retrieval-augmented reasoning systems.

Building on this framework, we collected a large-scale corpus of over 30,000 papers published in top NLP and ML venues including NeurIPS, ICLR, ICML, ACL, EMNLP, and NAACL between 2022 and 2025, a period marked by the rapid emergence and evolution of large language models. Each paper abstract was evaluated using a GPT-based relevance scoring pipeline. Considering both cost and effectiveness, we selected GPT-4o-mini for its strong zero-shot reasoning ability and efficiency. Papers scoring $\geq 8$ out of 10 according to our taxonomy-aligned task definitions were retained, yielding a curated set of 3,923 high-relevance papers. To ensure reliability, we conducted \textbf{manual validation} and \textbf{recall checks} on randomly sampled subsets, confirming that the threshold of 8 provides a balanced trade-off between precision and recall.

To highlight impactful work while mitigating publication-age bias, we introduced the Relative Citation Index (RCI), a log-log regression based, time-normalized metric adapted from the RCR framework \citep{10.1371/journal.pbio.1002541}. RCI adjusts raw citation counts according to publication age, enabling fair comparisons across papers and years. Empirical results showed that the log-log regression model achieved the best fit ($R^2=0.97$) and produced intuitive outcomes, with expected citations converging to zero for newly released papers. By integrating semantic relevance filtering and RCI-based impact assessment, we establish a balanced and reproducible foundation for analyzing research progress, trends, and topic-specific impact dynamics across the four core areas.

\subsection{Contribution and Structure}

This survey contributes to both the research and industrial communities by offering a comprehensive and structured perspective on memory in AI agents. For the research community, our sruvey establishes a comprehensive conceptual foundation. It not only systematically organizes memory representations, types, and core operations, but also frames frontier topics through the lens of the memory lifecycle to elucidate how memory is encoded, evolved, and adapted in AI agents. For the industrial community, it provides an extensive overview of tools, products, and benchmarks, coupled with analyses of their functionalities and deployment scenarios. Thereby, our survey serves as a practical reference for designing and implementing memory-enabled applications. Furthermore, this survey synthesizes emerging trends and outstanding challenges, outlining promising avenues for future research and development in this rapidly evolving domain.

\begin{figure}[t!]
    \centering
    \includegraphics[width=\linewidth]{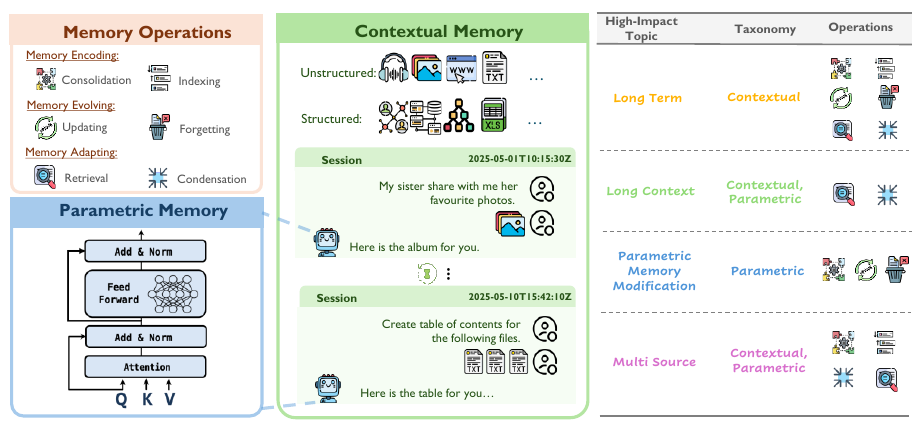}
    \caption{A unified framework of memory: Taxonomy, Core Operations, and Applications in LLM-based agents.}
    \label{fig:memory-structure}
\end{figure}

The remainder of the paper is organized as follows. Section~\ref{sec:memory_foundations} provides readers with a comprehensive understanding of memory representation, memory types, functional memory categories, and core memory operations, forming a solid foundation for studying memory in agents. Section~\ref{sec:memory_topics} maps high-impact topics to this foundation and summarizes key methods and datasets.  Section~\ref{sec:memory_practice} outlines real-world applications, products, and practical tools for building memory-enabled AI systems.  Section~\ref{sec:human_vs_agent_memory} compares human and agent memory systems, highlighting operational parallels and differences. Section~\ref{sec:future_direction} concludes with future directions for memory-centric agent (see Figure~\ref{fig:memory-structure} for an overview).

\section{Memory Foundations}
\label{sec:memory_foundations}

Memory in agents can be understood through four complementary dimensions: \textbf{representation},  \textbf{timescale},  \textbf{functional type}, and  \textbf{operations}. These perspectives jointly describe \textit{what memory is}, \textit{how long it persists}, \textit{what role it serves}, and \textit{how it evolves}. Representation defines the structural form of stored knowledge, timescale characterizes its temporal persistence, functional type captures the cognitive or computational role of stored content, and operation describes the dynamic processes that govern encoding, evolving, and adapting. Together, they provide an integrated framework linking the structure, function, and dynamics of memory in both human cognition and artificial intelligence. Together, these dimensions form the core dimensions for analyzing memory mechanisms in agents.

\subsection{Memory Representation}
\label{sec:memory_taxonomy}
From the perspective of memory representation, we divide memory into \textbf{Parametric Memory} and \textbf{Contextual Memory}, where the latter comprises both \textit{Unstructured} and \textit{Structured} forms.

\textbf{Parametric Memory} refers to the knowledge implicitly stored within model internal parameters \citep{berges2024memory, wang2024wise, prashanth2024recite}. Acquired during pretraining or post-training, this memory is embedded in the model’s weights and accessed through feedforward computation at inference. It serves as a form of long-term and persistent memory enabling fast, context-free retrieval of factual and commonsense knowledge. However, it lacks transparency and is difficult to update selectively in response to new experiences or task-specific contexts.

\textbf{Contextual Memory} denotes explicit, external information that complements model parameters and is categorized into unstructured and structured forms. \textit{\textbf{Contextual Unstructured Memory}} refers to an explicit and dynamically evolving memory system that stores, retrieves, and updates situational information in unstructured formats such as text \citep{zhong2024memorybank}, images \citep{wang2025new}, audio \citep{long2025seeinglisteningrememberingreasoning} , videos \citep{ wang2023lifelongmemory, yu2025context} or embeddings, derived from users, systems, and environments. It captures temporal \citep{ge2025tremu}, emotional \citep{yang-etal-2024-iterative}, procedural \citep{fang2025memp}, and semantic aspects of context without predefined schemas, enabling adaptive reasoning and continuity across interactions. The short-term form of Contextual Unstructured Memory includes concatenated prompts and Key–Value (KV) cache \citep{yang2024memory3}, which retains token-level representations during inference to maintain local coherence and efficient context reuse. In contrast, its long-term form extends to memory buffers \citep{packer2023memgpt}, retrieval databases \citep{wang2024symbolic}, or episodic storage \citep{hu2025evaluating} that store and refine contextual signals over time to support personalization and lifelong learning \citep{li2024hello}. Meanwhile, \textit{\textbf{Contextual Structured Memory}} denotes an explicit memory organized into predefined, interpretable formats or schemata, such as knowledge graphs \citep{wang-etal-2024-absinstruct,oguz-etal-2022-unik}, relational tables \citep{lu2023memochat}, experiences \citep{ouyang2025reasoningbank} or ontologies \citep{qiang2023agent}, which remain easily queryable. These structures support symbolic reasoning and precise querying, often complementing the associative capabilities of pretrained language models (PLMs). While usually used as long-term memory, structured memory can be short-term, constructed at inference for local reasoning, or long-term, storing curated knowledge across sessions.

\subsection{Memory Timescale}
Besides the representation, the timescale serves as another critical dimension. Drawing on the temporal persistence of information, memory is typically categorized into long-term and short-term memory.

\textbf{Long-term Memory} refers to a cognitive system with virtually unlimited capacity for storing information over extended periods of time, ranging from hours to an entire lifetime \citep{sumers2024cognitive}. In LLM agents, it refers to the ability to store, manage, and utilize information persistently across extended interactions with extended environment. This capability is essential for enabling continuity, personalization, and knowledge grounding in real-world applications such as multi-session agents \citep{ge2025tremu}, retrieval-augmented generation (RAG) \citep{qian2024memorag}, personalized assistants \citep{du2024perltqa, zhong2024memorybank}, and long-term planning agents \citep{xu2025mem}. It encompasses both contextual memory (such as dialogue histories \citep{hu2025evaluating} and user-specific preferences \citep{zhang-etal-2024-llm-based, memobase2025} and parametric memory (knowledge encoded within model parameters \citep{cao2025memory}). Meanwhile, it is closely aligned with functional perspectives, including semantic \citep{wang2024symbolic}, episodic \citep{wang2025episodic, liu2025echo}, procedural \citep{fang2025memp} memory, which will be introduced in the following sections.

\textbf{Short-term Memory} refers to the temporary storage of information for immediate use \citep{lumer2025memtool}. In LLM-based agents, it typically denotes the KV cache \citep{yang2024memory3} or current context window \citep{packer2023memgpt}, which holds task-relevant information to support real-time reasoning and decision-making. Similar to human cognition, this short-term memory can be consolidated into long-term memory through processes such as summarization \citep{lu2023memochat} and storage in external databases or model parameters. In practice, short-term memory is especially critical in long-context scenarios, where it helps mitigate hallucinations, address the “lost in the middle” problem \citep{liu-etal-2024-lost} in ultra-long contexts \citep{yu2025memagent}, reduce error accumulation in multi-turn interactions \citep{zhou2025mem1}, and enhance the reliability of multi-turn tool usage \citep{lumer2025memtool}.

\subsection{Memory Functional Type}
Beyond temporal persistence, memory can also be characterized by its functional roles in supporting agentic intelligence. Drawing on cognitive science, we further distinguish memory into episodic, semantic, procedural, and working memory.

\textbf{Episodic memory}, a core type of \textit{long-term memory} originating from cognitive psychology \citep{tulving1972episodic}, refers to the storage of past experiences linked to temporal cues, events, dialogue histories, and spatial contexts, and it dynamically evolves as the environment changes. It is widely regarded as a form of long-term memory \citep{fountas2024human} and, in modern agent systems, often functions as an external memory module \citep{pink2025position} that complements parametric knowledge. Recent work on agents \citep{miao2024episodic} increasingly explores how to update episodic memories \citep{das2024larimar,li2024linking}, perform temporal reasoning \citep{ge2025tremu}, and retrieve and utilize relevant experiences \citep{miao2024episodic} to enhance adaptability and decision-making in dynamic environments \citep{yan2025memory}.

\textbf{Semantic memory} another fundamental form of \textit{long-term memory}, refers to memory for facts concepts about the world \citep{tulving1972episodic}. In computational systems, it is often formalized into explicit, queryable structures such as knowledge graphs \citep{rasmussen2025zep}, relational tables \citep{hu2023chatdb}, or implicit model parameters \citep{packer2023memgpt}. Within model parameters, semantic knowledge is encoded in distributed representations that capture general world facts and concepts learned during pretraining. In contrast to context-dependent episodic memory, semantic memory is relatively stable, generalizable, and abstracted from cumulative experiences. While in LLM, the boundary between semantic and episodic memory is often blurred, as parametric representations may intertwine factual knowledge with contextual associations. This integration of implicit and explicit semantic memory provides the foundation for memory-augmented reasoning and adaptive knowledge use in modern agents \citep{zhou2025mem1}.

\textbf{Procedural memory}, also categorized as a form of \textit{long-term memory}, refers to memory that supports the execution of learned skills and action sequences without conscious awareness of prior experiences \citep{fang2025memp, ouyang2025reasoningbank, zhou2025agentfly}. In intelligent agents, procedural memory is typically formed in two ways: stored explicitly in external skill repositories for reuse \citep{zhou2025agentfly}, or encoded implicitly through large-scale training \citep{liu2024deepseek}. Training on execution data like trajectories and CoT reasoning fosters consistent task performance, particularly in tool-augmented \citep{lumer2025memtool} and RL-based systems \citep{yu2025memagent}. Procedural memory underpins the automation and generalization of task-oriented behaviors. 

\textbf{Working Memory}, a functional extension of \textit{short-term memory}, functions as a \textit{dynamic control mechanism} that not only temporarily stores information but also actively manipulates and updates it to support ongoing cognition \citep{wang2025graphcogent, rasmussen2025zep}. Its primary function is to actively select and integrate information from diverse sources, such as short-term context (e.g., dialogue history) and activated long-term memory (e.g., retrieved knowledge or parametric outputs) and transient computational buffers like the \textbf{Key–Value (KV) cache} \citep{chen2024improving, mem0}. In practice, working memory acts as the control layer \citep{li2025memos} for the agent context window, dynamically assembling the necessary inputs, including retrieved reasoning experience \citep{ouyang2025reasoningbank}, tool outputs \citep{lumer2025memtool}, and user data \citep{wang2025new}, to support complex reasoning, planning, and goal-directed behavior.

\subsection{Memory Operations}
\label{sec:memory_operation}
To enable dynamic memory beyond static storage, modern agents require operations that govern the lifecycle of information and support its effective use during interaction with the external environment. These operations can be grouped into three functional categories: Memory Encoding, Memory Evolving, and Memory Adapting.

\subsubsection{Memory Encoding}
\textbf{Memory encoding} governs how information is transformed into storable representations and linked for later retrieval. It primarily involves two complementary processes: Consolidation and Indexing. These operations naturally incorporate the temporal nature of memory, where information evolves over time.

\textbf{Consolidation} \citep{squire2015memory} refers to transforming $m$ short-term experiences $\mathcal{E}{[t, t+\Delta{t}]} = (\epsilon_{1}, \epsilon_{2}, \dots, \epsilon_{m})$ between $t$ and $t+\Delta_{t}$ into persistent memory $\mathcal{M}_{t}$. It encodes interaction histories (e.g., dialogs, trajectories) into durable forms such as model parameters \citep{wang2024towards}, graphs \citep{zhao2025eventweave}, or knowledge bases \citep{lu2023memochat}. It is essential for continual learning \citep{feng2024tasl}, personalization \citep{zhang-etal-2024-llm-based}, external Memory Bank construction \citep{zhong2024memorybank}, and knowledge graph construction \citep{xu2024generate}.
\begin{equation}
    \mathcal{M}_{t+\Delta_{t}}=\texttt{Consolidate}(\mathcal{M}_{t}, \mathcal{E}_{[t, t+\Delta_{t}]})
\end{equation}
\textbf{Indexing} \citep{maekawa2023generative} constructs auxiliary codes $\phi$ such as entities, attributes, or content-based representations \citep{wu2024longmemeval} that serve as access points to stored memory. Beyond access, indexing encodes temporal \citep{maharana2024evaluating} and relational structures \citep{mehta2022dsi++} across memories, enabling efficient and coherent retrieval through traversable index paths. It further supports scalable retrieval across symbolic, neural, and hybrid memory systems.
\begin{equation}
    \mathcal{I}_{t} = \texttt{Index}(\mathcal{M}_{t}, \phi)
\end{equation}

\subsubsection{Memory Evolving}

\textbf{Memory evolving} describes how stored information dynamically changes over time through two complementary processes: \textit{memory updating} and \textit{memory forgetting}.

\textbf{Updating} \citep{kiley2022mechanisms} reactivates existing memory representations in $\mathcal{M}_t$ and temporarily modify them with new knowledge $\mathcal{K}_{t+\Delta_{t}}$. Updating parametric memory typically involves a locate-and-edit mechanism \citep{fang2025alphaedit} that targets specific model components. Meanwhile, contextual memory updating involves summarization \citep{zhong2024memorybank}, pruning, or refinement \citep{bae2022keep} to reorganize or replace outdated content. Those updating operations support continual adaptation while maintaining memory consistency. 
\begin{equation}
    \mathcal{M}_{t+\Delta_{t}} = \texttt{Update}(\mathcal{M}_{t}, \mathcal{K}_{t+\Delta_{t}})
\end{equation}
\textbf{Forgetting} \citep{davis2017biology, wang2009concept} is the ability to selectively suppress memory content $\mathcal{F}$ from $\mathcal{M}_{t}$ that may be outdated, irrelevant, or harmful. In parametric memory, it is commonly implemented through unlearning techniques \citep{jia2024wagle, machine-unlearning-lina-2025} that modify model parameters to erase specific knowledge. In contextual memory, forgetting involves time-based deletion \citep{zhong2024memorybank} or semantic filtering \citep{wang2024machine} to discard content that is no longer relevant. These operations help maintain memory efficiency and reduce interference.
\begin{equation}
    \mathcal{M}_{t+\Delta_{t}} = \texttt{Forget}(\mathcal{M}_{t}, \mathcal{F})
\end{equation}
However, these operations introduce inherent risks and limitations. Attackers can exploit vulnerabilities to alter or poison memory contents. Once corrupted, memory fragments may persist undetected and later trigger malicious actions. As discussed in Section \ref{sec:future_direction}, such threats call for robust approaches that address not only the memory operations but also the entire memory lifecycle.

\subsubsection{Memory Adapting}
Memory adapting refers to how stored memory is retrieved and used during inference, encompassing two operations: retrieval and compression. \newline \textbf{Retrieval} is the process of identifying and accessing relevant information from memory in response to inputs, aiming to support downstream tasks such as response generation, visual grounding, or intent prediction.
Inputs $\mathcal{Q}$ can range from a simple query \citep{du2024perltqa} to a complex multi-turn dialogue context \citep{wang2025new}, and from purely textual inputs to visual content \citep{zhou2024vista} or even more modalities. Memory fragments are typically scored with a function \textit{sim()} with those above a threshold $\tau$ deemed relevant. Retrieval targets include memory from multiple sources \citep{tan-etal-2024-blinded}, modalities \citep{wang2025new}, or even parametric representations \citep{luo-etal-2024-landmark} within models. 
\begin{equation}
\begin{split}
    \texttt{Retrieve}(\mathcal{M}_{t}, \mathcal{Q}) &= m_{\mathcal{Q}} \in \mathcal{M}_{t} \\ \text{with } &\text{sim}(\mathcal{Q}, m_{\mathcal{Q}}) \geq \tau
\end{split}
\end{equation}
\newline \textbf{Condensation} enables efficient context usage under limited context window by retaining salient information and discarding redundancies with a compression ratio $\alpha$ before feeding it into models. It can be broadly divided into pre-input compression and post-retrieval compression. Pre-input compression applies in long-context models without retrieval, where full-context inputs are scored, filtered, or summarized to fit within context constraints \citep{yu-etal-2023-trams, chung-etal-2024-selection}. Post-retrieval compression operates after memory access, reducing retrieved content either through contextual compression before model inference \citep{xu2024recomp} or through parametric compression by integrating retrieved knowledge into model parameters \citep{safaya-yuret-2024-neurocache}. Unlike memory consolidation, which summarizes information during memory construction \citep{zhong2024memorybank}, compression focuses on reducing memory at inference \citep{pmlr-v235-lee24c}.
\begin{equation}
\begin{split}
    \mathcal{M}_{t}^{comp}=\texttt{Compress}(\mathcal{M}_{t}, \alpha)
\end{split}
\end{equation}

\definecolor{longtermcolor}{RGB}{251, 231, 158}   
\definecolor{longcontextcolor}{RGB}{132, 195, 103}  
\definecolor{parametriccolor}{RGB}{113, 184, 237}     
\definecolor{multisourcecolor}{RGB}{242, 168, 218}  

\begin{figure*}[t!]
  \begin{forest}
    forked edges,
    ver/.style={rotate=90, child anchor=north, parent anchor=south, anchor=center},
    for tree={
      grow=east,
      reversed=true,
      anchor=base west,
      parent anchor=east,
      child anchor=west,
      base=left,
      rectangle,
      rounded corners, 
      draw=lightblue,
      align=left,
      minimum width=3.5em,
      s sep=6pt,
      inner xsep=2pt,
      inner ysep=1pt,
    },
    where level=1{text width=4.2em,font=\scriptsize, draw=black, rounded corners}{},
    where level=2{text width=4.7em ,font=\tiny, draw=black, rounded corners}{},
    where level=3{text width=5.9em, font=\tiny, draw=black, rounded corners}{},
    where level=4{ font=\tiny, draw=black, rounded corners}{},
    where level=5{font=\tiny, draw=black, rounded corners}{}, 
    [Memory in AI Agent, ver
        [Long Term, ver,fill=longtermcolor!30, for descendants={fill=longtermcolor!30},
            [
            Encoding
                [Consolidation, text width=5em
                    [MyAgent \citep{hou2024myagent}{,} MemoChat \citep{lu2023memochat}{,} MemOS \citep{li2025memos}{,} LightMem \citep{zhang2025lightmem}, text width=20em
                    ]
                ]
                [Indexing, text width=5em
                    [HippoRAG \citep{gutierrez2024hipporag}{,} G-Memory \citep{zhang2025gmemory}{,}LongMemEval \citep{wu2024longmemeval}{,} GraphCogent\citep{wang2025graphcogent}, text width=20em
                    ]
                ]
            ]
            [Evolving
                [Updating, text width=5em
                    [Memory-R1 \citep{yan2025memory}{,} {O-Mem} \citep{wang2025omem}{,} NLI-transfer \citep{bae2022keep}{,} RCSum \citep{wang2025recursively}{,} Mem-$\alpha$ \citep{wang2025mem}, text width=20em]
                ]
                [Forgetting, text width=5em
                    [FLOW-RAG \citep{wang2024machine}{,} MemoryBank \citep{zhong2024memorybank}, text width=20em]
                ]
            ]
            [Adapting
                [Retrieval, text width=5em
                    [{LoCoMo \citep{maharana2024evaluating}{,} MemoChat \citep{lu2023memochat}{,} MemGuide \citep{DuEtAl2025MemGuide}{,} MemTool \citep{lumer2025memtool}}, text width=20em]
                ]
                [Condensation, text width=5em
                    [MoT \citep{li-qiu-2023-mot}{,} SCM \citep{wang2024enhancing}{,} Optimus-1 \citep{li2024optimus}{,} A-MEM\citep{xu2025mem}, text width=20em]
                ]
                [Generation, text width=5em
                    [{MEMORAG \citep{qian2024memorag}{,} ReadAgent \citep{pmlr-v235-lee24c}}{,} COMEDY \citep{chen2024compress}, text width=20em]
                ]
            ]
            [Personalization
                [Adaptation, text width=5em
                    [{PersonaMem-v2 \citep{jiang2025personamem}{,} Mem-U \citep{NevaMindAI2025memU}{,} MALP \citep{zhang-etal-2024-llm-based}{,} Per-Pcs \citep{tan2024personalized}}, text width=20em]
                ]
                [Augmentation, text width=5em
                    [{EMG \citep{wang2024crafting}{,} LDAgent \citep{li2024hello}}{,} PerLTQA \citep{du2024perltqa}, text width=20em]
                ]
            ]
        ]
        [Long Context, ver, fill=longcontextcolor!30, for descendants={fill=longcontextcolor!30}
            [Compression
            [Context Compression, text width=6em
                    [{
                    RECOMP \citep{xu2024recomp}{,} 
                    xRAG \citep{cheng2024xrag}{,}  
                    LongLLMLingua \citep{jiang-etal-2024-longllmlingua}{,}
                    AgentFold \citep{ye2025agentfoldlonghorizonwebagents}
                    }, text width=19em
                    ]
                ]
                [{KV Cache Eviction}, text width=6em
                    [
                    {H$_{2}$O \citep{zhang2023ho}{,} 
                    StreamingLLM \citep{xiao2024efficient}{,} 
                    SnapKV \citep{NEURIPS2024_28ab4182}{,}
                    RocketKV \citep{behnam2025rocketkv}
                    }, text width=19em
                    ]
                ]
                [KV Cache Storing \\ Optimization, text width=6em
                    [{LESS \citep{pmlr-v235-dong24f}{,}
                    KVQuant \citep{NEURIPS2024_028fcbcf}{,} 
                    KIVI \citep{pmlr-v235-liu24bz}{,}
                    ShadowKV \citep{sun2025shadowkv}
                    }, text width=19em
                    ]
                ]
            ]
            [Retrieval
                [Context Retrieval, text width=6em
                    [{GraphReader \citep{he2025graphiti}{,}
                    Ziya-Reader \citep{he-etal-2024-never}{,}}, text width=19em
                    ]
                ]
                [KV Cache Selection, text width=6em
                    [{QUEST \citep{pmlr-v235-tang24l}{,}
                    RetrievalAttention \citep{liu2024retrievalattentionacceleratinglongcontextllm}
                    }, text width=19em
                    ]
                ] 
            ]
        ]
        [Parametric Motification, ver, text width=8em,fill=parametriccolor!30, for descendants={fill=parametriccolor!30}
            [Editing
                [Locating then\\ Editing , text width=7em
                    [{ROME \citep{meng2022locating}{,} MEMIT \citep{meng2022mass}{,}  AlphaEdit\citep{fang2025alphaedit}{,} AnyEdit\citep{jiang2025anyedit}}{,} M2Edit\citep{zhou-etal-2025-m2edit}, text width=18em
                    ]
                ] 
                [Meta Learning , text width=7em
                    [KE \citep{de2021editing}{,} MEND \citep{mitchell2022fast}{,} DAFNET \citep{zhang2024dafnet}{,} MALMEN\citep{tan2024massive}, text width=18em]
                ] 
                [Prompt, text width=7em
                    [IKE \citep{zheng2023can}{,} MeLLo \citep{zhong2023mquake}{,} EditCoT~\citep{wang-etal-2025-knowledge-editing}{,} DR-IKE\citep{nafee-etal-2025-dynamic}{,} LTE\citep{jiang-etal-2024-learning}, text width=18em]
                ] 
                [Additional Parameters, text width=7em
                    [CaliNET \citep{dong2022calibrating} {,} SERAC \citep{mitchell2022memory}{,} Titans \citep{behrouz2024titans} {,} MLP-Memory \citep{wei2025mlp}, text width=18em]
                ]
            ]
            [Unlearning
                [Locating then\\ Unlearning , text width=7em
                    [{DEPN \citep{wu2023depn}{,} MemFlex \citep{tian2024forget}{,} WAGLE \citep{jia2024wagle}{,} NeuMuter\citep{hou-etal-2025-decoupling}}, text width=18em]
                ]
                [Training Objective, text width=7em
                    [FLAT \citep{wang2025llm}{,} GA+Mismatch \citep{yao2024large}{,}  SOUL \citep{jia2024soul}{,} Relearn~\citep{xu-etal-2025-relearn}{,} UL\citep{jang-etal-2023-knowledge}, text width=18em]
                ] 
                [Prompt, text width=7em
                    [ICUL\citep{pawelczyk2023context}{,} ECO\citep{liu2024large}{,} SEPS\citep{jeung-etal-2025-seps}{,} ReversingIKE\citep{youssef-etal-2025-make}{,} ERASE\citep{muresanu2025fast},  text width=18em]
                ]
                [Additional Parameters, text width=7em
                    [ULD\citep{ji2024reversing}{,} EUL\citep{chen2023unlearn}{,} LoKU\citep{cha2025towards}{,} LLMEraser\citep{ding2025unified}{,} S3T\citep{chowdhury2025towards}, text width=18em]
                ] 
            ]
            [{Lifelong \\ (Continual) \\ Learning}
                [{Regularization-based Learning}, text width=9em
                    [TaSL \citep{feng2024tasl}{,} SELF-PARAM \citep{wangself}, text width=16em]
                ]
                [{Replay-Based Learning}, text width=9em
                    [DSI++ \citep{mehta2022dsi++}{,} Memento \citep{zhou2025agentfly}{,} Reasoning Bank \citep{ouyang2025reasoningbank}, text width=16em]
                ]
                [Interactive Learning, text width=9em 
                    [LSCS \citep{wang2024towards}{,} ACE \citep{zhang2025agentic}{,} Early Experience \citep{zhang2025agent}, text width=16em]
                ]
            ]
        ]
        [Multi-source, ver, fill=multisourcecolor!30, for descendants={fill=multisourcecolor!30}
            [{Cross-Textual\\ Integration}
                [Reasoning, text width = 5em
                    [ Mirix \citep{wang2025mirix}{,} StructRAG \citep{li2024structrag}{,} ChatDB \citep{hu2023chatdb}, text width=20em] 
                ]
                [Conflict, text width = 5em
                    [RKC-LLM \citep{wang2023resolving}{,} BGC-KC \citep{tan-etal-2024-blinded}, text width=20em]
                ]
            ]
            [{Multi-modal\\ Coordination}
                [Fusion, text width = 5em
                    [{LifelongMemory \citep{wang2023lifelongmemory}{,} Ma-llm\citep{he2024ma}{,} $M^3$\citep{long2025seeinglisteningrememberingreasoning}}, text width=20em]
                ]
                [Retrieval, text width = 5em
                    [VISTA \citep{zhou2024vista}{,} IGSR \citep{wang2025new}{,} MMLongBench \citep{wang2025mmlongbench}, text width=20em] 
                ]
            ]
        ]
    ]
  \end{forest}
  \caption{Operation-driven key research topics in AI agents, mapping core memory operations to four key research topics.}
  \label{fig:nlm_to_plm}
\end{figure*}

\section{From Operations to Key Research Topics}

\label{sec:memory_topics}

This section analyzes how real-world systems manage and utilize memory through core operations. We examine four key research topics introduced in Section~\ref{sec:introduction}, guided by the framework in Figure~\ref{fig:memory-structure}, using the Relative Citation Index (RCI)—a time-adjusted metric that normalizes citation counts by publication age to highlight influential work. RCI surfaces emerging trends and enduring contributions across memory research. Figure~\ref{fig:nlm_to_plm} shows the architectural landscape of these topics.

\subsection{Long-term Memory}
Long-term memory, as a research topic, examines how agents preserve and leverage information across extended interactions to achieve continuity, personalization, and cumulative learning. In this section, we discuss contextual long-term memory as a functional system that integrates both structured and unstructured forms, highlighting its core operations of encoding, evolving, and adapting in supporting complex reasoning and temporal coherence.

\subsubsection{Memory Encoding} Memory Encoding is the foundational process of transforming raw inputs —such as dialogue histories or agent observations—into durable representations suitable for long-term storage. This process is realized through two critical and complementary operations: Memory Consolidation and Memory Indexing.


\textit{\textbf{Memory Consolidation}} plays a central role in shaping long-term memory by stabilizing short-term context into enduring representations. In LLM-based agents, consolidation unfolds across multiple levels: (1) \textit{dialogue summarization or structuring} converts interaction histories into retrievable traces \citep{zhong2024memorybank, wang-etal-summarizing-dialogue-memory-2025, lu2023memochat, hou2024myagent}; (2) \textit{reasoning experience consolidation} encodes successful tool-use trajectories and problem-solving strategies \citep{ouyang2025reasoningbank, fang2025memp}; (3) \textit{parametric consolidation} embeds stable knowledge directly into model parameters through methods such as continual pretraining \citep{jiang2024long}, supervised finetuning \citep{chu2025sft}, or reinforcement learning \citep{wang2025mem, yan2025memory}; (4) \textit{event-level consolidation} organizes episodic information into structured event graphs \citep{zhao2025eventweave}; (5) \textit{knowledge-level consolidation} populates knowledge graphs with factual triples for symbolic reasoning \citep{rasmussen2025zep}. Collectively, these processes extend the agent's temporal memory horizon, enabling the persistent retention of contextual, episodic, and semantic memory across extended interaction with the external environment. Nevertheless, robust long-term consolidation remains challenging-requiring the balance between stability and adaptability, mitigating context loss from over-compression, and preserving relevance under continuous updates \citep{zhang2025agentic}. This highlights the critical need for dynamic consolidation strategies within complex, evolving memory systems.

\textit{\textbf{Memory Indexing}} provides the foundational structure for long-term memory, transforming vast collections of experiences into a searchable repository that enables efficient and accurate retrieval. Recent work categorizes memory indexing into three paradigms: graph-based indexing, exemplified by HippoRAG \citep{gutierrez2024hipporag}, constructs lightweight knowledge graphs to explicitly map the relational structure between memory fragments; signal-enhanced indexing, where systems like LongMemEval \citep{wu2024longmemeval} enrich memory keys with metadata such as timestamps or summaries to refine retrieval accuracy; and timeline-based indexing, as demonstrated in Theanine \citep{ong2025towards}, which organizes memories along temporal and causal chains to enable chronologically-informed retrieval. These strategies highlight the need to integrate structure, retrieval signals, and temporal dynamics for effective long-term memory management. These paradigms signal a shift from simple semantic similarity to a crucial synthesis of relational structure, metadata signals, and temporal dynamics, enabling the development of scalable and contextually aware memory systems.

\subsubsection{Memory Evolving}
Memory Evolving involves operations such as forgetting and updating. Here, memory is dynamically refined through the incorporation of new knowledge, the correction of outdated or erroneous content, and the selective removal of low-value information. These processes ensure that the memory remains accurate, efficient, and contextually relevant, enabling agents to adapt to evolving tasks and environments.

\textit{\textbf{Memory Updating}} is the dynamic process of maintaining the internal consistency and accuracy of long-term memory by continually creating new representations \citep{chen2024compress}, integrating them with existing knowledge, and pruning outdated or irrelevant information \citep{bae2022keep}. Recent research are broadly categorized as either intrinsic or extrinsic. \textit{Intrinsic Updating} operates through self-contained processes to refine its knowledge base: selective editing \citep{bae2022keep} improves memory by selectively deleting outdated information; recursive summarization \citep{wang2025recursively} compresses dialogue histories through iterative summarization; memory blending merges past and present representations to form evolved insights \citep{kim2024ever}; and self-reflective evolving enhances factual consistency by verifying memories against retrieved evidence \citep{sun-etal-2024-towards-verifiable}. \textit{Extrinsic Updating} relies on external signals, such as incorporating direct user corrections into memory to enable continual system improvement \citep{dalvi-mishra-etal-2022-towards}. Ultimately, the success of any memory update hinges on its ability to integrate new information without corrupting critical prior knowledge or violating the factual and stylistic consistency.

\begin{figure}[t!]
  \centering
  \begin{minipage}[c]{0.48\textwidth}
    \includegraphics[width=\linewidth]{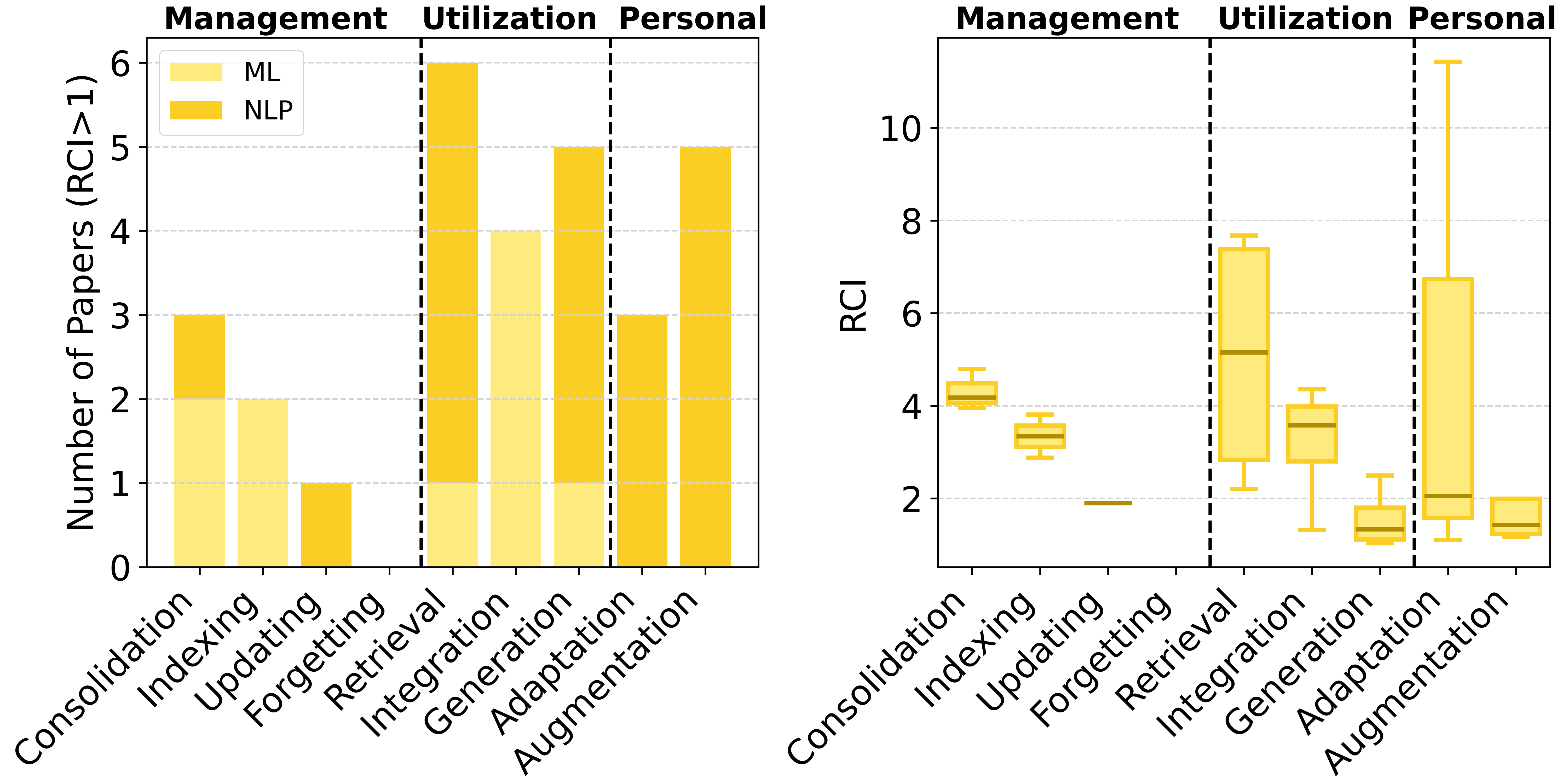}
    \caption{Publication statistic of highlighted papers (RCI > 1) discussed in long-term memory.}
    \label{fig:longterm-pub}
  \end{minipage}%
  \hfill
  \begin{minipage}[c]{0.48\textwidth}

\textit{\textbf{Memory Forgetting}} involves the removal of previously consolidated long-term memory representations. Distinct from the passive decay studied in cognitive psychology, forgetting in LLM agents is an active, targeted "unlearning" of consolidated information, driven by the need to expunge sensitive, harmful, or private content to meet external constraints \citep{chen2024compress, mitchell2022memory}. Consequently, developing robust unlearning capabilities for safety, privacy, and compliance has become a critical research focus \citep{liu2024towards,eldan2024whos,ji2024reversing,machine-unlearning-lina-2025, machine-unlearning-threat-survey-ziyao-etal-2025}. The foremost challenge is to achieve precise and complete removal of targeted data without causing collateral damage to the integrity and performance of the remaining valid knowledge.
  \end{minipage}
\end{figure}

\subsubsection{Memory Adapting}
Memory adapting focuses on retrieving and condensing relevant information from stored long-term memory to support reasoning, decision-making, and generation. It bridges the gap between the vast, passive repository of stored knowledge and the immediate context required for effective reasoning and generation. \textbf{Memory Retrieval} selects relevant information, while \textbf{Memory Condensation} transforms that information into a structured, compact context for the model to use. The ultimate success of these operations is measured by their ability to support the final stage of \textbf{Memory Grounded Generation}.

\textit{\textbf{Memory Retrieval}} focuses on selecting the most relevant memory entries for a given query. Retrieval methods can be categorized into three primary paradigms: (1) \textit{query-centered retrieval}, which refines the query itself for better search accuracy, as seen in FLARE \citep{jiang-etal-2023-active} and IterCQR \citep{jang-etal-2024-itercqr}; (2) \textit{memory-centered retrieval}, which improves the organization and ranking of stored information through enhanced indexing \citep{wu2024longmemeval} or reranking \citep{du2024perltqa}; and (3) \textit{event-centered retrieval}, which leverages temporal and causal structures for context-aware selection, as explored in LoCoMo \citep{maharana2024evaluating} and MSC \citep{xu2021beyond}. While techniques like multi-hop graph traversal further enrich this process \citep{gutierrez2024hipporag}, the core challenge remains in developing adaptive retrieval strategies that can dynamically adjust to the evolving structure and relevance of the memory store itself.

\textit{\textbf{Memory Condensation}} is the inference-time process of transforming raw, retrieved long-term memories into a structured and compact context for the LLM. Integration may span multiple memory sources (e.g., long-term dialogue histories, external knowledge bases) and modalities (e.g., text, images, or videos), enabling richer and contextually grounded generation. Recent efforts on memory integration can be broadly categorized into two strategies. \textbf{Static contextual integration} approaches, such as EWE \citep{chen2024improving} and Optimus-1 \citep{li2024optimus}, focus on retrieving and combining static memory entries at inference time to enrich context and improve reasoning consistency. In contrast, \textbf{dynamic memory evolving} approaches, exemplified by A-MEM \citep{hou2024myagent}, Synapse \citep{zheng2024synapse}, R2I \citep{samsami2024mastering}, and SCM \citep{wang2024enhancing}, emphasize enabling memory to grow, adapt, and restructure over the course of interactions, either through dynamic linking or controlled memory updates. While static integration strengthens immediate contextual grounding, recent work has transformed condensation into a more agentic paradigm, Agentic Context Engineering (ACE) \citep{zhang2025agentic}, in which an autonomous agent proactively refines, prioritizes, and restructures retrieved contexts to maximize reasoning efficiency. This agent-driven evolution of memory condensation represents a crucial step toward building adaptive, self-improving, and lifelong learning agents.

\textit{\textbf{Memory Grounded Generation}} can be broadly categorized into three types based on how memory influences generation. \textit{Self-Reflective Reasoning} uses memory of prior thinking processes to guide guide intermediate reasoning steps, such as MoT \citep{li-qiu-2023-mot} and StructRAG \citep{li2024structrag}. \textit{Feedback-Guided Correction}leverages knowledge of past errors or user feedback to constrain decoding and prevent their repetition \citep{qian2024memorag, tandon2021learning}; \textit{Contextually-Aligned Long-Term Generation} integrates summaries of distant history to maintain coherence throughout long dialogues or documents \citep{chen2024compress, lu2023memochat}. The primary challenge across all these methods is mitigating the impact of noise or inaccuracies from the earlier retrieval operations, ensuring the final output is both reliable and factually grounded.

\subsubsection{Personalization}
Personalization is key but challenging for long-term memory, limited by data sparsity, privacy, and changing user preferences. Current methods can be broadly categorized into two lines: model-level adaptation and external memory augmentation.

\textit{\textbf{Model-Level Adaptation}} encodes user preferences into model parameters via fine-tuning or lightweight updates. One strategy involves embedding user traits into a latent space, where methods like CLV use contrastive learning to cluster persona representations that guide generation \citep{tang-etal-2023-enhancing-personalized}. A more prevalent strategy employs parameter-efficient techniques; for instance, RECAP injects user histories via a prefix encoder \citep{liu2023recap}, while Per-Pes assembles modular adapters that reflect user behaviors \citep{tan2024personalized}. In specialized domains, MaLP \citep{zhang-etal-2024-llm-based} introduces a dual-process memory for modeling short- and long-term personalization in medical dialogues. The central challenge for this paradigm is managing the personalization-generalization trade-off: effectively specializing the model to an individual without compromising its broad, pre-trained capabilities.

\textit{\textbf{External Memory Augmentation}} personalizes responses by retrieving user-specific information from an external repository at inference time. This approach varies by memory format: structured memories like user profiles or knowledge graphs are used to create personalized prompts in LaMP \citep{salemi2023lamp}; unstructured memories, such as dialogue histories, provide rich contextual data for alignment in systems like LAPDOG \citep{huang-etal-2023-learning}; and hybrid systems like SiliconFriend \citep{zhong2024memorybank} maintain persistent, cross-session memory stores. While these approaches scale well, they often treat long-term memory as a passive buffer, leaving its potential for proactive planning and decision-making largely untapped.

\begin{figure*}[t!]
  \centering
  \includegraphics[width=\textwidth]{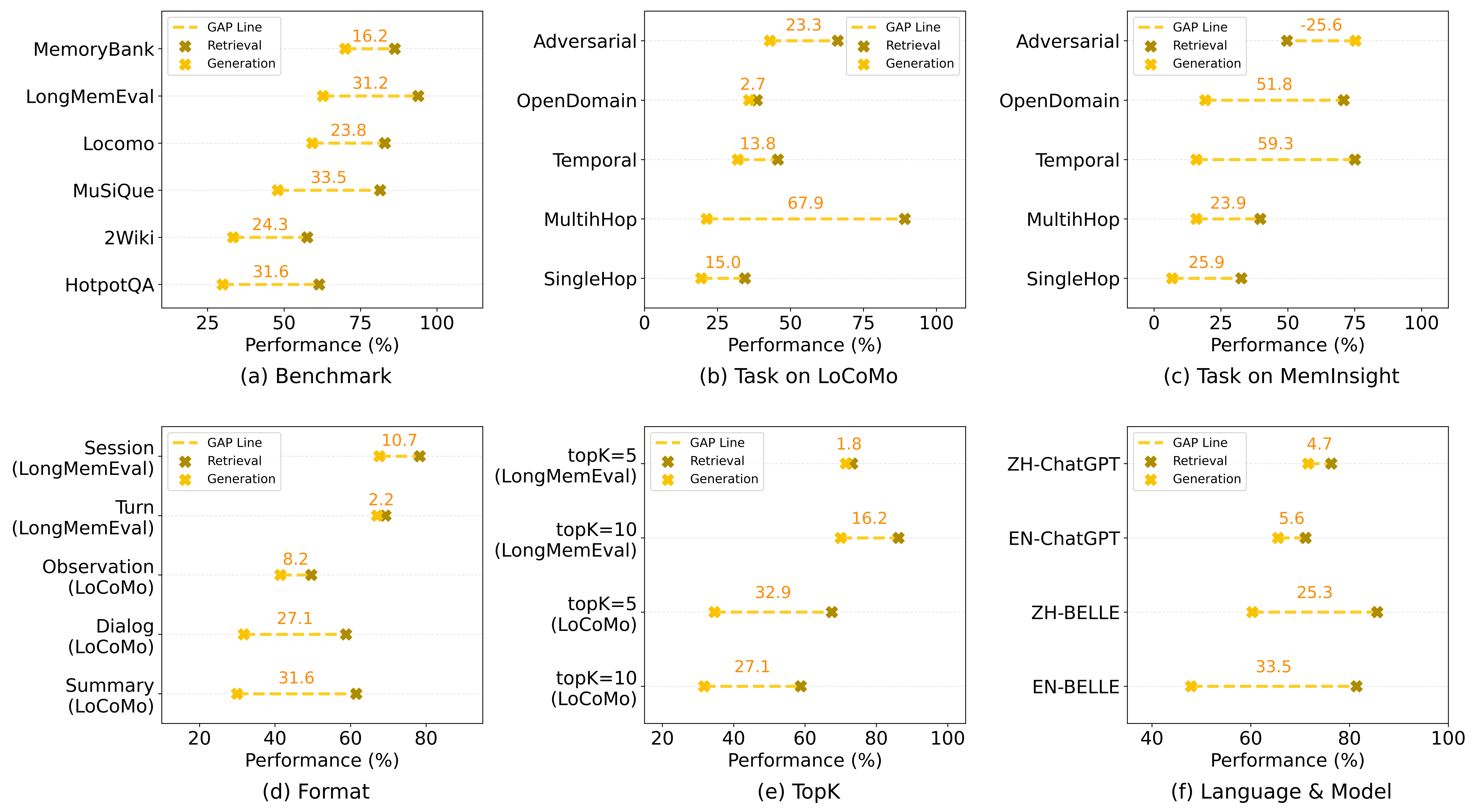}  
  \caption{Benchmark for evaluating \textbf{long-term memory}. “Mo” denotes modality. “Ops” denotes operability. “DS Type” indicates dataset type (QA – question answering, MS – multi-session dialogue). “Per” and “TR” indicate whether persona and temporal reasoning are present.}
  \label{fig:memory_retrieval_generation_gap}
\end{figure*}

\subsubsection{Discussion}
\label{sec:discussion}

\textbf{Long-term memory evaluation remains constrained by static assumptions.} Current benchmarks mainly follow two paradigms: knowledge-based question answering (QA) and multi-turn dialogue. QA tasks test a model’s ability to retrieve and reason over factual knowledge, leveraging both parametric memory \citep{yang2024memory3, berges2024memory, de2019episodic} and unstructured contextual memory \citep{salama2025meminsight, jin2024disentangling}. Techniques like self-evolution alignment \citep{zhang2025self} and salient memory distillation \citep{lu2023memochat, lanchantin2023learning} enhance factual grounding. However, these benchmarks often assume static memory and overlook dynamic operations such as updating, selective retention, and temporal continuity \citep{wu2024longmemeval, maharana2024evaluating}.
In contrast, multi-turn dialogue benchmarks (e.g., LoCoMo \citep{maharana2024evaluating}, LongMemEval \citep{wu2024longmemeval}) better capture real-world memory use by spanning 20–30 turns and enabling analyses of cross-session retrieval, updating, and event reasoning. Yet most still treat dialogue history as static context, focusing narrowly on QA accuracy while neglecting operations like indexing, consolidation, forgetting, and user adaptation. This static lens limits understanding of how memory evolves over time, especially in interactive settings requiring temporal adaptation. Recent work has begun addressing these challenges through agent-based systems \citep{xu2025mem} that integrate long-term memory into multi-turn planning and generation.

\paragraph{Mismatch between memory retrieval and memory-grounded generation reveals context engineering bottlenecks.} We analyze retrieval–generation performance gaps reported in recent studies \citep{gutierrez2024hipporag, maharana2024evaluating, wu2024longmemeval, zhong2024memorybank}. As shown in Figure~\ref{fig:memory_retrieval_generation_gap}, state-of-the-art models achieve Recall@5 above 90 on 2Wiki and MemoryBank \citep{gutierrez2024hipporag, zhong2024memorybank}, yet generation metrics (e.g., F1) lag by over 30 points. This indicates that high retrievability does not guarantee effective generation. Several factors contribute: compact memory formats (e.g., dialogue turns or task-level observations) better support generation than verbose entries; longer temporal distance between memory and query, as in MemInsight on LoCoMo \citep{salama2025meminsight}, degrades generation even with accurate retrieval, highlighting temporal reasoning as a key bottleneck in memory-grounded generation. Recent efforts such as TREMU \citep{ge2025tremu} attempt to address this via chain-of-thought supervision, yet empirical gains remain limited, further suggesting that long-horizon agents will increasingly encounter this constraint; retrieving more items introduces noise that impairs decoding; and multilingual settings reveal a persistent language gap, with English outperforming Chinese. These findings show that while current systems retrieve relevant memories, they remain limited in structuring and leveraging them for downstream generation.

\paragraph{Memory operations remain under-evaluated in current benchmarks.} Despite growing interest in memory-augmented models, current evaluations primarily focus on retrieval accuracy (e.g., Recall@k, Hit@k, NDCG) and post-retrieval generation quality (e.g., F1, BLEU, ROUGE-L), as seen in LoCoMo and LongMemEval. While some studies incorporate human assessments of memorability, coherence, and correctness, these efforts largely overlook procedural aspects of memory use—such as consolidation, updating, forgetting, and selective retention. Some recent efforts, such as MemoryBank and ChMapData-test \citep{wu2025interpersonal}, begin to address aspects of memory updating and long-term planning, but remain isolated and narrow in scope. There remains a pressing need for comprehensive benchmarks that span parametric, contextual unstructured, and structured memory, along with dynamic evaluation protocols that assess memory reliability, temporal reasoning, and multi-session dialogue consistency beyond static QA accuracy.

\paragraph{Publication Trend.} As shown in Figure~\ref{fig:longterm-pub}, retrieval and generation dominate recent literature, especially in NLP. Core operations like consolidation and indexing receive more attention in ML, while forgetting remains underexplored. Personalization is largely limited to NLP due to practical application needs. In terms of citation impact, consolidation, retrieval, and integration play key roles—driven by advances in memory-aware fine-tuning, summarization, retrieval-augmented generation, and prompt fusion.

\newcommand{\lightbulb}{
\begin{tikzpicture}[scale=0.15, line width=0.6pt]
  \shade[ball color=yellow!70!white] (0,0) circle (1.2);
  \draw[fill=gray!60] (-0.5,-1.2) rectangle (0.5,-1.6);
  \draw[fill=black] (-0.4,-1.6) rectangle (0.4,-2.0);
\end{tikzpicture}
}
\begin{tcolorbox}[
    colback=longtermcolor!30!white,
    colframe=black!50,
    arc=6pt,
    boxrule=0.8pt,
    width=\linewidth,
    enhanced,
    left=4pt,
    right=4pt,
    top=4pt,
    bottom=4pt,
    fontupper=\small\bfseries
]
\lightbulb \textbf{Shift evaluation from isolated memory operations toward systematic assessment of memory encoding, evolving, and adapting.}

\lightbulb \textbf{Effective methods for long-horizon temporal reasoning beyond dialogue are still lacking.}

\lightbulb \textbf{Addressing the retrieval–generation disconnect requires context engineering strategies that prioritize concise, reliable memory condensation.}

\lightbulb \textbf{Advance personalized agents by moving beyond memory storage toward adaptive reuse and personalization of session-spanning memories.}
\end{tcolorbox}

\definecolor{LT}{HTML}{fbefcf}
\subsection{Long-context Memory}

Managing vast quantities of multi-sourced external memory (short-term memory) in conversational search presents significant challenges in long-context language understanding. While advancements in model design and long-context training have enabled LLMs to process millions of input tokens \citep{ding2023longnetscalingtransformers1000000000,ding2024longropeextendingllmcontext}, effectively managing memory within such extensive contexts remains a complex issue. These challenges can be broadly categorized into two main aspects with respect to memory operations: 1) \textbf{Memory Compression}, which focuses on compressing the short-term memory of the context tokens or KV cache to enable efficient long context decoding and \textbf{Memory Retrieval} optimizes the selection of contextual memory for effective long context processing. In this section, we systematically review efforts made in handling these challenges.


\begin{figure}[t!]
  \centering
  \begin{minipage}[t]{0.48\textwidth}
    \includegraphics[width=\linewidth]{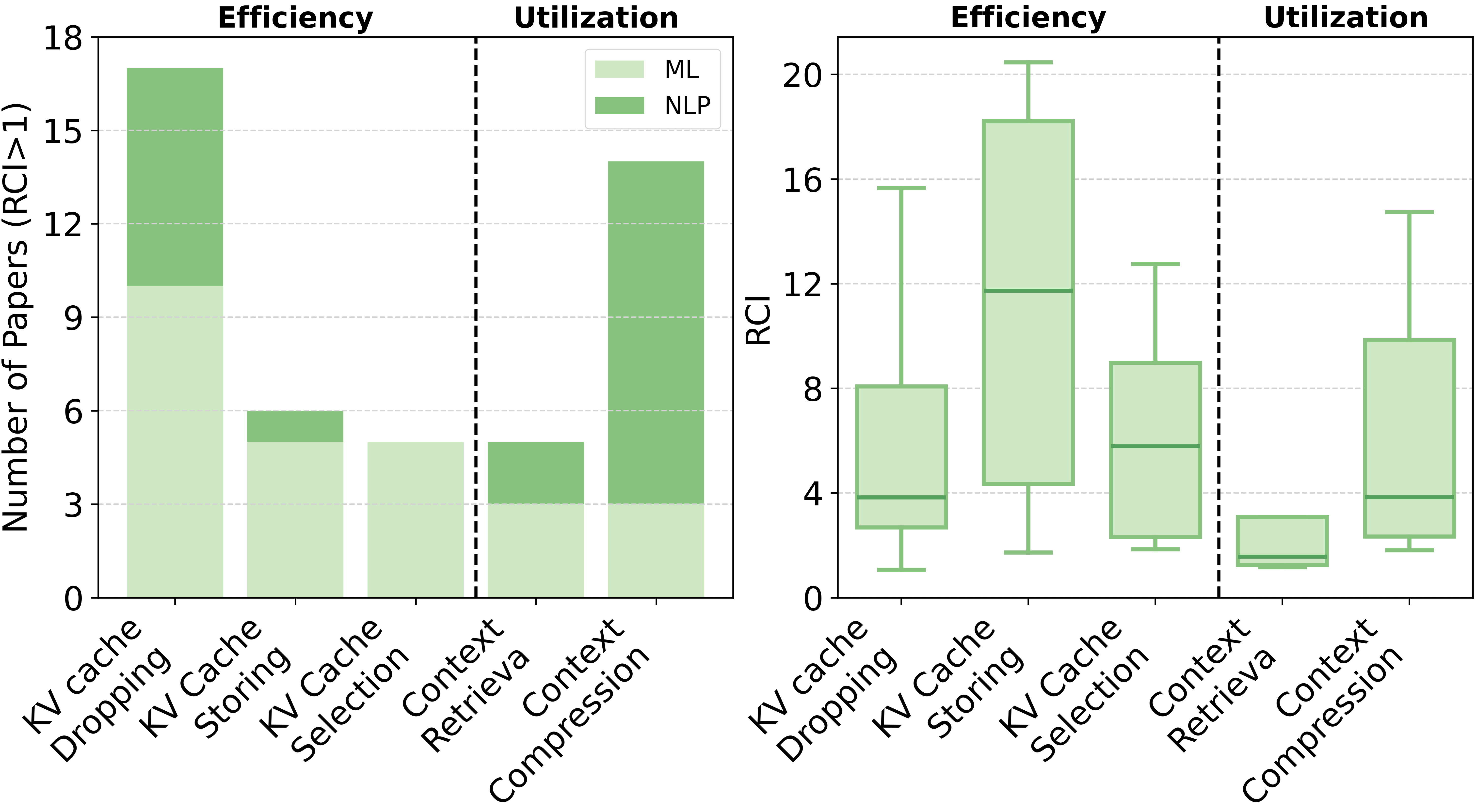}
    \caption{Publication statistic of highlighted papers (RCI > 1) discussed in long-term memory.}
    \label{fig:longcontext-pub}
  \end{minipage}%
  \hfill
  \begin{minipage}[t]{0.48\textwidth}
      \includegraphics[width=\textwidth]{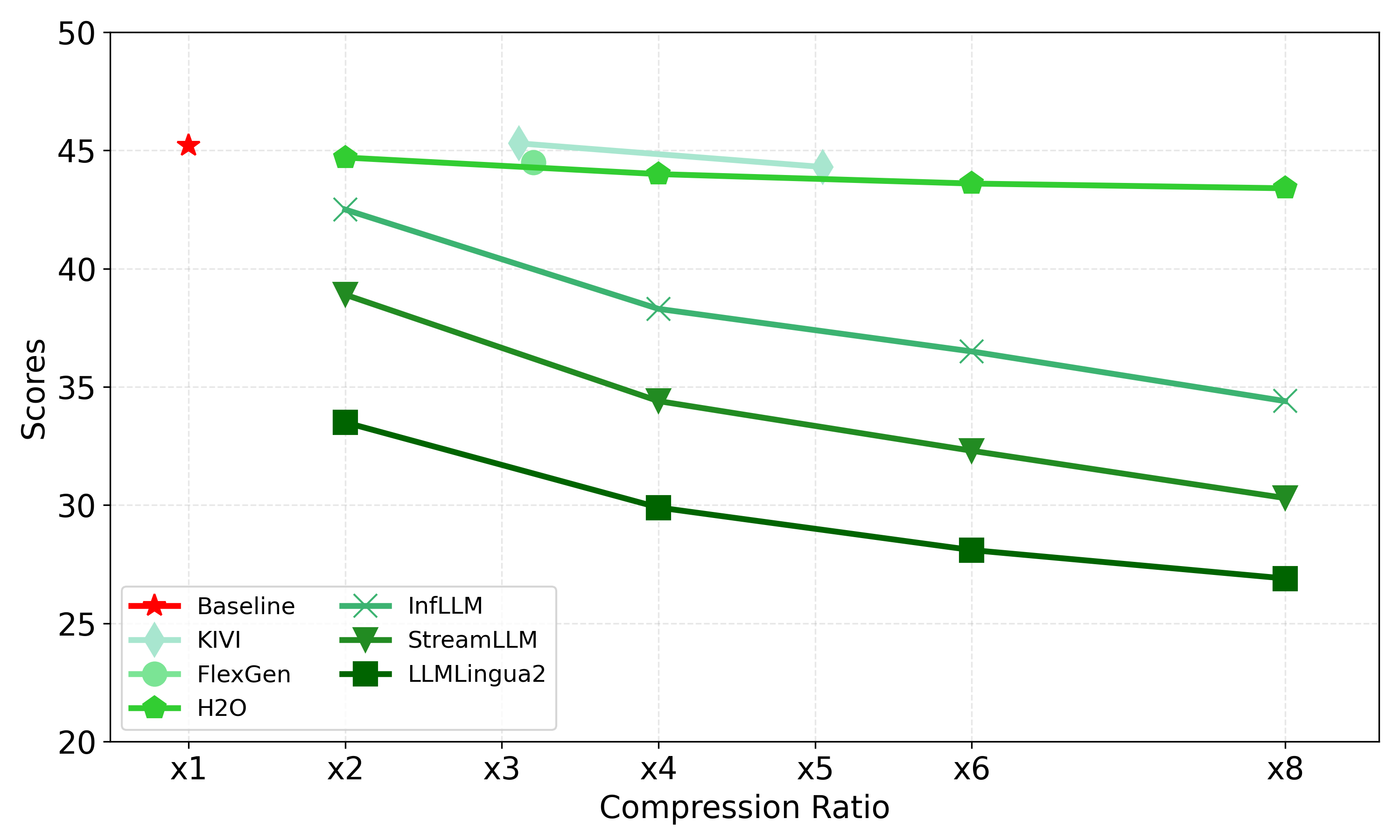}
    \caption{Compression based method performance vs. compression rate on LongBench \citep{bai-etal-2024-longbench}. Data borrowed from \citet{yuan-etal-2024-kv}.}
  \label{fig:long-context-eval}
  \end{minipage}
\end{figure}

\subsubsection{Memory Compression}
To manage extensive amounts of multi-sourced external memory, LLMs must be optimized to efficiently process lengthy contexts. In this section, we discuss approaches for efficiently processing long-context short-term memory, which focuses on memory compression. Specifically, we discuss both memory operations of context tokens (e.g., documents, dialogue histories), and memory operations of KV cache (i.e., working memory). 

\textit{\textbf{Context Compression}} utilizes memory compression operations to optimize contextual memory utilization, which generally involves two major approaches: soft prompt compression and hard prompt compression \citep{li2024prompt}. Soft prompt compression focuses on compressing chunks of input tokens into the continuous vectors in the inference stage (e.g., AutoCompressors \citep{chevalier-etal-2023-adapting}, xRAG \citep{cheng2024xrag}, CEPE \citep{yen-etal-2024-long}), or encoding task-specific long context (e.g., database schema) to parametric memory of finetuned models in the training stage (e.g., YORO \citep{kobayashi-etal-2025-read}), to reduce the input sequence length.


While hard prompt compression directly compresses long input chunks into shorter natural language chunks. Eviction based methods selectively prune uninformative tokens (e.g., Selective Context \citep{li-etal-2023-compressing}, Adaptively Sparse Attention \citep{NEURIPS2023_cdaac2a0}, HOMER \citep{song2024hierarchical}) or chunks (e.g., Semantic Compression \citep{fei-etal-2024-extending}) from the context to shorten the input. Summarization based methods (e.g., RECOMP \citep{xu2024recomp}, CompAct \citep{yoon-etal-2024-compact}, Nano-Capsulator \citep{chuang-etal-2024-learning}, LLMLingua series \citep{jiang-etal-2023-llmlingua, jiang-etal-2024-longllmlingua, pan-etal-2024-llmlingua}) in contrast compress long inputs by abstracting the key information. Hybrid methods (e.g., TCRA-LLM \citep{liu-etal-2023-tcra}) combine the features of evicting uninformative tokens and abstracting context chunks to empower context compression. With both soft prompts and hard prompts, LLMs are allowed to more effectively utilize the context via memory compression.

Beyond static compression, RL-based Active Management has recently emerged, treating context utilization as a dynamic decision-making process. Methods such as AgentFold \citep{ye2025agentfoldlonghorizonwebagents} and FoldGRPO \citep{sun2025scalinglonghorizonllmagent} utilize Reinforcement Learning from Verifiable Rewards (RLVR) to train LLM agents to actively manage and compress long-context information during task execution. Unlike traditional summarization or pruning, these approaches allow the agent to learn an optimal policy for memory retention by optimizing against task-specific rewards. By transitioning from static soft and hard compression to RL-driven active management, LLMs can move beyond simple token reduction toward task-aware memory optimization for long-context memory processing.

\textit{\textbf{KV Cache Eviction.}} In long-context processing, KV cache (working memory) aims to minimize unnecessary key-value computations by storing past key-value pairs as external parametric memory. However, as context length increases, the memory requirement for storing these memory grows quadratically, making it infeasible for handling extremely long contexts. KV Cache Eviction aims to reduce cache size by eliminating unnecessary KV cache. Static eviction approaches select unnecessary cache with fixed pattern. For instance, StreamingLLM \citep{xiao2024efficient} and LM-Infinite \citep{han-etal-2024-lm} use an $\Lambda$-shaped sparse pattern, LCKV \citep{wu-tu-2024-layer} only retain the KV cache from top layer, while LaCache \citep{shi2025lacache} use a ladder-shaped eviction pattern to retain long-range dependency. In contrast, dynamic eviction approaches are more flexible, which decide the KV cache to be eliminated with respect to the query (e.g., H$_2$O \citep{zhang2023ho}, FastGen \citep{ge2024model}, Keyformer \citep{MLSYS2024_48fecef4}, Radar \citep{hao2025radar}, NACL \citep{chen-etal-2024-nacl}), or the model behavior (attention weight) during inference (e.g., SnapKV \citep{NEURIPS2024_28ab4182}, HeadKV \citep{fu2025not}, Scissorhands \citep{liu2023scissorhands}, PyramidInfer \citep{yang-etal-2024-pyramidinfer}, L$_2$ Norm \citep{devoto-etal-2024-simple}, SirLLM \citep{yao-etal-2024-sirllm}, D-LLM \citep{NEURIPS2024_03469b1a}, CateKV \citep{jiang2025catekv}, RocketKV \citep{behnam2025rocketkv}). Considering the risk of potential information loss when discarding KV cache, merging based approaches (e.g., MiniCache \citep{NEURIPS2024_fd070571}, InfiniPot \citep{kim-etal-2024-infinipot}, CHAI \citep{pmlr-v235-agarwal24a}) merge similar KV cache or storing KV cache with special tokens (Activation Beacon \citep{zhang2025long}) instead of directly discarding to reduce information loss. 

\textit{\textbf{KV Cache Storing Optimization.}} In another way of conducting compressing KV cache, KV Cache Storing Optimization considers the potential information loss when removing less important elements, and focus on how to preserve the entire KV cache at a smaller footprint. For instance, LESS \citep{pmlr-v235-dong24f}, Eigen \citep{saxena-etal-2024-eigen} and ShadowKV \citep{sun2025shadowkv} compress KV cache entries into low-rank representations, while FlexGen \citep{10.5555/3618408.3619696}, Atom \citep{ZhaoLZYC0CK0K24}, KVQuant \citep{NEURIPS2024_028fcbcf}, ZipCache \citep{NEURIPS2024_7e57131f}, KIVI \citep{pmlr-v235-liu24bz} dynamically quantize KV cache to reduce memory allocation. More recently, dynamic methods (e.g., Kelle \citep{xia2025kellecodesignkvcaching}) propose software-hardware co-design solution to reduce the cost of storing KV cache. These approaches provide less performance drop compared with KV cache eviction methods but remain limited due to the quadratic nature of the growing memory. Future works should continue focusing on the trade-off between less memory cost and less performance drop.

\subsubsection{Memory Retrieval}
Apart from compressing contextual memory to reduce the load for processing long context, optimizing memory retrieval from long-context raises another important challenge, for effectively identity key information from the noisy context. Considering the type of contextual memory, these efforts can be summarized as contextual retrieval and KV cache selection.

\textit{\textbf{Context Retrieval}} aims to enhance LLM's ability in identifying and locating key information from the contextual memory. Graph-based approaches such as CGSN \citep{nie-etal-2022-capturing} and GraphReader \citep{li-etal-2024-graphreader} decompose documents into graph structures for effective context selection. Token-level context selection approaches (e.g., TRAMS \citep{yu-etal-2023-trams}, Selection-p \citep{chung-etal-2024-selection}, PASTA \citep{zhang2024tell}) pruning and (or) selecting tokens deemed most important. In contrast, methods such as NBCE \citep{su-etal-2024-naive}, FragRel \citep{yue-etal-2024-fragrel}, and Sparse RAG \citep{zhu2025accelerating} perform context selection at the fragment level, choosing the relevant context fragments based on their importance to the specific task. Furthermore, training-based approaches as Ziya-Reader \citep{he-etal-2024-never} and FILM \citep{NEURIPS2024_71c3451f} train LLMs with specialized data to help improve their context selection ability. Other methods like MemGPT \citep{packer2023memgpt}, Neurocache \citep{safaya-yuret-2024-neurocache} and AWESOME \citep{cao-wang-2024-awesome} preserve an external vector memory cache to effectively store and retrieve first encode external memory into vector space, and this external vector memory can be effectively updated or retrieved to enable long-term memory utilization. Together with these methods, LLMs are allowed to better identify key information in the context via memory retrieval. 

\textit{\textbf{KV Cache Selection}} selectively loads essential KV caches to accelerate inference, focusing on efficient memory retrieval. QUEST \citep{pmlr-v235-tang24l}, TokenSelect \citep{wu2025tokenselectefficientlongcontextinference}, and Selective Attention \citep{leviathan2025selective} apply query-aware KV cache selection to identify critical caches for faster inference. Similarly, RetrievalAttention \citep{liu2024retrievalattentionacceleratinglongcontextllm} employs Approximate Nearest Neighbor (ANN) search to locate important caches. By storing KV caches externally and retrieving them during inference, Memorizing Transformers \citep{wu2022memorizing}, LongLLaMA \citep{NEURIPS2023_8511d06d}, ReKV \citep{di2025streaming}, and ArkVale \citep{NEURIPS2024_cd4b4937} efficiently process long contexts. These methods provide flexibility by avoiding KV cache eviction and integrating with storage optimization techniques (e.g., \citet{pmlr-v235-tang24l} shows QUEST is compatible with Atom \citep{ZhaoLZYC0CK0K24}).

\subsubsection{Discussion}
\paragraph{Lost in the Context.}
Despite claims that context length can extend to millions of tokens, long-context LLMs have been found to miss crucial information in the middle of the context during tasks such as question answering and key-value retrieval \citep{liu2024lost, ravaut-etal-2024-context}. This ``lost in the middle'' issue is especially critical when managing vast amounts of external memory, as essential information may be located at various positions within the long context.
Such limitations also extend to the multimodal contexts; as demonstrated in MMLongBench \citep{wang2025mmlongbench}, Long-Context Vision Language Models (LCVLMs) exhibit a similar “lost-in-the-middle” phenomenon when processing lengthy interleaved text-image documents.
In addition, in more complex scenarios requiring reasoning based on contextual memory, LLMs also fail to effectively aggregate memory across different part of the context \citep{huang2025maskingmultihopqaanalysis}. Furthermore, though higher recall can be obtained with larger retrieval set, irrelevant information will mislead LLMs and harm the generation quality \citep{pmlr-v202-shi23a, jin2025longcontext}. Effective contextual utilization become a key challenge in addressing these limitations, encompassing context retrieval and context compression across memory operations.

\paragraph{Trade-off between compression rate and performance drop.} Compression, as one of the major memory operations involved in long context memory, is widely used in compressing both parametric memory (KV cache) and contextual memory (Context), to balance the efficiency (compression rate) and effectiveness (performance drop). Different compression-based strategies have their own pros and cons. For example, KV cache eviction methods typically achieve higher compression rates but result in greater information loss and, consequently, a more significant performance drop.  \citet{yuan-etal-2024-kv} propose an universal benchmarking on these different strategies, qualitatively showcase the pros and cons according to different strategies. As illustrated in Figure~\ref{fig:long-context-eval}, generally, KV cache storage optimization methods (with 'x' marker) achieves best trade-off between effectiveness and efficiency. In contrast, KV cache eviction methods (with $\nabla$ marker) are more flexible, with fully customization compression rate, but less effective. In the other hand, compressing the contextual memory (with $\Delta$ marker) are less effective compared with compressing the parametric memory, as evidenced by the comparatively poor performance of LLMLingua2.

\paragraph{Publication Trending.} Figure~\ref{fig:longcontext-pub} summarizes publication trends on long context. The NLP community focuses more on utilization with contextual memory, while the ML community dedicates more effort to efficiency via parametric memory. From an RCI perspective, KV cache storage optimization dominates discussions on long context topics. This dominance is not only for balancing efficiency and effectiveness, but also due to its compatibility with other long context methods. Comparing the two memory operation, retrieval methods generally get less attention. One reason for this is the overlap between context retrieval and other topics, such as long-term memory and multi-source memory, which leads to context retrieval being somewhat underestimated in Figure~\ref{fig:longcontext-pub}. Additionally, understanding the relationship between RAG and long-context \citep{li-etal-2024-retrieval, jin2025longcontext} is crucial for the development of memory-based LLM agent. However, impactful work on contextual utilization in complex environments is still lacking. Addressing this gap is a valuable future direction.

\begin{tcolorbox}[
    colback=longcontextcolor!30!white,
    colframe=black!50,
    arc=6pt,
    boxrule=0.8pt,
    width=\linewidth,
    enhanced,
    left=4pt,
    right=4pt,
    top=4pt,
    bottom=4pt,
    fontupper=\small\bfseries
]
\lightbulb \textbf{Balancing the trade-off between reduced memory usage and minimized performance degradation in KV cache optimization represents an exciting area for future research.} 

\lightbulb \textbf{Contextual utilization with complex environment (e.g., multi-source memory) is a pivotal research direction for advancing the development of intelligent agents.}

\end{tcolorbox}

\subsection{Parametric Memory Modification}
Modifying parametric memory, which is encoded knowledge within the LLM parameters, is crucial for dynamically adapting stored memory. Methods for parametric memory modification can be broadly categorized into three types: (1) \textbf{Editing} is the localized modification of model parameters without requiring full model retraining; (2) \textbf{Unlearning} selectively removes unwanted or sensitive information; and (3) \textbf{Continual Learning} incrementally incorporates new knowledge while mitigating catastrophic forgetting. This section systematically reviews recent research in these categories, with detailed analyses and comparisons presented in subsequent subsections.


\subsubsection{Editing.}
Parametric memory editing updates specific knowledge stored in the parametric memory without full retraining.
One prominent line of work involves directly modifying model weights.
A dominant strategy is locating-then-editing method~\cite{meng2022locating,meng2023massediting,mela2024mass,huang2024commonsense,fang2025alphaedit,guo2025towards,deng2025everything}, which uses attribution or tracing to find where facts are stored, then modifies the identified memory directly.
Another approach is meta-learning~\cite{de2021editing,mitchell2022fast,tan2024massive,li2024pmet,zhang2024dafnet,guo2025towards}, where an editor network learns to predict targeted weight changes for quick and robust corrections.
Some methods avoid altering the original weights altogether. 
Prompt-based methods~\cite{zheng2023can,zhong2023mquake} use crafted prompts like ICL to steer outputs indirectly.
Additional-parameter methods~\cite{wang2024wise,dong2022calibrating,mitchell2022memory,wang2024memoryllm,Larimar2024} add external parametric memory modules to adjust behavior without touching model weights.
These approaches vary in efficiency and scalability, though most focus on entity-level edits.

\subsubsection{Unlearning.}
Parametric memory unlearning enables selective forgetting by removing specific memory while retaining unrelated memory.
Recent work explores several strategies.
Additional-parameter methods add components such as logit difference modules~\cite{ji2024reversing} or unlearning layers~\cite{chen2023unlearn} to adjust memory without retraining the whole model. 
Prompt-based methods manipulate inputs~\cite{liu2024large} or use ICL~\cite{pawelczyk2024context} to externally trigger forgetting.
Locating-then-unlearning methods~\cite{jia2024wagle, tian2024forget, wu2023depn} first identify responsible parametric memory, then apply targeted updates or deactivations.
Training objective-based methods~\cite{wang2025llm, liu2024towards,jia2024soul, yao2024large} modify the training loss functions or optimization strategies explicitly to encourage memory forgetting.
These approaches aim to erase memory when given explicit forgetting targets, while preserving non-targeted knowledge and balancing efficiency and precision.

\subsubsection{Continual Learning.}
Continual learning \citep{wang2024comprehensive} enables long-term memory persistence by mitigating catastrophic forgetting in model parameters. Two main approaches are regularization-based and replay-based methods. Regularization constrains updates to important weights, preserving vital parametric memory; methods like TaSL \citep{feng2024tasl}, SELF-PARAM \citep{wangself}, EWC \citep{kirkpatrick2017overcoming}, and POCL \citep{wu2024mitigating} apply such constraints to embed knowledge without replay. In contrast, replay-based methods reinforce memory by reintroducing past samples, particularly suited to incorporating retrieved external knowledge or historical experiences during training. For example, DSI++ \citep{mehta2022dsi++} leverages generative memory to supplement learning with pseudo queries, maintaining retrieval performance without full retraining. Beyond these paradigms, agent-based work such as LifeSpan Cognitive System (LSCS) \citep{wang2024towards} extends continual learning into an interactive setting, enabling agents to incrementally acquire and consolidate memory through real-time experience. LSCS provides valuable insights into how external memory can be encoded into model parameters continually. 

\subsubsection{Discussion}

\begin{figure}[t!]
  \centering
  \begin{minipage}[t]{0.48\textwidth}
    \includegraphics[width=\linewidth]{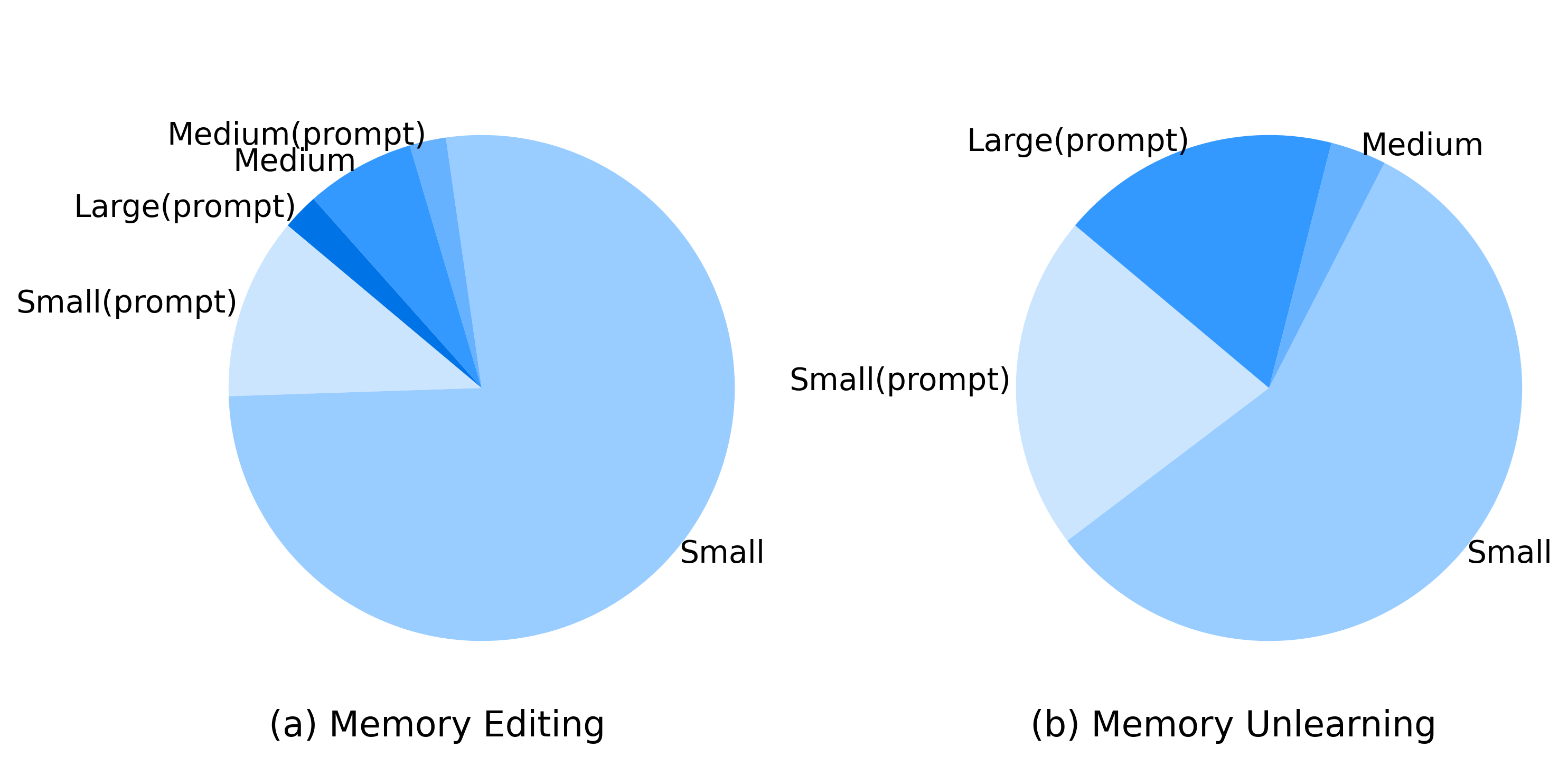}
    \caption{Model size distribution in memory editing and unlearning. }
    \label{fig:para-modelsize}
  \end{minipage}%
  \hfill
  \begin{minipage}[t]{0.48\textwidth}
        \includegraphics[width=\linewidth]{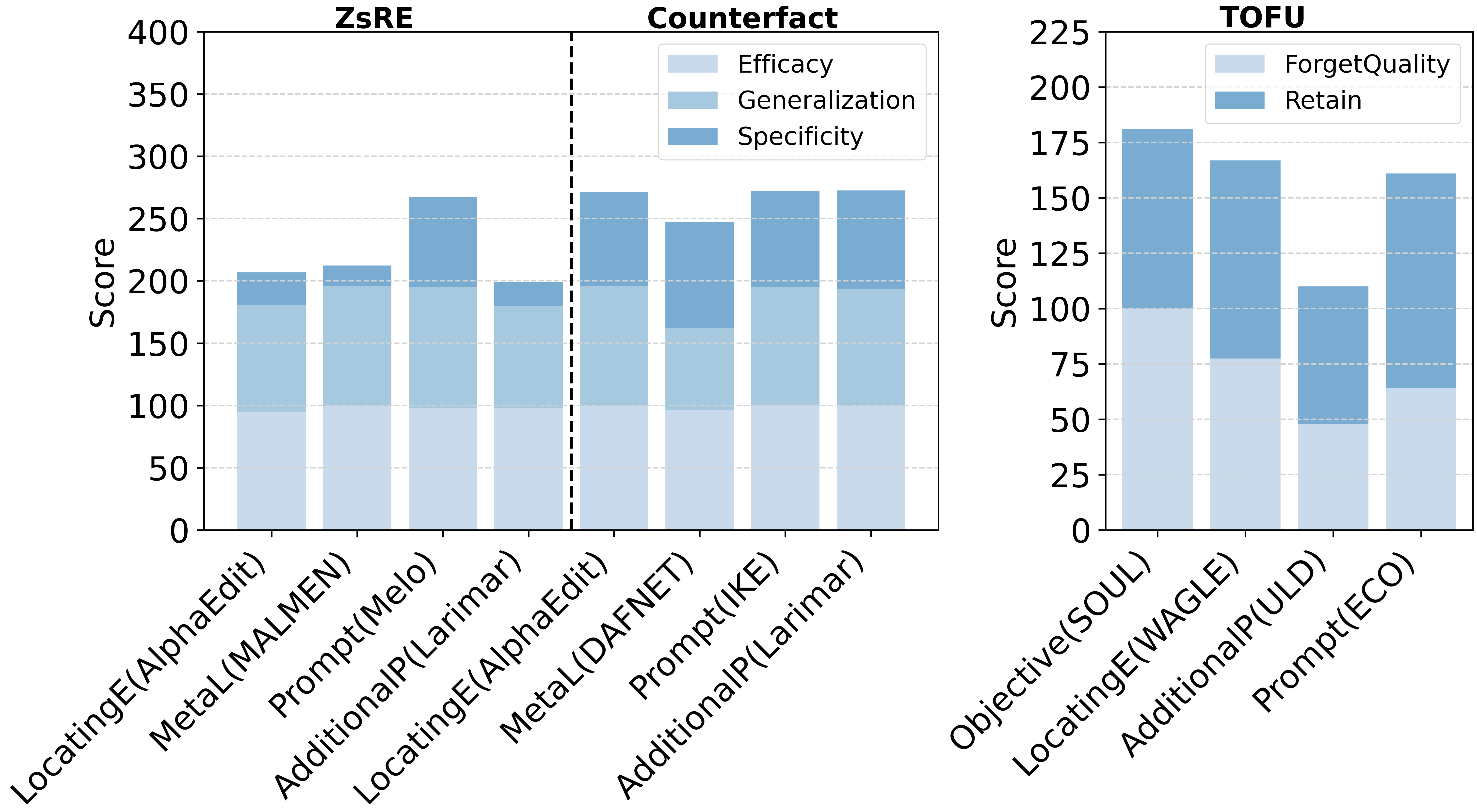}
    \caption{SOTA solutions across different categories on the CounterFact (editing), ZsRE (editing) and TOFU (unlearning) benchmark. }
    \label{fig:para-sota}
  \end{minipage}
\end{figure}

\paragraph{SOTA Solution Analysis.}
We select recent SOTA methods across different categories and report their performance in Figure~\ref{fig:para-sota} on the most widely used datasets for memory editing (CounterFact~\cite{meng2022locating} and ZsRE~\cite{levy2017zero}) and memory unlearning (ToFU~\cite{maini2024tofu}). We aim to ensure a fair comparison by using consistent base models and appropriate evaluation metrics. Specifically, for CounterFact and ZsRE, we follow~\citet{meng2022locating}, where 2,000 samples are randomly selected from the dataset for updates, with 100 samples per edit. All methods on CounterFact use GPT-J as the base model; for ZsRE, most use GPT-2, except MELO, which uses T5-small. For the ToFU benchmark, all methods use LLaMA2-7B-chat under the 10\% forgetting setting.
Prompt-based methods achieve strong overall performance across all benchmarks, while meta-learning methods generally underperform compared to others.
We observe that the same methods tend to perform worse on ZsRE than on CounterFact. This drop is primarily due to significantly lower specificity scores on ZsRE, which in turn lowers the overall score.
This highlights the challenge of achieving precise, targeted edits and suggests that improving specificity remains a promising research direction.
Additionally, we find that most current SOTA methods achieve high scores on the ToFU benchmark, suggesting it may be insufficiently challenging and that new unlearning benchmarks are needed.

\begin{figure}[t!]
  \centering
  \begin{minipage}[t]{0.48\textwidth}
    \includegraphics[width=\linewidth]{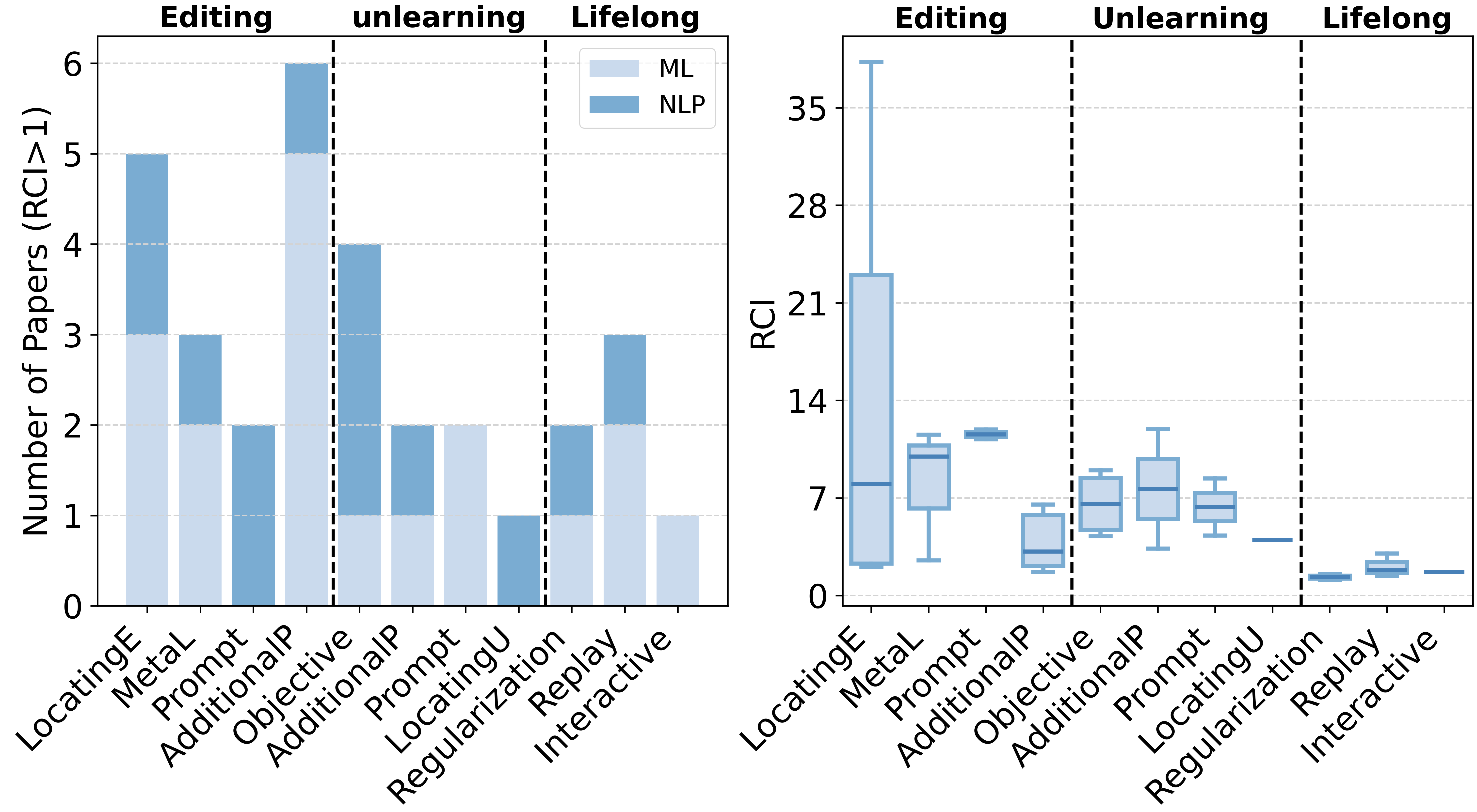}
    \caption{Publication statistic of highlighted papers (RCI > 1) discussed in this section.}
    \label{fig:para-publication}
  \end{minipage}%
  \hfill
  \begin{minipage}[t]{0.48\textwidth}
          \includegraphics[width=\linewidth]{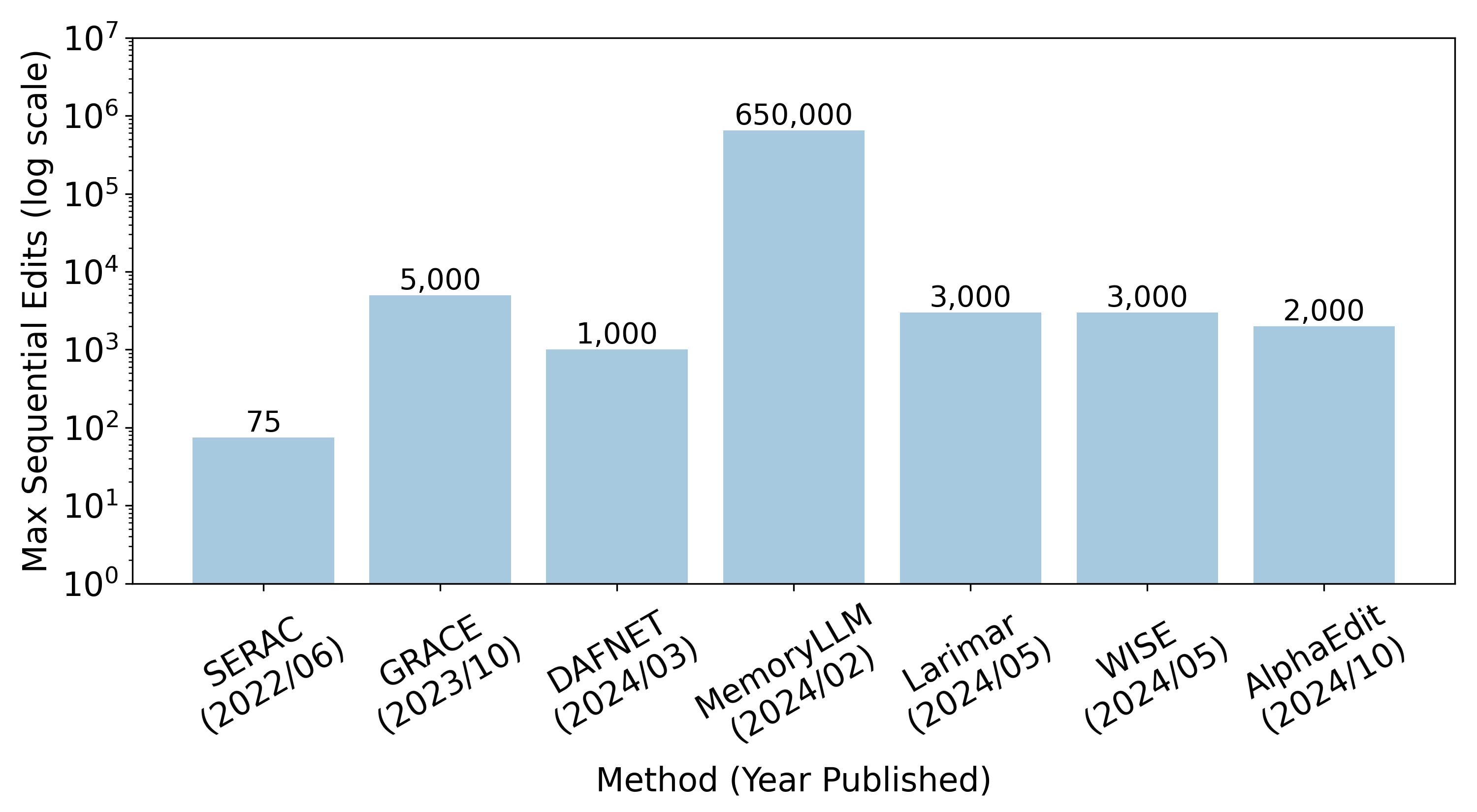}
    \caption{Maximum sequential edits supported by different model editing methods}
    \label{fig:para-sequenceeditng}
  \end{minipage}
\end{figure}

\paragraph{Scaling Challenges.}

Figure~\ref{fig:para-sequenceeditng} shows the maximum number of sequential edits supported by different methods. Except for MemoryLLM, which supports up to 650k updates, most methods only test 1,000 to 5,000 edits. We also note that research on sequential unlearning remains sparse and presents an open area for future exploration.
Figure~\ref{fig:para-modelsize} illustrates the distribution of model sizes used across different methods. In both editing and unlearning, non-prompt-based methods are typically applied to medium or small models ($\leq 20\text{B}$). In contrast, prompt-based approaches are more commonly evaluated on larger models, likely due to their reliance on stronger instruction-following and in-context learning capabilities. Non-prompt methods, on the other hand, often face scalability challenges due to higher computational costs, making them difficult to apply to large models. This highlights the need to further investigate how to balance model size with editing or unlearning effectiveness and efficiency.

\paragraph{Publication Trending.}
Figure~\ref{fig:para-publication} presents publication statistics of papers with RCI > 1 across editing, unlearning, and lifelong learning. Editing has attracted the most attention, especially locating-then-editing and additional-parameter methods. NLP venues focus more on editing, while ML work is more evenly distributed across the three areas. Locating-then-editing also shows the highest RCI variance, reflecting several highly influential studies. Although unlearning is less represented, it demonstrates strong potential in objective- and parameter-based categories. Lifelong learning, by contrast, remains relatively underexplored.

\begin{tcolorbox}[
    colback=parametriccolor!30!white,
    colframe=black!50,
    arc=6pt,
    boxrule=0.8pt,
    width=\linewidth,
    enhanced,
    left=4pt,
    right=4pt,
    top=4pt,
    bottom=4pt,
    fontupper=\small\bfseries
]
\lightbulb \textbf{Current editing methods often lack specificity, while unlearning benchmarks like TOFU may be too simple to reveal real limitations.}

\lightbulb \textbf{Agents should leverage continual learning to self-evolve through sustained interaction with the environment, without overwriting stable parametric memory.}

\end{tcolorbox}
\subsection{Multi-source Memory}
Multi-source memory is essential for real-world AI deployment, where systems must reason over internal parameters and external knowledge bases spanning structured data (e.g., knowledge graphs, tables) and unstructured multi-modal content (e.g., text, audio, images, videos). This section examines key challenges across two dimensions: cross-textual integration and multi-modal coordination. 

\subsubsection{Cross-textual Integration}
Cross-textual integration enables an AI agent to perform deeper reasoning and resolve conflicts from multiple textual sources to support more contextually grounded responses.

\textit{\textbf{Reasoning}} focuses on integrating multi-format memory to generate factually and semantically consistent responses. One line of research investigates reasoning over memories from different domains, particularly through the precise manipulation of structured symbolic memories, as demonstrated by ChatDB \citep{hu2023chatdb} and Neurosymbolic \citep{wang2024symbolic}. Other works \citep{nogueira-dos-santos-etal-2024-memory, wu-etal-2022-efficient} explore the dynamic integration of domain-specific parameterized memories to enable more flexible reasoning. Multi-source reasoning across diverse document sources has also been studied, as seen in DelTA \citep{wang2025delta} and dynamic-MT \citep{du-etal-2022-dynamic}. Additionally, several studies \citep{li2024structrag, DBLP:conf/acl/LeePF024, zhao2024divknowqa, xu2024generate} have investigated heterogeneous knowledge integration by retrieving information from both structured and unstructured sources. While these efforts have made substantial progress in combining parameterized and external memories for reasoning, achieving unified reasoning over heterogeneous, multi-source memories remains a major open challenge. In particular, more work is needed to effectively integrate parameterized memories with both structured and unstructured external knowledge sources.

\textit{\textbf{Conflict}} in multi-source memory refers to factual or semantic inconsistencies that arise during the retrieval and reasoning over heterogeneous memory representations. These conflicts often emerge when integrating parametric and contextual memories, or combining structured and unstructured knowledge such as triples, tables, and free text \citep{xu2024knowledge}. Prior work has focused on identifying and localizing such inconsistencies. For example, RKC-LLM \citep{wang2023resolving} proposes an evaluation framework to assess models' ability to detect contextual contradictions, while BGC-KC \citep{tan-etal-2024-blinded} highlights models' tendency to favor internal knowledge over retrieved content, motivating source attribution and trust calibration. These methods offer important foundations for memory conflict understanding, though many remain limited to static scenarios or single-source reasoning.

\begin{figure}[t!]
  \centering
  \begin{minipage}[t]{0.48\textwidth}
    \includegraphics[width=\linewidth]{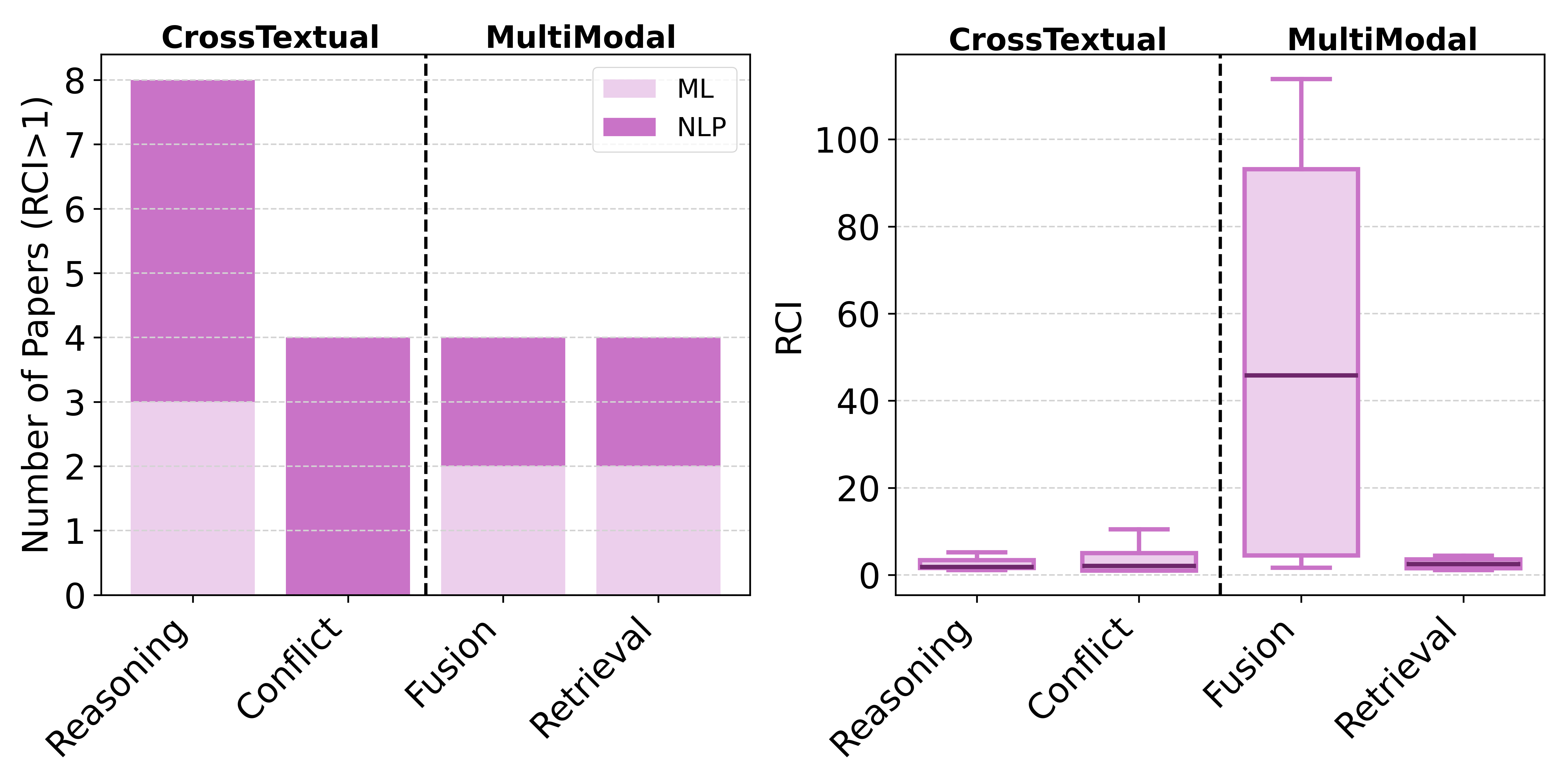}
    \caption{Publication statistic of highlighted papers (RCI > 1) discussed in multi-source memory.}
    \label{fig:multisource-pub}
  \end{minipage}%
  \hfill
  \begin{minipage}[t]{0.48\textwidth}
      \includegraphics[width=\textwidth]{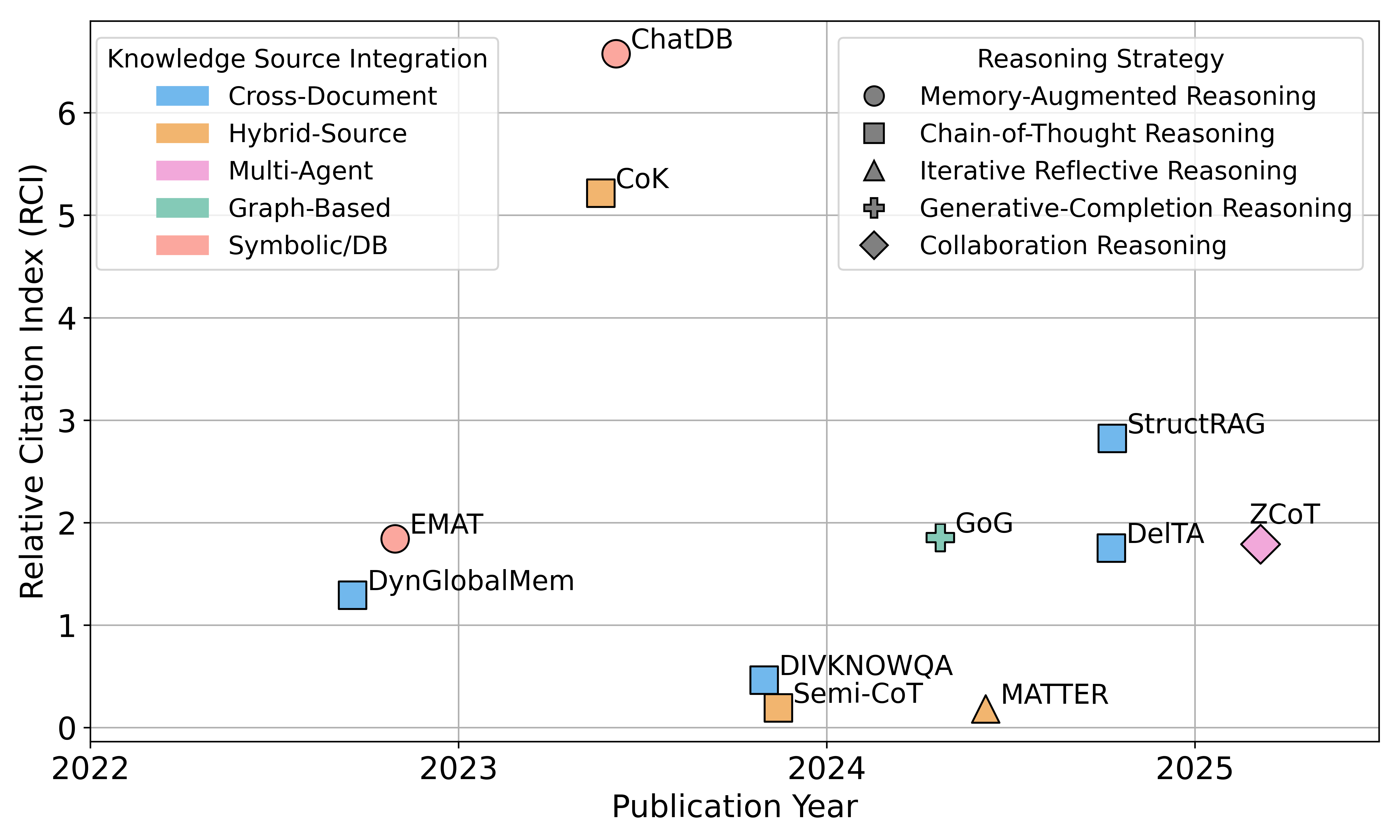}  
      \caption{Trends in cross-textual reasoning: memory sources and reasoning strategies.}
      \label{fig:cross-textual_memory_reasoning}
  \end{minipage}
\end{figure}

\begin{figure}[t!]
  \centering
  \begin{minipage}[t]{0.35\textwidth}
      \includegraphics[width=\textwidth]{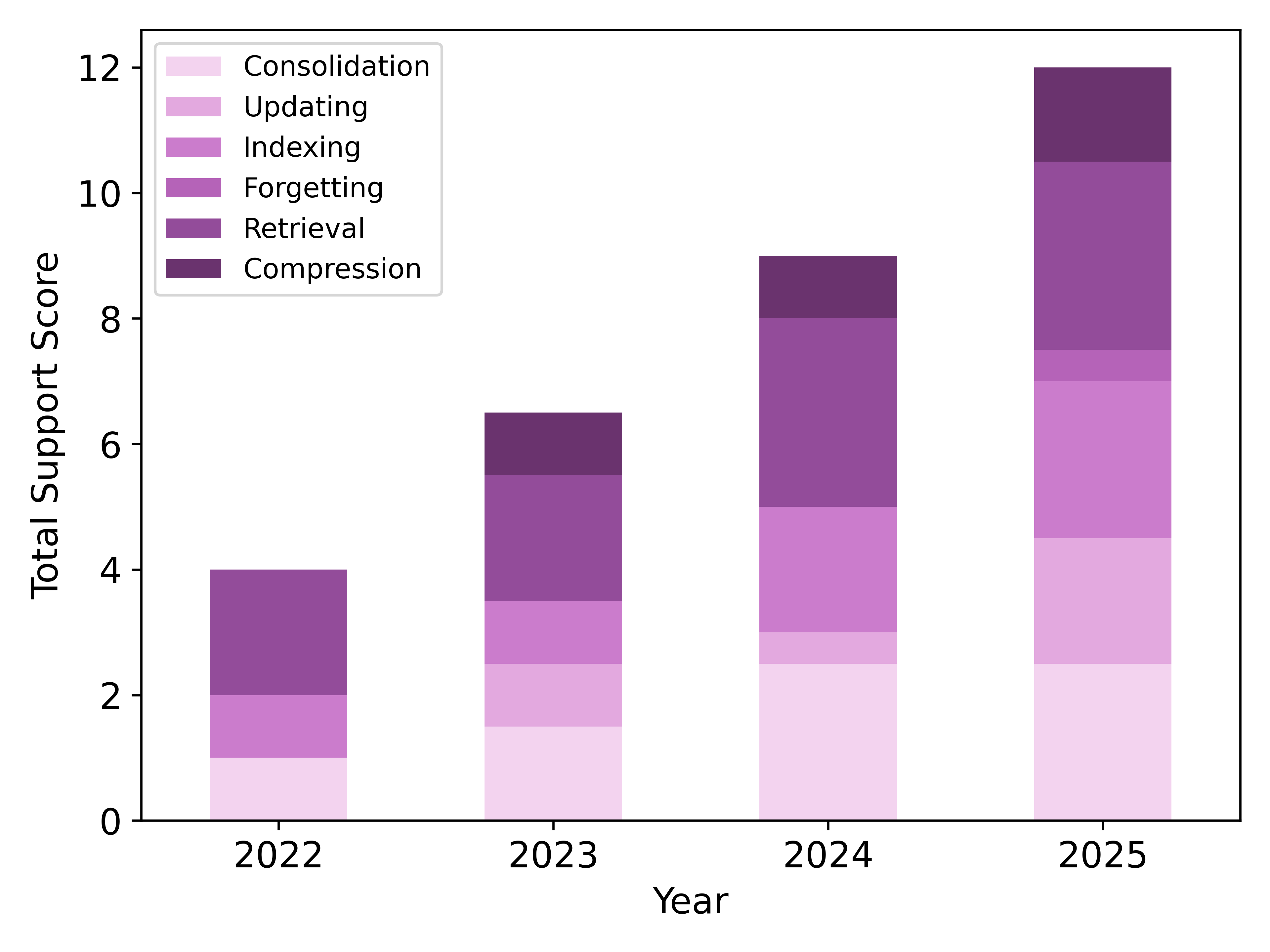}  
      \caption{Evolution of memory operation support across Years.}
      \label{fig:memory_operation_mm_trend}
  \end{minipage}%
  \hfill
  \begin{minipage}[t]{0.6\textwidth}
      \includegraphics[width=\textwidth]{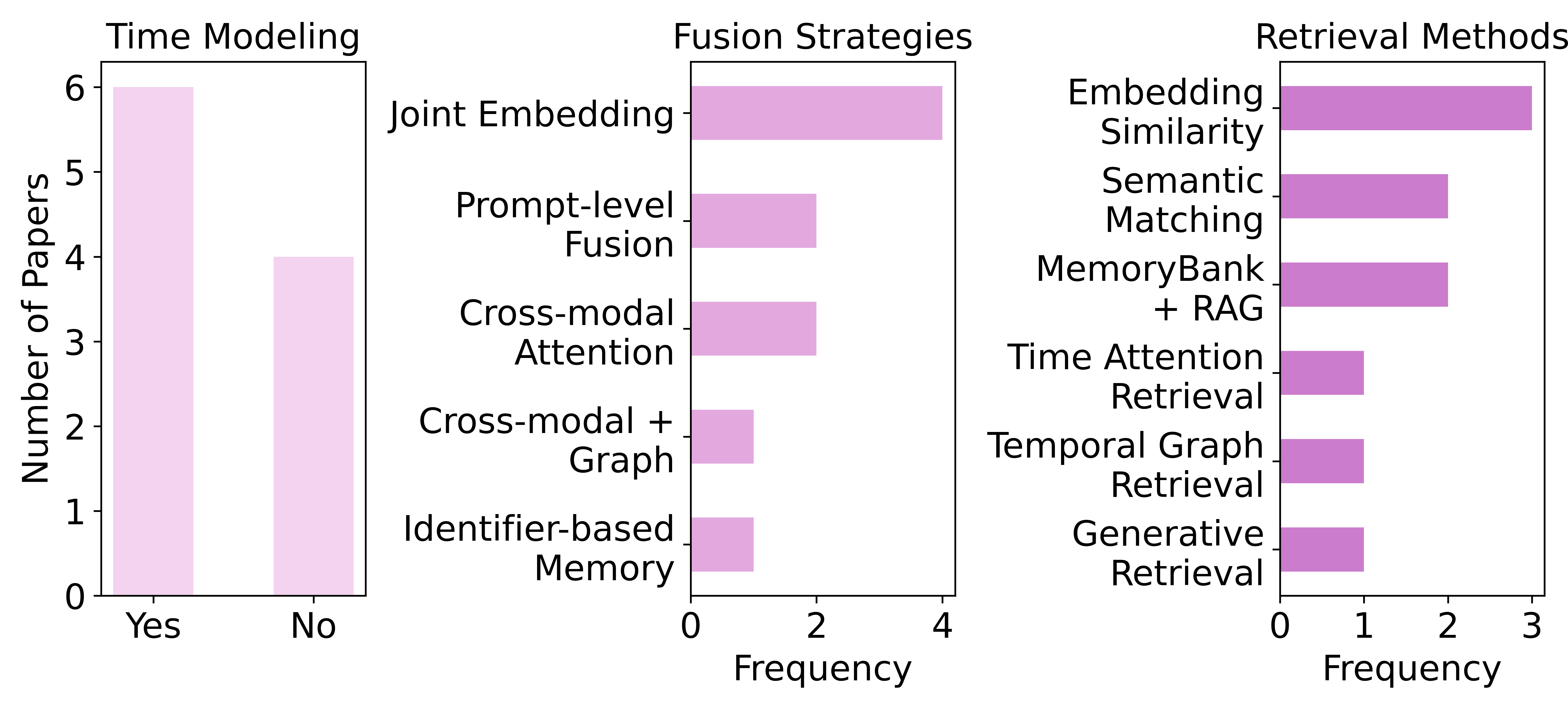}  
      \caption{Analysis of temporal modeling, fusion strategies, and retrieval methods in multi-modal coordination.}
      \label{fig:multi_modal_analysis}
  \end{minipage}
\end{figure}

\subsubsection{Multi-Modal Coordination.}
As memory-augmented systems evolve toward multi-modal settings, a key challenge lies in fusion and retrieval over heterogeneous modalities such as text, image, audio, and video.

\textit{\textbf{Fusion}} refers to aligning the retrieved information across diverse modalities. From a memory perspective, fusion serves as a key mechanism for integrating cross-modal information over time. Existing approaches can be broadly divided into two lines. The first focuses on \textbf{unified semantic projection}, where models such as UniTransSeR \citep{ma-etal-2022-unitranser}, MultiInstruct \citep{xu-etal-2023-multiinstruct}, PaLM-E \citep{driess2023palme}, and NExT-Chat \citep{zhang2023nextchat} embed heterogeneous inputs into a shared representation space for reuse and query.
The second line emphasizes long-term cross-modal memory integration. For example, LifelongMemory \citep{wang2023lifelongmemory} introduces a transformer with persistent memory to accumulate visual-textual knowledge across patient records. Similarly, MA-LMM \citep{he2024ma} maintains a multimodal memory bank to extend temporal understanding in long videos. While effective at aligning modalities, current fusion methods often fall short in supporting long-term multimodal memory management. Key challenges include dynamic memory updates and maintaining consistency across heterogeneous sources.

\textit{\textbf{Retrieval}} in multi-modal systems enables access to stored knowledge across modalities such as text, image, and video. Most existing methods rely on embedding-based similarity computation, grounded in vision-language models like QwenVL \citep{bai2023qwenvl}, CLIP \citep{pmlr-v139-radford21a} or other multi-modal models \citep{li-etal-2024-generative}. These models project heterogeneous inputs into a shared semantic space, allowing for cross-modal retrieval. For instance, VISTA \citep{zhou2024vista} enhances retrieval via visual token representations, while UniVL-DR \citep{liu2023univldr} integrates video and language through a unified dual encoder. More recently, IGSR \citep{wang2025new} extends retrieval to multi-session conversations by introducing intent-aware sticker retrieval, yet it still remains anchored in similarity-based retrieval. The limitations of such approaches are underscored by MMLongBench \citep{wang2025mmlongbench}, which reveals that even state-of-the-art Large Vision-Language Models (LVLMs) struggle with cross-modality retrieval. Consequently, these methods often lack the capacity for reasoning-driven retrieval and neglect critical modalities like audio and sensorimotor signals required for embodied interaction. To bridge these gaps, M3 \citep{long2025seeinglisteningrememberingreasoning} introduces a Multi-modal Memory Modelling framework for open-ended agents that unifies storage and reasoning across diverse data types, including audio and sensorimotor signals. By enabling dynamic updates and reasoning-driven retrieval, M3 moves beyond shallow alignment to ensure robust long-term memory management.

\subsubsection{Discussion}

\paragraph{Trends in Multi-Source Memory Integration.}
Recent studies \citep{wang2025new, song2024moviechat} reveal a steady evolution in how multi-source memory is organized, retrieved, and reasoned over. While diverse methods have been proposed for \textbf{cross-textual integration} and \textbf{multi-modal coordination}, a closer look at representative models (Figures~\ref{fig:cross-textual_memory_reasoning}, \ref{fig:memory_operation_mm_trend}, \ref{fig:multi_modal_analysis}) highlights shared challenges and emerging trends. These developments reflect a broader shift from static retrieval pipelines toward dynamic, context-sensitive memory systems capable of supporting temporally grounded, cross-source reasoning across tasks and sessions.

\textbf{Cross-textual integration} involves two key design axes: source type and reasoning mechanism. Early models such as ChatDB \citep{hu2023chatdb} and EMAT \citep{wu-etal-2022-efficient} use symbolic memory (e.g., databases, tables) accessed via explicit queries, offering transparency but limited scalability in open-domain settings. More recent systems like StructRAG \citep{li2024structrag}, DelTA \citep{wang2025delta}, and Chain-of-Knowledge \citep{li2024chain} adopt unstructured memory and neural retrieval, combining attention-based fusion with chain-of-thought reasoning. Yet, most still treat memory as static, disconnected from real-time inference. Newer models such as MATTER \citep{DBLP:conf/acl/LeePF024}, GoG \citep{xu2024generate}, and ZCoT \citep{michelman2025enhancing} move toward inference-aware memory, using retrieval-generation loops and collaborative agents to evolve memory dynamically.  Despite this shift, resolving conflicts across heterogeneous sources remains a major challenge. Retrieved and parametric content are often merged without consistency checks or source attribution, leading to hallucinations and factual drift \citep{tan-etal-2024-blinded, zhou-etal-2023-context}. Preliminary solutions such as multi-step conflict resolution \citep{wang2023resolving} and epistemic calibration \citep{xu2024knowledge} are promising but lack scalability. Future work should pursue integrated, conflict-aware memory systems capable of dynamic reasoning under uncertainty and source ambiguity.

\textbf{Multi-modal memory coordination} has advanced across three key dimensions: fusion, retrieval, and temporal modeling. As shown in Figure~\ref{fig:multi_modal_analysis}, common strategies include joint embedding \citep{he2024ma, zhou2024vista, ma-etal-2022-unitranser, wang2025new, wang2025mmlongbench} and prompt-level fusion \citep{wang2023lifelongmemory, guo2024embodied}, while recent methods such as identifier-based memory \citep{li-etal-2024-generative} and cross-modal graph fusion \citep{nguyen-etal-2023-conversation} enable more selective, task-adaptive integration. Retrieval has evolved from static similarity toward temporally contextualized approaches, including temporal graphs and time-aware attention \citep{xiao2025worldmem}, facilitating reasoning over extended interactions. Notably, 60\% of surveyed models encode temporal information, underscoring the importance of time in long-horizon tasks. Beyond retrieval and fusion, operational control—such as memory updating, indexing, and compression—is becoming increasingly essential. While earlier systems (2022–2023) mainly focused on retrieval, newer agents like E-Agent \citep{glocker2025llm} and WorldMem \citep{xiao2025worldmem} adopt self-maintaining architectures that continuously refine memory content over time. For example, WorldMem compresses multi-modal logs, while E-Agent dynamically updates internal memory to support long-horizon planning. These systems highlight a shift from passive memory querying to active, operationally rich architectures.

\paragraph{Publication Trend.} As shown in Figure~\ref{fig:multisource-pub}, cross-textual reasoning dominates publication volume, reflecting its foundational role in multi-source integration. Fusion research, particularly work driven by CLIP \citep{pmlr-v139-radford21a}, demonstrates the highest citation impact and influence on multi-modal learning. In contrast, dynamic retrieval and conflict resolution remain underexplored. Together, these trends suggest a field transitioning from surface-level integration toward deeper, operation-aware, and temporally structured memory architectures.

\definecolor{efcfef}{HTML}{EFCFEF}
\begin{tcolorbox}[
    colback=efcfef!40!white,
    colframe=black!50,
    arc=6pt,
    boxrule=0.8pt,
    width=\linewidth,
    enhanced,
    left=4pt,
    right=4pt,
    top=4pt,
    bottom=4pt,
    fontupper=\small\bfseries
]
\lightbulb \textbf{Design conflict-aware memory mechanisms that explicitly detect, attribute, and resolve inconsistencies across evolving memories and heterogeneous representations.}

\lightbulb \textbf{Develop self-maintaining memory architectures with built-in indexing, updating, compression, and consistency checks for long-term, cross-session use.}

\lightbulb \textbf{Advance long-horizon reasoning by integrating multi-modal long-context understanding with multi-turn dialogue reasoning, a core requirement for real-world agents.}
\end{tcolorbox}

\section{Memory In Practice}
\label{sec:memory_practice}

Memory augmentation agents operationalize theoretical memory concepts through an interdependent hierarchy of products, development tools, and infrastructure. Products such as assistants and copilots utilize parametric and contextual memory to support personalization and long-horizon reasoning. Development tools translate practical demands into frameworks that manage storage, retrieval, and adaptation. Infrastructure provides the computational backbone that supports memory operations at scale. The interaction among these layers is bidirectional: product requirements drive development tool design, tools constrain infrastructural implementation, and infrastructural advances enable richer product capabilities. Understanding and bridging the gaps among them clarifies both the technical and conceptual frontiers of agent memory mechanisms.

\begin{table}[t]
\centering
\small
\setlength{\tabcolsep}{2pt}
\caption{Product Memory Design Trade-offs. Representative products are compared to highlight recurring design choices and limitations of memory systems.}
\begin{tabular}{@{}>{\centering\arraybackslash}p{1.3cm}
                >{\centering\arraybackslash}p{1.6cm}
                >{\centering\arraybackslash}p{1.5cm}
                >{\centering\arraybackslash}p{2cm}
                >{\centering\arraybackslash}p{3cm}
                p{5cm}@{}}
\toprule
\makecell[tc]{\textbf{Products}} &
\makecell[tc]{\textbf{Domain}} &
\makecell[tc]{\textbf{Dominant}\\\textbf{Memory}} &
\makecell[tc]{\textbf{Prioritized}\\\textbf{Operations}} &
\makecell[tc]{\textbf{User}\\\textbf{Experience}} &
\makecell[tl]{\textbf{Limitations}} \\
\midrule
\textbf{ChatGPT} \newline \citep{openai2023chatgpt} &
General &
Parametric &
\makecell[tc]{Consolidation\\Retrieval\\Condensation} &
Consistency / Accuracy / General &
Hallucination; limited personalization and modal memory management. \\
\midrule
\textbf{Replika} \newline \citep{replika2025} &
Personal &
Contextual &
\makecell[tc]{Updating\\Retrieval} &
Empathy / Adaptation &
Privacy risks; memory drift; limited cross-session continuity; simple slot-based memory management. \\
\midrule
\textbf{GitHub Copilot} \newline \citep{githubcopilot} &
Task-oriented &
Parametric &
\makecell[tc]{Condensation} &
Efficiency / Reliability &
No cross-session task continuity; no user-level personalization; no persistent long-term memory. \\
\midrule
\textbf{Doubao} \newline \citep{doubao2025} &
Multi-modal &
Parametric &
\makecell[tc]{Consolidation\\Retrieval\\Condensation} &
General / Stylization / Low latency &
Hallucination; modality gap; session-bound. \\
\bottomrule
\end{tabular}
\label{tab:agent_comparison_dim7}
\end{table}

\subsection{Products}
The agent products can be broadly categorized based on their dominant memory types and application focus. \textbf{General agents} like ChatGPT \citep{openai2023chatgpt}, Gemini \citep{gemini2024}, Claude \citep{claude2023}, Grok \citep{grok2023}, and DeepSeek \citep{liu2024deepseek} rely predominantly on large-scale parametric memory to encode broad cross-domain knowledge within model weights and underpin stable reasoning and factual generalization. Limited user contextual memory is layered on top to improve retrieval and situational grounding. \textbf{Personal agents} primarily leverage contextual memory to capture user preferences, interaction history, and affective cues, enabling personalized and adaptive responses \citep{li2024hello, qin2025ui, hong2023cogagent} such as \textbf{Replika} \citep{replika2025}, \textbf{Character.AI \citep{characterai2023}},  \textbf{Me.bot}, \textbf{Tencent ima.copilot} \citep{imacopilot2025} and \textbf{Doubao \citep{doubao2025}}. These agents achieve long-term personalization and social coherence, though at the cost of privacy management and memory drift. \textbf{Task-oriented agents} rely on contextual memory and specific domain knowledge to execute multi-step reasoning and maintain session continuity like GitHub Copilot \citep{githubcopilot}, Cursor \citep{cursor2024}, Coze \citep{coze2024}, DeepResearcher \citep{tongyidr}, WebSearcher \citep{tongyidr} and CodeBuddy \citep{zhao2024codebuddy}. For these agents, achieving a high task success rate remains the primary consideration for user satisfaction and practical effectiveness. \textbf{Multi-modal agents} represent a more integrated paradigm that unifies parametric and contextual memory across language, vision, and action modalities. Representative examples such as Mobile assistants (Doubao \citep{doubao2025}, Siri \citep{siri2025}, Xiaoyi \citep{xiaoyi2025}) and Embodied Agent extend memory beyond text to perception and embodiment, marking a step toward general, long-horizon agents.

Although these products have partially integrated memory-related functions, their memory scope and modality differ substantially across domains. ChatGPT and Doubao support long-range and cross-session adaptation through large-model backbones, but their memory management remains relatively simple and prone to hallucination. Their multimodal memory functions are limited to basic image-grounded retrieval rather than integrated cross-modal reasoning. Replika, as a personalization-oriented companion system, relies heavily on transparent and user-driven memory updates. However, its stored content depends entirely on user input, lacking autonomous management and raising privacy concerns, while higher-level session memory remains undeveloped. In contrast, GitHub Copilot, constrained by the complexity of programming tasks, operates mainly within a short-term working memory window without persistent task-level or project-level memory coordination, and lacks personalized code adaptation. Overall, these systems remain in an early stage of memory integration, where memory operations are largely prompt-based rather than dynamically managed. This gap highlights the need for more advanced development tools to support scalable, transparent, and adaptive memory mechanisms across products and domains.

\begin{table*}[t]
\centering
\small
\setlength{\tabcolsep}{3pt}
\renewcommand{\arraystretch}{1.2}
\caption{Memory Development Tools Trade-offs. Representative development tools are compared to highlight the special design and potential limitations.}
\begin{tabular}{@{}>{\bfseries}p{1.2cm}
                >{\raggedright\arraybackslash}p{2cm}
                >{\centering\arraybackslash}p{1.5cm}
                >{\raggedright\arraybackslash}p{1.5cm}
                >{\raggedright\arraybackslash}p{4cm}
                p{3cm}@{}}
\toprule
\textbf{Tool} & \textbf{Category} & \textbf{Memory Type} & \textbf{Prioritized Operations} & \textbf{Key Features} & \textbf{Limitations} \\ 
\midrule

EasyEdit \citep{wang-etal-2024-easyedit} &
 \textbf{Parametric Editing} &
Parametric &
Updating &
Directly modifies LLM weights (WISE \citep{wang2024wise}) &
Ripple Effects: Editing facts may damage general reasoning; high computational cost. \\
\midrule

Zep \newline \citep{rasmussen2025zep} &
\textbf{Temporal} Memory Construction &
Contextual &
Consolidation, Updating &
Temporal knowledge graphs; incremental summarization; robust temporal reasoning. &
Controllability; Information loss \\
\midrule

Mem0 \citep{mem0} &
\textbf{Personalized} Memory Layer &
Contextual &
Consolidation \newline Indexing \newline Updating &
User-level personalization across sessions; hybrid search (Vector + Graph); developer-friendly API. &
Lossy condensation; user-centric personalization rather than complex task or world-state memory. \\
\midrule

MemOS \citep{li2025memos} &
Memory \textbf{Scheduling} \& Hierarchical Management &
Contextual  &
Updating \newline Retrieval & Hierarchical OS-style scheduling (Short/Long/Working) for optimized context window and memory management. &
Control overhead; latency \\
\midrule

Graphiti \citep{he2025graphiti} &
\textbf{Graph} Memory Construction &
Contextual &
Indexing \newline Updating \newline Retrieval &
Dynamic construction of knowledge graphs from unstructured streams; semantic relationship tracking. &
Strictly typed graphs can be brittle with high token consumption for graph construction. \\


\bottomrule
\end{tabular}
\label{tab:tools_comparison}
\end{table*}

\subsection{Development Tools}
\label{sec:tools}

\textbf{Frameworks.} On top of core infrastructure, frameworks offer modular interface for memory-related operations. Examples include \textbf{Graphiti} \citep{he2025graphiti}, \textbf{LlamaIndex} \citep{llamaindex}, \textbf{LangChain} \citep{langchain}, \textbf{LangGraph} \citep{langgraph2025}, \textbf{EasyEdit} \citep{wang-etal-2024-easyedit}, \textbf{CrewAI} \citep{duan2024exploration}, MemU \citep{NevaMindAI2025memU}, and \textbf{Letta} \citep{packer2023memgpt}. These frameworks abstract complex memory processes into configurable pipelines, enabling developers to construct multi-modal, persistent, and updatable memory modules that interact with LLM agents. \textbf{Memory Layer Systems.} These systems operationalize memory as a service layer, providing orchestration, persistence, and lifecycle management. Tools like \textbf{Mem0} \citep{mem0}, \textbf{Zep} \citep{rasmussen2025zep}, \textbf{Memary} \citep{memary2025}, \textbf{MemOS} \citep{li2025memos} and \textbf{Memobase} \citep{memary2025} focus on maintaining temporal consistency, indexing memory by session or topic, and ensuring efficient recall. These platforms often combine symbolic and sub-symbolic memory representations and provide internal APIs for memory access and manipulation over time.


\subsection{Infrastructure}
Memory tools rely on a robust foundational infrastructure to operationalize the storage, retrieval, and evolution of memory. This infrastructure is anchored by persistent storage systems, such as graph databases like \textbf{Neo4j} \citep{neo4j2012} and vector stores \citep{douze2024faiss}, which work in tandem with retrieval mechanisms ranging from sparse \textbf{BM25} \citep{robertson1995okapi} to dense embedding retrieval \citep{izacard2021unsupervised, openai_embeddings_2025} to ensure precise access. The execution of complex memory lifecycle operations—including dynamic updating and targeted forgetting—depends on the reasoning capabilities of LLMs \citep{achiam2023gpt, liu2024deepseek} guided by optimized prompt engineering. Crucially, to support the high throughput and scalability required by these tools, the underlying computational layer incorporates acceleration technologies such as \textbf{FlashAttention} \citep{dao2024flashattention}, sequence parallelism, and efficient Key-Value (KV) cache management strategies \citep{kwon2023efficient}, all designed to enable the effective processing of ultra-long contexts and massive interaction histories.

\section{The Cognitive Gap between Biological and Agent Memory}
\label{sec:human_vs_agent_memory}

Human memory is not a monolithic storage but a complex, hierarchical interaction between sensory, short-term, and long-term systems \citep{baddeley1988cognitive}. While agents aim to emulate these functions to support reasoning, their underlying mechanisms are different from biological cognitive architectures. As summarized in Table~\ref{tab:memory_comparison}, current agentic implementations remain focused on static persistence, lacking the dynamic sophistication of biological memory in terms of encoding, evolving, and adapting.

\paragraph{Encoding: From Verbatim Recording to Constructive Schematization} Human encoding is inherently \textit{constructive}; we do not record snapshots but restructure the past through present cognition to fit internal schemas. In contrast, agents typically perform verbatim recording (in databases) or static parameterization (in weights), leading to an accumulation of fragmented traces rather than a synthesized self-model. This reliance on "raw" data prevents agents from pruning noise at the point of entry. While current training (e.g., pre/post-training \citep{liu2024deepseek}) attempts internalization, it remains a discrete process that fails to bridge the gap between static "knowing that" and the adaptive cognitive structures required to filter environmental complexity.

\paragraph{Evolving: From Summarization to Internalization} Memory evolution in humans relies on sleep-dependent reconsolidation, where episodic traces are distilled into semantic structures of general world knowledge. This active synthesis prunes noise and extracts causal patterns to prevent overfitting to immediate reality. In contrast, agent memory evolution depends on explicit operations like summarization or hard deletion to simulate memory dynamics. While frameworks such as ACE \citep{zhang2025agentic} utilize summarization for short-term buffer condensation, they primarily address immediate task resolution rather than long-term cognitive growth. Conversely, although systems like G-Memory \citep{zhang2025gmemory} construct long-term archives via hierarchical graphs, this remains a symbolic approach to evolution. These mechanisms treat memory like a static library that needs filing, whereas human memory is like a muscle. While agents can summarize a book (ephemeral experience), they fail to turn that knowledge into the instinctive skill (procedural wisdom) needed to perform a task naturally. Consequently, contemporary agents remain reactive note-takers limited by artificially compressed histories, lacking the capacity for long-lifecycle evolving or the construction of a consistent self-representation.

\paragraph{Adapting: From Retrieval to Meaning Construction}
Human memory utilization is a process of dynamic reconstruction driven by homeostatic needs and self-consistency \citep{baddeley1988cognitive}. In contrast, current agents predominantly rely on Retrieval-Augmented Generation (RAG) and extremely long context windows. While expanding the context window provides a larger buffer, it represents a brute-force architectural scaling that bypasses the necessity of semantic internalization. Such "long-context" dependency leads to a diminishing signal-to-noise ratio and prohibitive computational costs. As evidenced by the 'lost-in-the-middle' phenomenon \citep{liu2024lost}, retrieval without inference-time reconsolidation, characterized by the active rewriting of historical traces like AgentFold \citep{ye2025agentfoldlonghorizonwebagents} and FoldGRPO \citep{sun2025scalinglonghorizonllmagent}, struggles to develop a coherent causal representation. The next frontier is to move beyond passive retrieval toward active thinkers who reconstruct their internal state in real-time to adapt to environmental dynamics.

\paragraph{Storage, Ownership and Volume}
The divergence between biological and agent memory is rooted in the fundamental properties of their physical and systemic substrates, primarily manifested in storage, data ownership, and resource scalability. Regarding \textbf{storage}, human memory is characterized by biological \textbf{holistic interconnection}, enabling associative recall across the entire brain. Conversely, agents rely on \textbf{heterogeneous representations}—segregating data into disconnected formats like documents, graphs, and vector embeddings—which prioritizes local pattern matching over global semantic coherence. This systemic gap extends to \textbf{data ownership}: human memory is inherently private and individual-bounded, whereas agent memory is \textbf{replicable and broadcastable}, enabling collective intelligence but challenging the ethical "right to be forgotten." Finally, while the human brain achieves complex memory evolving with extreme metabolic frugality ($\sim$20W) \citep{balasubramanian2021brain}, agent memory remains constrained by the computational and environmental costs of silicon-based scaling, necessitating a shift toward bio-inspired efficiency that prioritizes semantic density over raw data volume.

\begin{table*}[t!]
\centering
\small
\renewcommand{\arraystretch}{1}
\begin{tabular}{p{2.5cm} p{5cm} p{5cm}}
\toprule
\textbf{Aspect} & \textbf{Human Memory} & \textbf{Agent Memory} \\
\midrule
\multirow{2}{*}{\raggedright\arraybackslash \textbf{Storage}} & Distributed, interconnected neural systems across brain regions & Model parameter, modular, and context-dependent \\ \addlinespace

\textbf{Ownership} & Individual and private. & Shareable, replicable, and broadcastable. \\
\addlinespace
\textbf{Volume} & Biologically limited & Scalable, bounded only by storage and compute limits \\ \addlinespace
\textbf{Memory Encoding} & Slow, biologically driven, passive & Fast, explicit, policy-driven and selective \\ \addlinespace
\textbf{Memory Evolving} & Indirect, reconsolidation-based, error-prone & Precise, programmable, supports rollback/unlearning \\
\addlinespace
\textbf{Memory Adaption} & Implicit, salience- and frequency-biased & Explicit, customizable (e.g., quantization, summarization) \\
\bottomrule
\end{tabular}
\caption{Key differences between human and agent memory.}
\label{tab:memory_comparison}
\end{table*}

\section{Open Challenges and Future Directions}
\label{sec:future_direction}
This section outlines the open challenges in core memory topics and proposes future research directions.  We then explore broader perspectives, including biologically inspired models, lifelong learning, multi-agent memory, and unified memory representation, which further extend the capabilities and theoretical grounding of memory systems. Together, these discussions provide a roadmap for advancing reliable, interpretable, and adaptive memory in AI.

\subsection{Topic-Specific Directions}
Designing memory-centric AI requires addressing core limitations and emerging demands. Guided by RCI analysis and trends, we outline key challenges shaping future memory research.

\textbf{Unified evaluation is needed to address consistency, personalization, and temporal reasoning in long-term memory.} Existing benchmarks rarely assess core operations such as consolidation, updating, retrieval, and forgetting in dynamic, multi-session settings. This gap contributes to the retrieval–generation mismatch, where retrieved content is often outdated, irrelevant, or misaligned due to poor memory maintenance. Addressing these issues requires temporal reasoning, structure-aware generation, and retrieval robustness, along with systems supporting personalized reuse and adaptive memory management across sessions.

\textbf{Long-context Processing: Efficiency vs. Expressivity.} Scaling memory length exacerbates trade-offs between computational cost and modeling fidelity. Techniques such as KV cache compression and recurrent memory reuse offer efficiency but risk information loss or instability. Meanwhile, reasoning over complex environments, especially in multi-source or multi-modal settings, requires selective context integration, source differentiation, and attention modulation. Bridging these demands, mechanisms that balance contextual bandwidth with task relevance and stability, increasingly pointing toward the use of RL-based frameworks to learn active optimal context management and folding policies.

\textbf{While promising, parametric memory modification requires further research to improve control, erasure, and scalability.} Current editing methods often lack specificity, while unlearning benchmarks like TOFU may be too simple to expose real limitations. Most approaches fail to scale beyond thousands of edits or support models over 20B parameters. Lifelong learning remains underexplored despite its potential. Future work should develop more realistic benchmarks, improve efficiency, and unify editing, unlearning, and continual learning into a cohesive framework.

\textbf{Multi-source Integration: Consistency, Compression, and Coordination.} Modern agents rely on heterogeneous memory comprising structured knowledge, unstructured histories, and multi-modal signals but face redundancy, inconsistency, and ambiguity. These stem from misaligned temporal scopes, conflicting semantics, and missing attribution across modalities. Resolving them requires conflict resolution, temporal grounding, and provenance tracking. Efficient indexing and compression are essential for scalability and interpretability in multi-session settings.

\subsection{Broader Perspectives}
In addition to the core topics outlined above, a range of broader perspectives is emerging that further enriches the landscape of memory-centric agents. 

\par{\textbf{Procedural Tool Memory and Skill Acquisition.}} As agents become more action-oriented, memory needs to evolve from static fact storage toward procedural tool memory, where tool use is internalized as reusable skills rather than repeatedly consulting the tool API during extended interactions. Frameworks such as ReAct \citep{yao2023react} already hint at this shift by coupling reasoning with action trajectories, enabling agents to learn from execution feedback instead of treating tools as stateless calls. Recent infrastructure, including MCP \citep{anthropic2024mcp} servers, further supports this evolving by framing tools as persistent services that allow for experience accumulation across interactions. Benchmarks like BFCL v4 \citep{patil2024berkeley} explicitly expose the need for memorizing execution traces, error-recovery strategies, and tool-chain compositions, rather than relying on ad hoc prompting. Industrial systems have begun to operationalize this idea, exemplified by Anthropic’s introduction of skills \citep{anthropic2025claudeskills} in Claude, which treat tool use as a form of procedural memory that improves reliability and reduces inference cost.

\par{\textbf{Parametric Sharing: Memory as Dynamic Weights.}} While textual memory (e.g., RAG) provides transparency, it inevitably suffers from information loss during compression and natural language conversion. We propose Parametric Sharing, where memory is exchanged as model-native representations \citep{berges2024memory}—such as dynamic adapters or specialized memory layers—directly within the latent space \citep{zou2025latentcollaboration}. This approach preserves high-dimensional semantic nuances and enhances the collective reasoning of fused systems by bypassing the "bottleneck" of explicit text. Future work should explore standardized neural memory protocols and cross-model weight alignment to enable heterogeneous agents to merge internalized experiences into a collaborative parametric intelligence.

\par{\textbf{Lifelong Learning.}} Future research should shift from discrete task-based learning to managing real-time environment streams \citep{zhang2025agentic}, focusing on mitigating catastrophic forgetting while maintaining rapid adaptation \citep{feng2024tasl}. Under extreme data sparsity, agents must utilize meta-learning to optimize a "memory value function," enabling the autonomous determination of "solidification value" for selective memory internalization \citep{tian2024forget}. Crucially, personalized representations must transcend the restrictive inductive bias of the base model's pre-trained distribution, which often suppresses unique individual traits. By integrating structural \citep{rasmussen2025zep} and unstructured memory \citep{bae2022keep} into a dynamic personalized parameter space (e.g., evolving LoRA or embeddings), agents can decouple personal traces from general knowledge. This ensures causal consistency across infinite horizons, evolving agents from task-oriented tools into longitudinal, habit-aware companions.

\par{\textbf{Memory in Multi-agent Systems.}} In multi-agent systems, memory is not only individual but also distributed. Agents must manage their own internal memories while interacting with and learning from others \citep{wang2025mirix}. This raises unique challenges such as memory sharing, alignment, conflict resolution, and consistency across agents. Effective multi-agent memory systems should support both local retention of personalized experiences and global coordination through shared memory spaces or communication protocols. Future work may explore decentralized memory architectures, cross-agent memory synchronization, and collective memory consolidation to enable collaborative planning, reasoning, and long-term coordination.

\par{\textbf{Multi-modal Memory.}} Multi-modal memory inherently reflects how humans perceive the real world. While advancements like M3-agent \citep{long2025seeinglisteningrememberingreasoning} and GUI-agent \citep{hong2023cogagent} have explored multi-modal memory processing capabilities, this field remains in its preliminary stages. Significant challenges persist in aligning multi-modal memories within a unified semantic space and enabling effective retrieval and reasoning. Specifically, current systems suffer from weak reasoning during multi-turn interactions and data misalignment, highlighting critical directions for future research.

\par{\textbf{Biological Inspirations for Memory Design.}} Memory in biological systems offers key insights for building more resilient and adaptive AI memory architectures. The brain manages the stability–plasticity dilemma through complementary learning systems: the hippocampus encodes fast-changing episodic experiences, while the cortex slowly integrates stable long-term memory \citep{mcclelland1995complementary, kumaran2016learning}. Inspired by this, AI models increasingly adopt dual-memory architectures, synaptic consolidation, and experience replay to mitigate forgetting \citep{ritter2018meta, wang2021dual}. Cognitive concepts like memory reconsolidation \citep{dudai2015consolidation}, bounded memory capacity \citep{cowan2001magical}, and compartmentalized knowledge \citep{franklin2020structured} further inform strategies for update-aware recall, efficient storage, and context-sensitive generalization. 

Meanwhile, the K-Line Theory~\cite{MINSKY1980117} points out that hierarchical memory structures are fundamental to biological cognition. These structures enable humans to efficiently organize memory across different levels of abstraction, as seen in how infants group specific objects like "apple" and "banana" into broader categories like "fruit" and "food." Organizing the agent memory with hierarchy structures for scalability and efficiency raises new challenges~\cite{wang-etal-2024-abspyramid, han-etal-2025-concept} and future directions~\cite{wang-etal-2024-absinstruct, hong-etal-2024-abstraction} for memory research.

\par{\textbf{Parametric Memory Retrieval.}} While recent knowledge editing methods \citep{fang2025alphaedit, wang2024wise} claim they can localize and modify specific representations, enabling models to selectively retrieve knowledge from their own parameters remains an open challenge. Efficient retrieval and integration of latent memory could significantly enhance memory utilization and reduce dependence on external indexing and memory management.

\par{\textbf{Spatio-temporal Memory}} captures not only the structural relationships among information but also their temporal evolution, enabling agents \citep{lei2025stma} to adaptively update knowledge while preserving historical context \citep{zhao2025eventweave}. For example, the agent may record that a user once disliked broccoli but later adjusts its memory based on recent purchase patterns. By maintaining access to both historical and current states, spatio-temporal memory supports temporally informed reasoning and nuanced personalization. However, efficiently managing and reasoning over long-term spatio-temporal memory remains a key challenge.

\par{\textbf{Unified Memory Representation.} While parametric memory \citep{yang2024text} provides compact and implicit knowledge storage, and external memory \citep{zhong2024memorybank} offers explicit and interpretable information, unifying their representational spaces and establishing joint indexing mechanisms is essential for effective memory consolidation and retrieval. Future work could focus on developing unified memory representation frameworks that support shared indexing, hybrid storage, and memory operations across modalities and knowledge forms.}

\par{\textbf{Memory Threats \& Safety.}}
While memory significantly enhances the utility of LLMs by enabling up-to-date and personalized responses, its management remains a critical safety concern. Memory often stores sensitive and confidential data, making operations like adding or removing information far from trivial. Recent research has exposed serious vulnerabilities in memory handling, particularly in machine unlearning techniques designed to selectively erase data. Multiple studies \citep{machine-unlearning-threat-survey-ziyao-etal-2025, barez2025openproblemsmachineunlearning} have demonstrated that these methods are prone to malicious attacks, which strengthens the need for more secure and reliable memory operations.

\section{Conclusions}
This survey provides a comprehensive overview of agent memory, classifying it into parametric and contextual types and mapping operations to encoding, evolving, and adapting. Complemented by functional perspectives like episodic, semantic, procedural, and working memory, this framework clarifies how memory supports reasoning, personalization, and collaboration. By analyzing four key topics, including long-term memory, long context memory, parametric modification, and multi-source memory, we highlight progress, challenges, and pathways for future work, while offering practical benchmarks and tool guidance for industry. 
\bibliographystyle{ACM-Reference-Format}
\bibliography{new_custom}

\newpage
\appendix
\section*{Appendix}
\section{GPT-based Pipeline Selection}
\label{app:relevance_scoring}
To facilitate large-scale relevance filtering aligned with our taxonomy, we design a GPT-based scoring pipeline to evaluate the alignment between paper abstracts and predefined task definitions (Table~\ref{tab:task_definitions}). Each abstract is paired with a corresponding task definition and scored on a 1–10 scale by the model, with a threshold of $\geq$ 8 used to retain high-relevance papers for further analysis. We adopt \textbf{GPT-4o-mini} as the scoring backbone due to its favorable trade-off between performance and efficiency. Despite its relatively lightweight architecture, GPT-4o-mini demonstrates strong zero-shot reasoning capabilities, making it a cost-effective and sufficiently accurate choice for abstract-level topic relevance estimation across a corpus of over 30,000 papers. The exact prompt format used in this evaluation process is illustrated in Figure~\ref{fig:prompt_relevance_task_eval}.

\section{Relative Citation Index}
\label{rel_citation_indx}
In this work, we identify impactful works by Relative Citation Index (RCI) metric inspired by the RCR metrics \citep{10.1371/journal.pbio.1002541}, which estimate the expected citations with respect to publication age to prevent bias between original citations from different publication dates. The age $A_i$ of a paper $p_i$ is computed as:
\begin{equation}
    A = T - Year_i 
\end{equation}
, where $T$ is the date when the citation is collected (20th April 2025) and $Year_i$ is the year where paper $i$ is first published. Thus, we can model the relation between citation number $C_i$ and age $A_i$ of paper $p_i$ in three different way, which are:

\textbf{linear model}: 
\begin{equation}
    C_i = \beta + \alpha A_i
\end{equation}

\textbf{exponential model}: 
\begin{equation}
    C_i = \exp(\beta + \alpha A_i)
\end{equation}

\textbf{log-log regression model}: 
\begin{equation}
    \log(C_i+1)=\beta+\alpha\log A_i + \epsilon_i
\end{equation}

\begin{figure}
    \centering
    \includegraphics[width=0.6\linewidth]{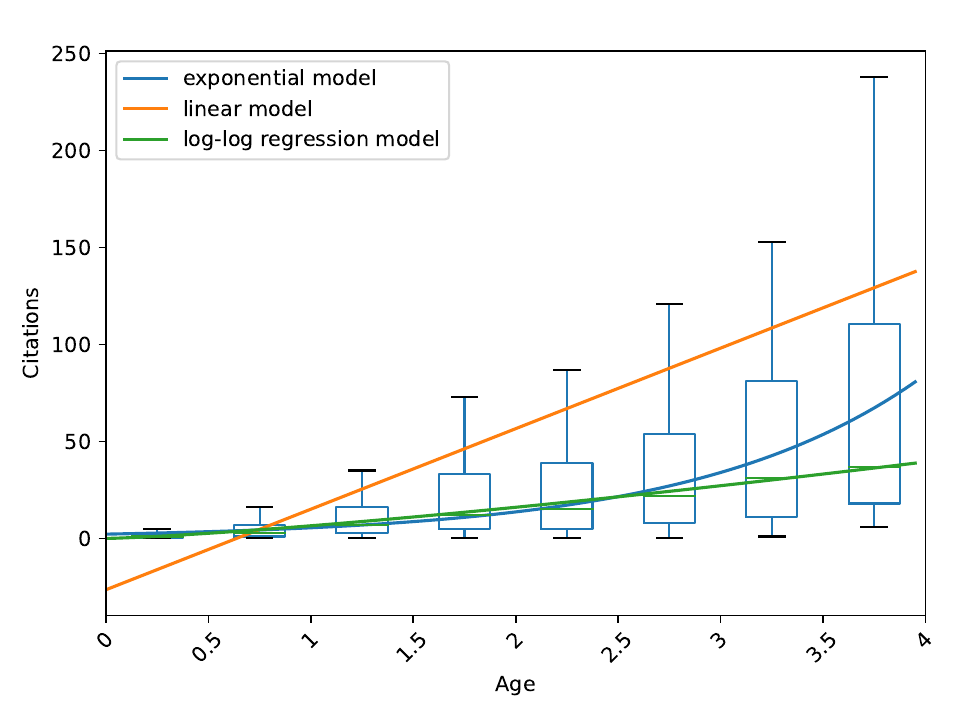}
    \caption{Boxplot of citation distributions from the 3,932 papers with respect to age, red curve is the expected citations $\hat{C_i}$. Generally $RCI>=1$ indicate the paper is above median citations in its age group, and  higher $RCI$ indicate higher research impact.}
    \label{fig:rci_boxplot}
\end{figure}

We collect papers from past 3 years (2022 to 2025) from Top NLP and ML conferences (i.e., ACL, NAACL, EMNLP, NeurIPS, ICML, ICLR). To reduce the bias from different research area, we use GPT to score the relevance of a paper with the four topics discussed in the paper, using the prompt shown in Figure~\ref{fig:prompt_relevance_task_eval}. We pick all the papers with score equal and higher than 8 and collect their publication date and citation numbers from Semantic Scholar API\footnote{\url{https://www.semanticscholar.org/product/api}}. For papers without publication date field, we use the first conference day as the publication date. We gather a total number of 3,932 valid papers after the processing and compute the estimated $\hat{\beta}$ and $\hat{\alpha}$ accordingly\footnote{Noted that not all papers mentioned in this work are considered in estimating $\hat{\beta}$ and $\hat{\alpha}$, but they will be assigned a RCI score based on the publication age.}. Figure~\ref{fig:rci_boxplot} shows the estimated age-citation model, where we can find that the log-log regression model best fit the data, which almost perfectly fitting the median citation with respect to publication age. In addition, log-log regression model grantees that the expected citation equals 0 when a paper is freshly released, which follows the intuition. Thus, we pick log-log regression model to compute the expected citation for next step\footnote{The estimation is: $\hat{\beta}=1.878$, $\hat{\alpha}=1.297$}, and we are able to obtain the expected citation number $\hat{C_i}$ of paper $p_i$ with age $A_i$ as:
\begin{equation}
    \hat{C_i} = \exp(\hat{\beta}) A_i^{\hat{\alpha}}
\end{equation}
Then we compute the relative citation index $RCI_i$ of paper $p_i$ as:
\begin{equation}
    RCI_i = \frac{C_i}{\hat{C_i}}
\end{equation}
When $RCI_i >= 1$, we consider this paper over-cited than its expectations, and vice versa. In this paper, we focus on the paper with $RCI>=1$, for which we believe has more influence.

\section{RCI-Driven Analysis of Topic Impact}
\begin{figure}[t!]
  \centering
  \includegraphics[width=0.5\textwidth]{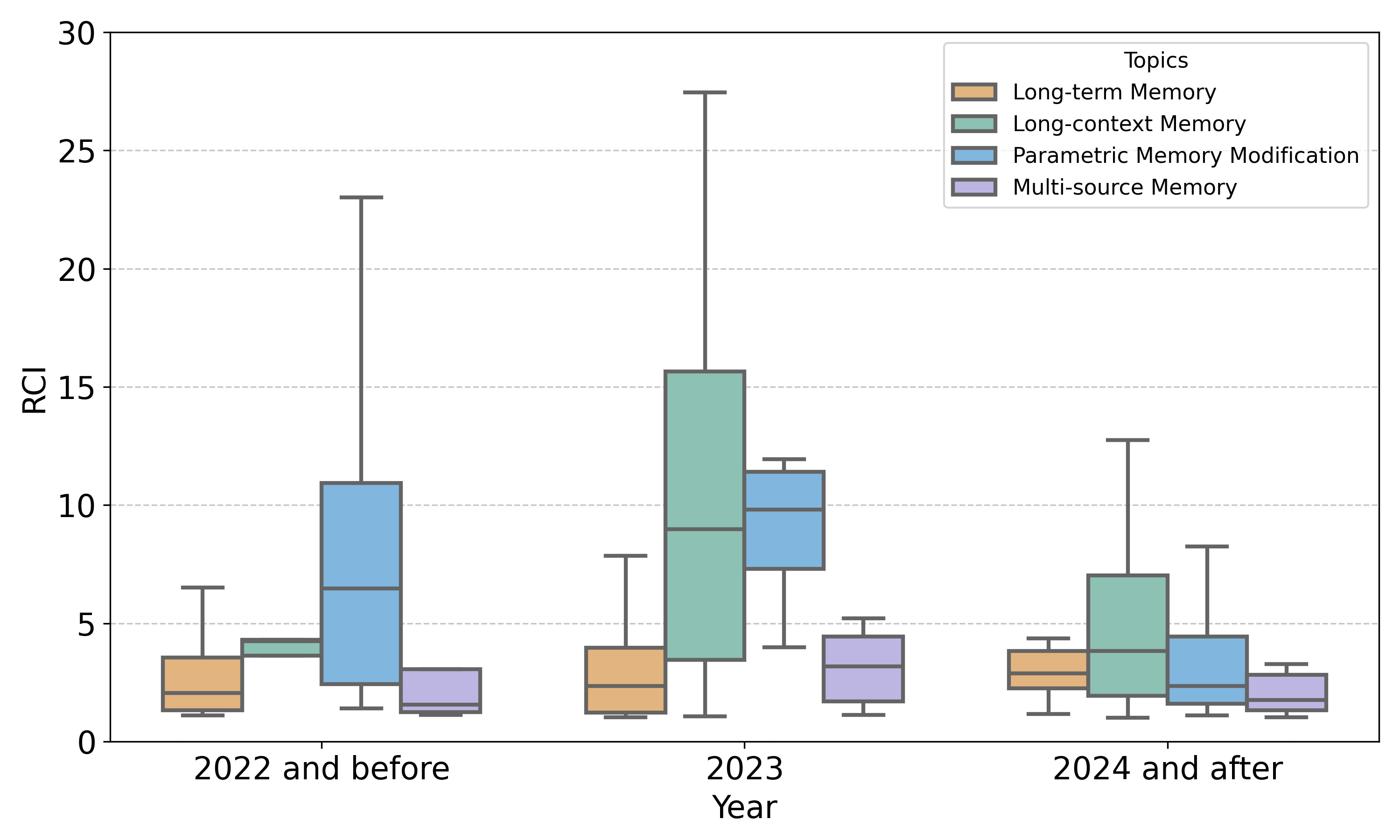}
  \caption{Overall distribution of median RCI across topics and years}
  \label{fig:appendix_box_trend}
\end{figure}

In this study, we leverage both RCI and publication volume trends to gain a clearer understanding of the development and influence of various memory-related research topics. As shown in Figure \ref{fig:appendix_box_trend}, boxplots illustrate the distribution of median Relative Citation Index (RCI) values across topics by year. Notably, 2023 stands out as a pivotal year following the emergence of large language models (LLMs), with a surge in both the quantity and quality of publications related to long-context and parametric memory, suggesting that these areas were directly shaped by the advancement of LLMs. In contrast, long-term memory and multi-source memory maintained relatively stable average impact levels, indicating continued activity without the emergence of disruptive or field-defining work during that period.

\begin{figure}[t!]
  \centering
  \includegraphics[width=0.5\textwidth]{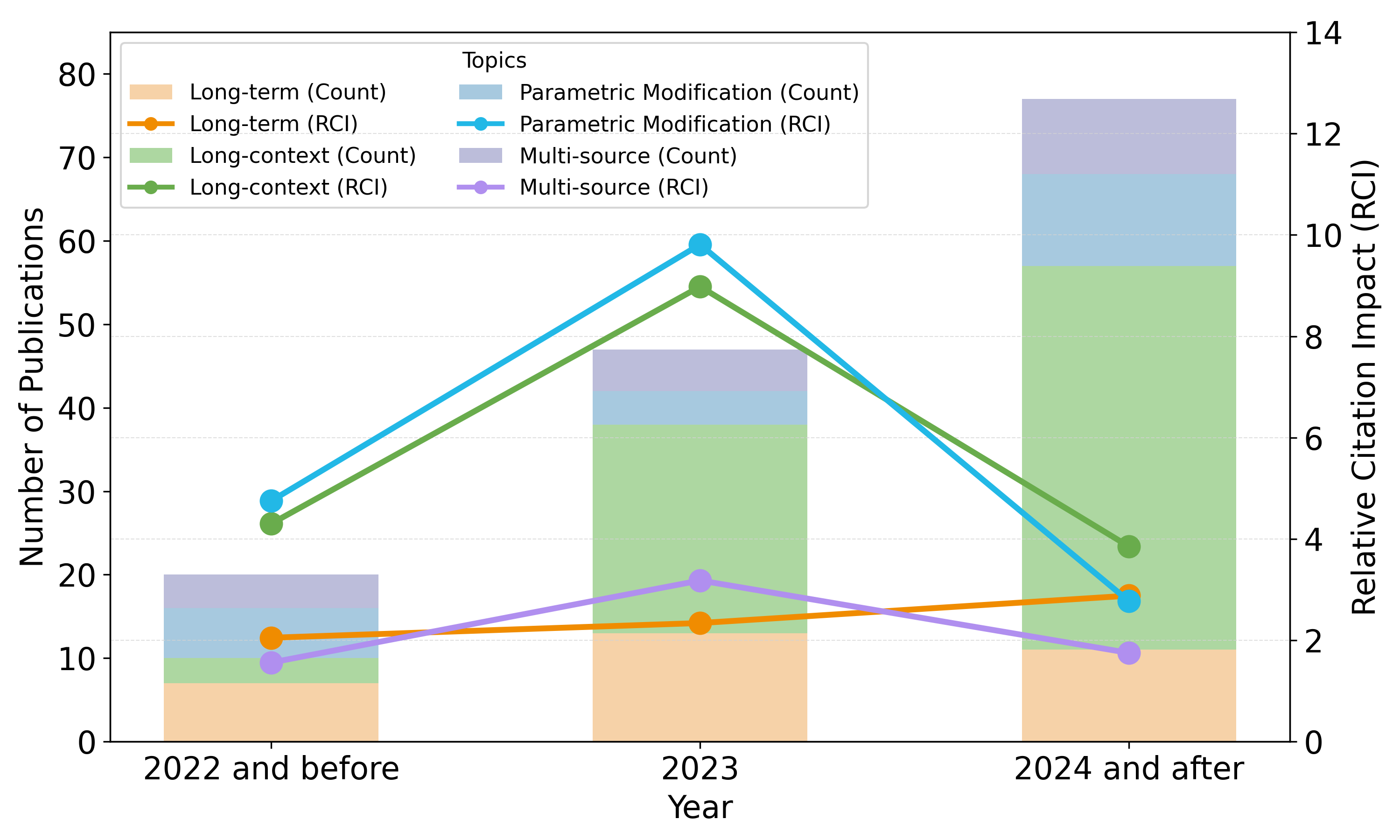}
  \caption{Overall temporal trends of topic-wise publication volume and median RCI.}
  \label{fig:appendix_volume_trend}
\end{figure}

Figure \ref{fig:appendix_volume_trend} visualizes the temporal trends in publication volume and median RCI for each topic. All topics experienced notable growth in publication counts, with long-context in particular expanding from one of the least represented topics before 2022 to the most prominent by 2024—largely driven by the rise of LLMs. Furthermore, the RCI of long-term memory has shown a steady increase, reflecting a growing body of valuable work in that domain. By contrast, other topics witnessed a noticeable decline in RCI medians after 2023, though their influence levels remained comparable to those seen prior to 2022. These patterns collectively underscore the substantial impact of large models in catalyzing progress across memory-related research, especially in the areas of long-context and parametric memory.

\section{Chord Analysis of Interactions Among Memory Types, Operations, Topics, and Venues}
We present a chord-based analysis of memory research from two perspectives: (1) the interactions among memory types, operations, and topics, and (2) their distribution across major ML and NLP conference venues.

\subsection{Memory Interactions Across Types, Operations, and Topics}
\begin{figure*}[htbp]
  \centering
  \includegraphics[width=0.7\textwidth]{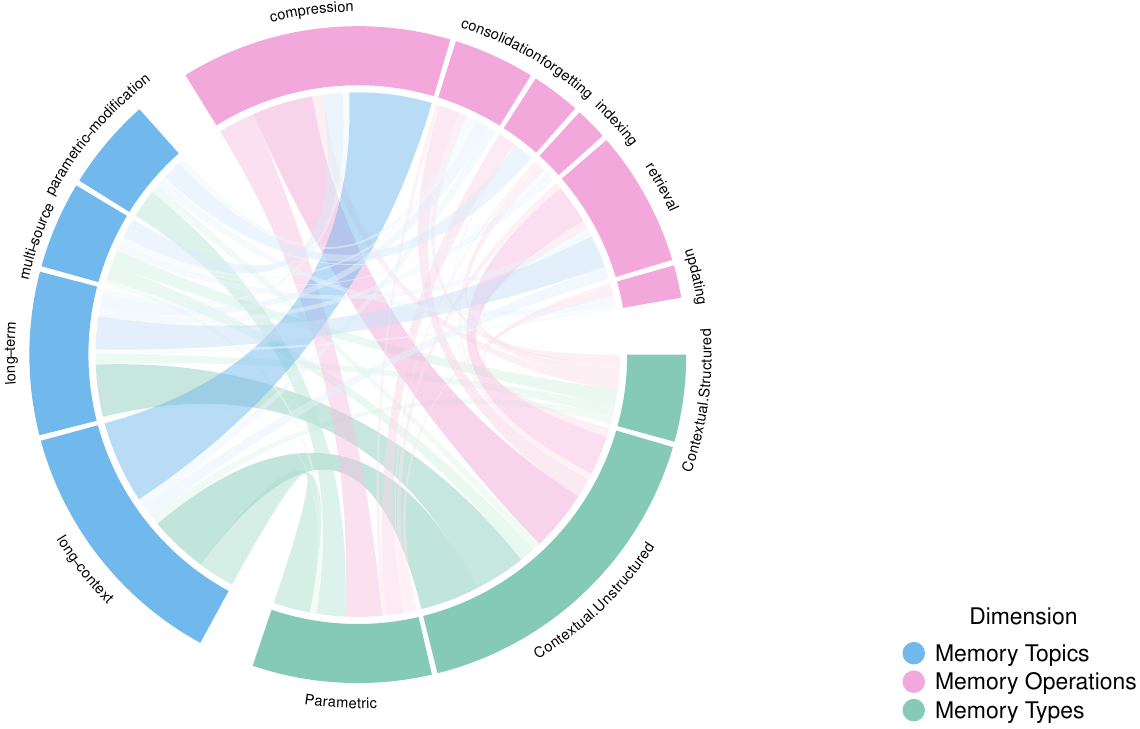}
  \caption{Chord Map of Interactions Across Memory Topics, Operations, and Types.}
  \label{fig:appendix_chord_map_type}
\end{figure*}

To intuitively analyze the strength of connections between memory types, operations, and research topics, we examine 132 method-focused papers with an RCI $\geq$ 1 and generate a final chord diagram (as shown in Figure \ref{fig:appendix_chord_map_type}) based on the analysis. 

From the perspective of memory types, research predominantly focuses on parametric memory and contextual unstructured memory, with most work centered on compression, retrieval, forgetting, and updating. In contrast, contextual structured memory is relatively underexplored, likely because LLMs are optimized for sequential text and perform less effectively on structured inputs.

From the operation perspective, compression and retrieval are the most frequently studied, while indexing receives comparatively less attention. This is largely because most existing works focus on the use of memory, where retrieval and compression are two fundamental operations. In the case of consolidation, most studies refer to storing knowledge either in model parameters via training on unstructured text or transforming it into a fixed external memory format. Updating and forgetting are mainly associated with knowledge editing and unlearning, typically within parametric memory. These directions aim to incrementally modify parameters in the model based on external input. However, due to the opaque nature of model internals, such memory operations remain at an early stage of active exploration. In contrast, memory indexing mechanisms for LLMs have received limited attention.

From the topic perspective, parametric modification studies are mostly centered on parametric memory, though some works attempt parameter adaptation through continual learning over unstructured text. Research under the long-context theme primarily focuses on compression and retrieval within unstructured memory, with some leveraging parameterized forms like key-value caches. In long-term memory studies, the emphasis is also on unstructured memory, particularly in terms of consolidation, compression, and retrieval. Research related to multi-source memory is still limited and typically involves integrating structured and unstructured information.

In summary, the limited exploration of contextual structured memory highlights an opportunity to develop more comprehensive memory operations by integrating it with unstructured memory. Second, research on multi-source memory remains scarce, despite the substantial challenges it poses—particularly the issue of memory conflicts arising from heterogeneous sources. Designing robust and consistent strategies for multi-source memory integration is thus a promising direction. Finally, although indexing has been extensively studied in traditional database systems, it remains underexplored in the context of LLM-based agents. The complexity of memory types and the need for vectorized or sparse retrieval methods call for new indexing approaches specifically tailored to reasoning and interaction in LLMs.

\subsection{Memory Interactions Across Conference Venues}
\begin{figure*}[htbp]
  \centering
  \includegraphics[width=0.7\textwidth]{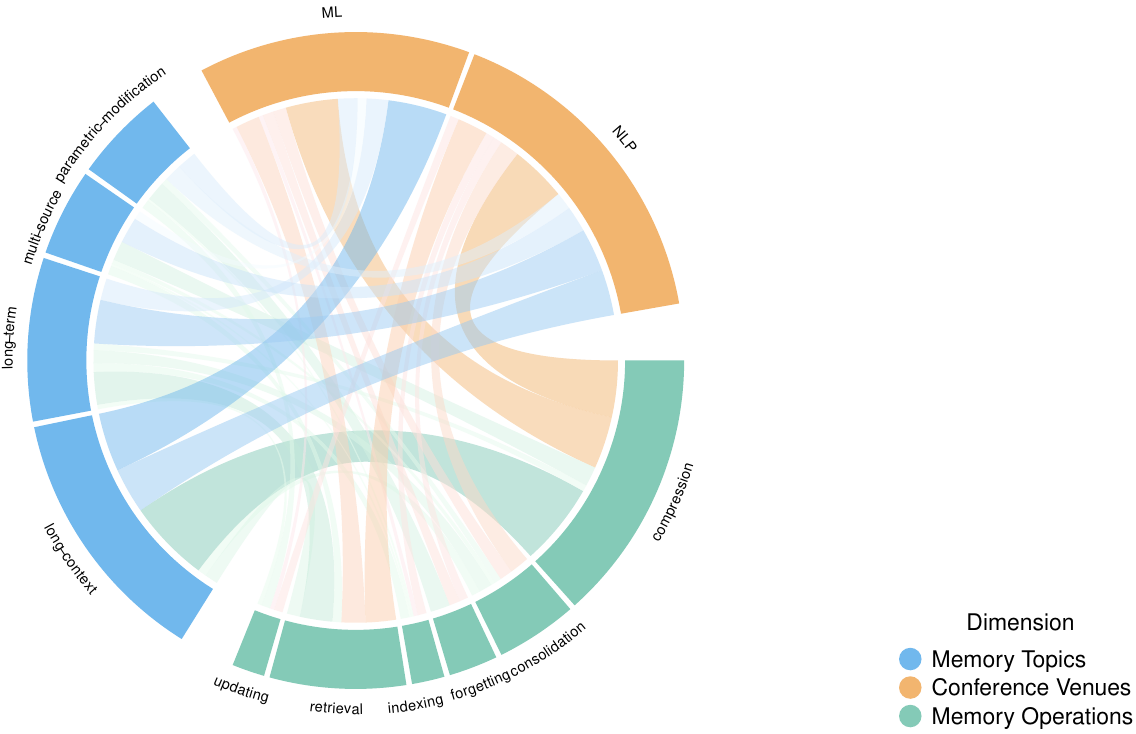}
  \caption{Chord Map of Interactions Across Memory Topics, Operations, and Conference Venues.}
  \label{fig:appendix_chord_map_conference}
\end{figure*}

In addition to our primary paper collection, we also analyzed 81 method-focused papers with RCI $\geq$ 1 across major conferences. As shown in Figure \ref{fig:appendix_chord_map_conference}, from the operation perspective, compression, forgetting, and updating appear more frequently in ML conferences (ICLR, ICML, NeurIPS), while retrieval and consolidation are more commonly featured in NLP conferences (ACL, EMNLP, NAACL). This distribution suggests that the former set of operations is still in the stage of theoretical exploration, whereas the latter is more grounded in practical application. Consequently, compression, forgetting, and updating still hold substantial potential for translation into real-world systems.

Indexing remains underrepresented in both ML and NLP venues. This may be partly due to its frequent co-occurrence with retrieval, and partly because current vector-based indexing approaches are relatively uniform, with few novel alternatives available.

From the topic perspective, long-term memory is more frequently addressed in NLP conferences, while long-context topics are more common in ML venues—likely reflecting the differing application- and theory-oriented focuses of these communities. Parameter modification appears more often in ML conferences, whereas multi-source memory is more prevalent in NLP conferences, highlighting the fact that multi-source memory challenges often arise during real-world applications and system integration.

\begin{table*}[ht]
\centering
\small
\renewcommand{\arraystretch}{1.2}
\begin{tabular}{p{3.4cm}p{11.6cm}}
\toprule
\textbf{Topic Name} & \textbf{Definition in Prompt} \\
\midrule
Long-Term Memory &
\textbf{Definition:} Creating systems that ensure knowledge from past interactions remains accessible as new tasks emerge, maintaining continuity in multi-turn conversations. \newline
\textbf{Features:} Memory retention, retrieval, and attribution—preserving, accessing, and contextualizing memory to support coherent interaction. \\
\midrule
Long-Context &
\textbf{Definition:} Efficiently processing, interpreting, and utilizing very long input sequences without performance degradation. \newline
\textbf{Features:} Optimized attention, context compression, and mitigation of the “lost-in-the-middle” problem. \\
\midrule
Parametric Memory Modification &
\textbf{Definition:} Managing and updating internal parameters to preserve accuracy, privacy, and adaptability without full retraining. \newline
\textbf{Features:} Selective unlearning, precise model editing, distillation, and lifelong learning. \\
\midrule
Multi-Source&
\textbf{Definition:} Integrating and harmonizing diverse data types into a unified framework while resolving inconsistencies. \newline
\textbf{Features:} Multi-modal fusion, semantic consistency, conflict resolution, and redundancy removal. \\
\midrule
Personalization* &
\textbf{Definition:} Building user-centric memory systems that adapt to individual preferences and history while preserving privacy. \newline
\textbf{Features:} Privacy-aware profiling, consistent personalization, and long-term continuity. \\
\bottomrule
\end{tabular}
\caption{Definitions and features of the five memory-centric evaluation topics. *Personalization is treated as a specialized form of long-term memory that focuses on user-centric adaptation across sessions.}
\label{tab:task_definitions}
\end{table*}

\begin{figure*}[ht]
\begin{tcolorbox}[
colframe=black!75!white,
colback=white, sharp corners,
boxrule=0.8pt, width=\textwidth,
title=Prompts of the Relevance Evaluation to Task Definitions
]
\small
\textbf{System Instruction:}  
Given the task and the abstract, evaluate the relevance of the abstract to the task.  

\textbf{Prompt Template:}\\
"""\\
You are tasked with evaluating the relevance of a given article to a specific task definition. \\
Please read the following task definition, article title, and abstract carefully. \\
Based on the content, rate the relevance on a scale from 1 to 10, \\where 1 means not relevant at all, 
and 10 means highly relevant.

Task Definition: $\{task_{def}\}$ \\
Article Title: $\{title\}$ \\
Abstract: $\{abstract\}$ \\

Please provide your rating in the format [[Rating]].\\For example, if the relevance is high, you might respond with [[9]].
"""
\end{tcolorbox}
\caption{Prompt for evaluating article relevance to specific task definitions.}
\label{fig:prompt_relevance_task_eval}
\end{figure*}

\definecolor{LT}{HTML}{fbefcf}

\begin{table*}[htbp]
\centering
\arrayrulecolor{black}
\resizebox{\textwidth}{!}{
\begin{tabular}{
    >{\arraybackslash}m{3cm}  
    >{\arraybackslash}m{1.3cm}  
    >{\arraybackslash}m{1.8cm}  
    >{\arraybackslash}m{1cm}  
    >{\arraybackslash}m{0.6cm}  
    >{\arraybackslash}m{0.6cm}  
    >{\arraybackslash}m{2.5cm}  
    >{\arraybackslash}m{6cm}  
    >{\arraybackslash}m{1cm}  
}
\toprule
\textbf{\rule{0pt}{1.2em}Datasets} & \textbf{Mo} & \textbf{Operations} & \textbf{DS Type} & \textbf{Per} & \textbf{TR} & \textbf{Metrics} & \textbf{Purpose} & \textbf{Year} \\
\toprule
\textbf{PersonaMem-v2} \newline \citep{jiang2025personamem} 
& text 
& Updating,\newline Retrieval 
& MS 
& ✓ 
& ✓ 
& Accuracy, \newline Persona Score 
& Benchmarks implicit user persona learning and agentic memory updates over long contexts. 
& 2025 \\

\textbf{MemoryBench} \newline \citep{ai2025memorybench} 
& text 
& Updating,\newline Retrieval,\newline Forgetting 
& QA 
& ✗ 
& ✓ 
& Accuracy, \newline Retention Rate 
& Comprehensive benchmark for memory correctness, persistence, and continual learning. 
& 2025 \\

\textbf{HaluMem} \newline \citep{chen2025halumem} 
& text 
& Retrieval 
& QA 
& ✗ 
& ✗ 
& Accuracy, \newline Hallucination Rate, \newline Omission Rate 
& Evaluates hallucinations in memory extraction, updating, and retrieval. 
& 2025 \\

\textbf{BFCL V4} \newline \citep{patil2024berkeley} 
& text (API/Code) 
& Updating,\newline Retrieval,\newline Reasoning 
& QA (API) 
& ✗ 
& ✗ 
& AST Accuracy, \newline Execution Success 
& Benchmarks function-calling capabilities, featuring a dedicated memory category for CRUD-style tool usage. 
& 2025 \\

\textbf{LongMemEval} \newline \citep{wu2024longmemeval} & text & Indexing, \newline Retrieval, \newline Compression & MS & ✗ & ✓ & Recall@K, NDCG@K, \newline Accuracy & Benchmarks chat assistants on long-term memory abilities, including temporal reasoning. & 2024 \\

\textbf{LoCoMo} \newline \citep{maharana2024evaluating} & text + image & Indexing, \newline Retrieval, \newline Compression & MS & ✗ & ✓ & Accuracy, ROUGE, \newline Precision, Recall, F1 & Evaluates long-term memory in LLMs across QA, event summarization, and multimodal dialogue tasks. & 2024 \\

\textbf{MemoryBank} \newline \citep{zhong2024memorybank} & text & Updating, \newline Retrieval & MS & ✓ & ✗ & Accuracy, \newline Human Eval & Enhances LLMs with long-term memory, adapting to user personalities and evolving contexts. & 2024 \\

\textbf{PerLTQA} \newline \citep{du2024perltqa} & text & Retrieval & MS & ✓ & ✗ & MAP, Recall,\newline Precision, F1, \newline Accuracy, GPT-4 score & Evaluates personalized long-term memory QA capabilities. & 2024 \\

\textbf{MALP} \newline \citep{zhang-etal-2024-llm-based} & text & Retrieval, \newline Compression & QA & ✓ & ✗ & ROUGE, Accuracy, Win Rate & Preference-conditioned dialogue generation; supports PEFT customization. & 2024 \\

\textbf{DialSim} \newline \citep{kim2024dialsim} & text & Retrieval & MS & ✓ & ✗ & Accuracy & Evaluates dialogue systems under realistic, real-time, and long-context multi-party conversations. & 2024 \\

\textbf{CC} \newline \citep{jang2023conversation} & text & Retrieval & MS & ✗ & ✓ & BLEU, ROUGE & Models long-term dialogues incorporating temporal and relationship contexts. & 2023 \\

\textbf{LAMP} \newline \citep{salemi2023lamp} & text & Consolidation,\newline Retrieval,\newline Compression & MS & ✓ & ✓ & Accuracy, F1, ROUGE & Provides user- and time-based splits for evaluating short- and long-term personalization. & 2023 \\

\textbf{MSC} \newline \citep{xu2021beyond} & text & Consolidation,\newline Retrieval,\newline Compression & MS & ✓ & ✗ & PPL & Multi-session human-human chats for evolving shared knowledge evaluation. & 2022 \\

\textbf{DuLeMon} \newline \citep{xu2022long} & text & Consolidation,\newline Updating,\newline Retrieval, Compression & MS & ✓ & ✗ & Accuracy, F1, \newline Recall, Precision, \newline PPL, BLEU, \newline DISTINCT & Supports dynamic persona tracking and consistent long-term human-bot interactions. & 2022 \\

\textbf{2WikiMultiHopQA} \newline \citep{ho2020constructing} & table + \newline knowledge base + text & Consolidation, \newline Indexing, \newline Retrieval, \newline Compression & QA & ✗ & ✗ & EM, F1 & Multi-hop QA combining structured and unstructured data with reasoning paths. & 2020 \\

\textbf{NQ} \newline \citep{kwiatkowski2019natural} & text & Retrieval, \newline Compression & QA & ✗ & ✗ & EM, F1 & Open-domain QA based on real Google search queries. & 2019 \\

\textbf{HotpotQA} \newline \citep{yang2018hotpotqa} & text & Retrieval,\newline Compression & QA & ✗ & ✗ & EM, F1 & Multi-hop QA with explainable reasoning and supporting facts at the sentence level. & 2018 \\
\bottomrule
\end{tabular}
}
\caption{Datasets used for evaluating \textbf{long-term memory}. “Mo” = modality; “Ops” = supported operations; “DS Type” = dataset type (QA – question answering, MS – multi-session dialogue); “Per” and “TR” = presence of persona and temporal reasoning.}
\label{tab:longterm_dataset}
\end{table*}

\definecolor{LC}{HTML}{DEF1D3}

\begin{table*}[htbp]
\centering
\arrayrulecolor{black}
\resizebox{\textwidth}{!}{
\begin{tabular}{
    >{\arraybackslash}m{3cm}
    >{\arraybackslash}m{2cm}
    >{\arraybackslash}m{2cm}
    >{\arraybackslash}m{2.5cm}
    >{\arraybackslash}m{6.5cm}
    >{\arraybackslash}m{1.3cm}
}
\toprule
\textbf{Datasets} & \textbf{Modality} & \textbf{Operations} & \textbf{Metrics} & \textbf{Purpose} & \textbf{Year} \\
\midrule
\textbf{LongBioBench} \newline \citep{yang2025a} & text & retrieval & EM & Controllable long-context multi-hop examination for LLMs & 2025 \\

\textbf{LongBench v2} \newline \citep{bai2025longbenchv2deeperunderstanding} & text + table + KG & compression, retrieval & Accuracy & Updated version of LongBench, longer and more challenging, with a consistent multi-choice format for reliable evaluation. & 2024 \\

\textbf{SWE-bench \newline Multimodal} \newline \citep{yang2025swebench} & text + image & compression, retrieval & Resolution rate, \newline Inference cost & Extends SWE-bench with multimodal tasks (517 instances) to test reasoning with images. & 2024 \\

\textbf{$\infty$Bench} \newline \citep{zhang-etal-2024-bench} & text & compression, retrieval & F1, Accuracy, ROUGE-L-Sum & Benchmark with 12 tasks targeting extreme long contexts (average >100K tokens). & 2024 \\

\textbf{L-Eval} \newline \citep{an-etal-2024-l} & text & compression, retrieval & ROUGE-L, F1, \newline GPT-4 & 20 tasks designed to evaluate long-context LLMs from various perspectives. & 2023 \\

\textbf{LongBench} \newline \citep{bai-etal-2024-longbench} & text & compression, retrieval & F1, ROUGE-L, \newline Accuracy, EM,\newline Edit Sim & Includes 14 English tasks, 5 Chinese tasks, and 2 code tasks for systematic long-context evaluation. & 2023 \\

\textbf{SWE-bench} \newline \citep{jimenez2024swebench} & text & compression, retrieval & Resolution rate (\% Resolved) & Tests LLMs on solving GitHub issues (2,294 instances), requiring reasoning over large codebases. & 2023 \\

\textbf{NIAH} \newline
\citep{niah} & text & retrieval & Recall Accuracy & Evaluating LLMs on finding specific data (needle) in long contexts (haystack). & 2023 \\

\textbf{LooGLE} \newline \citep{li-etal-2024-loogle} & text & compression, retrieval & BLEU-1, BLEU-4,\newline  ROUGE-1, ROUGE-4, \newline ROUGE-L, Meteor, \newline Bert, GPT-4 scores & Benchmark with 7 tasks for extreme long context (each doc >24K tokens). & 2023 \\

\textbf{GovReport} \newline \citep{huang-etal-2021-efficient} & text & compression & ROUGE-1, ROUGE-2,\newline ROUGE-L, Bert Score & Government research reports for evaluating long-document summarization. & 2021 \\

\textbf{MusiQue} \newline \citep{trivedi-etal-2022-musique} & text & retrieval & F1 & Multi-hop QA benchmark for reasoning and long-context QA tasks. & 2021 \\

\textbf{LRA} \newline \citep{tay2021long} & text + image & compression, retrieval & Accuracy & Six tasks designed to evaluate efficient long-context models. & 2020 \\

\textbf{NaturalQuestions} \newline \citep{kwiatkowski2019natural} & text & retrieval & EM, F1 & QA dataset used for retrieval and reasoning tasks with long contexts. & 2019 \\

\textbf{PG-19} \newline \citep{Rae2020Compressive} & text & compression & PPL & Book corpus from Project Gutenberg for long-context language modeling. & 2019 \\

\textbf{TriviaQA} \newline \citep{joshi-etal-2017-triviaqa} & text & retrieval & EM, F1 & QA dataset suitable for testing long-context understanding and reasoning. & 2017 \\

\textbf{NarrativeQA} \newline \citep{kocisky-etal-2018-narrativeqa} & text & retrieval & BLEU-1, BLEU-4, \newline Meteor, ROUGE-L, \newline MRR & QA dataset for comprehension over narrative texts with long contexts. & 2017 \\

\textbf{CNN/DailyMail} \newline \citep{nallapati-etal-2016-abstractive} & text & compression & ROUGE-1, ROUGE-2, \newline ROUGE-L & News articles for summarization tasks, suitable for long-context evaluation. & 2016 \\

\textbf{WikiText-103} \newline \citep{merity2017pointer} & text & compression & PPL & Wikipedia-based corpus of 100M tokens for long-context language modeling. & 2016 \\
\bottomrule
\end{tabular}
}

\caption{Datasets for \textbf{long-context memory} evaluation, sorted from newest to oldest.}
\label{tab:longcontext_dataset}
\end{table*}

\definecolor{Para}{HTML}{D1ECF9}

\begin{table*}[t!]
\centering
\arrayrulecolor{black}
\resizebox{\textwidth}{!}{
\begin{tabular}{
    >{\raggedright\arraybackslash}m{3cm}
    >{\raggedright\arraybackslash}m{2cm}
    >{\raggedright\arraybackslash}m{2cm}
    >{\raggedright\arraybackslash}m{3cm}
    >{\raggedright\arraybackslash}m{6cm}
    >{\raggedright\arraybackslash}m{1cm}
}
\toprule
 \textbf{Dataset} 
 & \textbf{Modality} 
 & \textbf{Operations}
 & \textbf{Metrics}
 & \textbf{Purpose} 
 & \textbf{Year} \\
\toprule

\textbf{KnowEdit}\newline\cite{zhang2024comprehensive} 
 & text
 & updating
 & Edit Success, Portability, Locality, Fluency
 & Provides a comprehensive benchmark of \textbf{six datasets} for evaluating knowledge insertion, modification, and erasure.
 & 2024 \\

\textbf{MUSE}\newline\cite{shi2024muse}
 & text
 & forgetting
 & VerbMem, KnowMem, PrivLeak
 & A benchmark for machine unlearning, assessing \textbf{six key properties} for unlearned models.
 & 2024 \\

\textbf{KnowUnDo}\newline\cite{tian2024forget}
 & text
 & forgetting
 & Unlearn Success, Retention Success, Perplexity, ROUGE-L
 & Evaluates unlearning in domains with \textbf{copyrighted content and user privacy}, checking if essential knowledge is inadvertently erased.
 & 2024 \\

\textbf{RWKU}\newline\cite{jin2024rwku}
 & text 
 & forgetting
 & ROUGE-L 
 & Tests real-world unlearning under \textbf{corpus-free conditions} with adversarial assessments.
 & 2024 \\

\textbf{WMDP}\newline\cite{li2024wmdp}
 & text
 & forgetting
 & QA Accuracy
 & Benchmarks hazardous knowledge detection and unlearning for \textbf{biosecurity, cybersecurity, and chemical security}.
 & 2024 \\

\textbf{TOFU}\newline\cite{maini2024tofu}
 & text 
 & forgetting
 & Probability, ROUGE, Truth Ratio
 & A dataset of facts about 200 \textbf{fictitious authors} for unlearning research.
 & 2024 \\

\textbf{ABSA}\newline\cite{ding-etal-2024-boosting}
 & text 
 & consolidation
 & F1
 & For aspect-based sentiment analysis to assess LLMs in \textbf{continual learning} tasks.
 & 2024 \\

\textbf{MQUAKE-CF}\newline\cite{zhong2023mquake} 
 & text 
 & updating
 & Edit-wise Success Rate, Instance-wise Accuracy, Multi-hop Accuracy
 & Evaluates \textbf{counterfactual knowledge propagation} through multi-hop reasoning (up to 4 hops).
 & 2023 \\

\textbf{MQUAKE-T}\newline\cite{zhong2023mquake} 
 & text 
 & updating
 & Edit-wise Success Rate, Instance-wise Accuracy, Multi-hop Accuracy
 & Assesses \textbf{temporal knowledge propagation} through multi-hop reasoning chains with one edit per chain.
 & 2023 \\

\textbf{Counterfact}\newline\cite{meng2022locating}  
 & text
 & updating
 & Efficacy Score, Efficacy Magnitude, Paraphrase Scores, Neighborhood Score
 & Evaluates \textbf{substantial and improbable factual changes} beyond superficial edits.
 & 2022 \\

\textbf{zsRE}\newline\cite{de2021editing}
 & text
 & updating
 & Success Rate, Retain Accuracy, Equivalence Accuracy, Performance Deterioration
 & One of the \textbf{earliest datasets} for evaluating knowledge editing.
 & 2021 \\

\textbf{SGD}\newline\cite{rastogi2020towards}
 & text 
 & consolidation
 & JGA, FWT, BWT
 & Multi-turn task-oriented dialogue for evolving user intents and continual schema updates.
 & 2020 \\

\textbf{INSPIRED}\newline\cite{hayati-etal-2020-inspired}
 & text 
 & consolidation
 & JGA, FWT, BWT
 & Dialogue dataset supporting incremental intent changes in task-oriented settings.
 & 2020 \\

\textbf{Natural Questions}\newline\cite{kwiatkowski2019natural}
 & text 
 & consolidation
 & Indexing Accuracy, Hits@1
 & Multi-purpose QA dataset with indexed documents for \textbf{dynamic continual learning}.
 & 2019 \\
\bottomrule
\end{tabular}
}
\caption{Datasets for parametric memory evaluation, sorted by year from newest to oldest.}
\label{tab:parametric-dataset}
\end{table*}

\begin{table*}[htbp]
\centering
\arrayrulecolor{black}
\resizebox{\textwidth}{!}{
\begin{tabular}{
    >{\arraybackslash}m{3.5cm}  
    >{\arraybackslash}m{1cm}  
    >{\arraybackslash}m{2cm}  
    >{\arraybackslash}m{1cm}  
    >{\arraybackslash}m{1cm}  
    >{\arraybackslash}m{1cm}  
    >{\arraybackslash}m{2cm}  
    >{\arraybackslash}m{5cm}  
    >{\arraybackslash}m{1cm}  
}
\hline
\textbf{\rule{0pt}{1.2em}Datasets} & \textbf{Mo} & \textbf{Ops} & \textbf{Src\#} & \textbf{Mod\#} & \textbf{Task} & \textbf{Metrics} & \textbf{Purpose} & \textbf{Year} \\
\hline

\textbf{MultiChat} \newline \citep{wang2025new} 
& text + image & Retrieval & 2 & 2 & Retrieval 
& Precision, mAP, GPT-4
& Image-grounded sticker retrieval with cross-session image-text dialogue context. 
& 2025 \\

\textbf{MovieChat-1K} \newline \citep{song2024moviechat} 
& text + video & Retrieval & 2 & 2 & QA 
& Accuracy
& Long-term video understanding for large multimodal models in video QA and captioning tasks.
& 2025 \\

\textbf{Context-conflicting} \newline \citep{tan-etal-2024-blinded} 
& text & Compression & 2 & 1 & Conflict 
& DiffGR, EM, Similarity
& Evaluate models' ability to handle conflicting evidence across sources. 
& 2024 \\

\textbf{EgoSchema} \newline \citep{mangalam2023egoschema} 
& video + text & Retrieval,\newline Compression & 3 & 2 & Fusion 
& Accuracy
& Combines episodic video memory, social schema, and conversation for long-term QA. 
& 2023 \\

\textbf{Ego4D NLQ} \newline \citep{hou2023groundnlq} 
& video + text & Retrieval,\newline Compression & 2 & 2 & Fusion 
& Recall@K
& QA over egocentric video with temporal memory using natural language queries. 
& 2022 \\

\textbf{2WikiMultihopQA} \newline \citep{ho2020constructing} 
& text & Indexing, \newline Retrieval, \newline Compression & 2 & 1 & Reasoning 
& EM, F1
& Multi-hop QA requiring reasoning across two Wikipedia passages with sentence-level evidence.
& 2020 \\

\textbf{HybridQA} \newline \citep{chen2021hybridqa} 
& text & Retrieval, \newline Compression & 2 & 1 & Reasoning 
& EM, F1
& QA requiring reasoning across structured tables and unstructured text. 
& 2020 \\

\textbf{CommonsenseVQA} \newline \citep{talmor-etal-2019-commonsenseqa} 
& text + image & Retrieval, \newline Compression & 2 & 2 & Fusion 
& Accuracy
& Commonsense QA over visual scenes requiring visual-text fusion. 
& 2019 \\

\textbf{NaturalQuestions} \newline \citep{kwiatkowski2019natural} 
& text & Retrieval, \newline Compression & >1* & 1 & Conflict 
& EM, F1
& Real-world QA over Google search snippets; also used for contradiction analysis. 
& 2019 \\

\textbf{ComplexWebQuestions} \newline \citep{talmor2018web} 
& text & Retrieval, \newline Compression & >1* & 1 & Reasoning 
& EM, F1
& Compositional QA requiring multi-step reasoning across web snippets. 
& 2018 \\

\textbf{HotpotQA} \newline \citep{yang2018hotpotqa} 
& text & Retrieval, \newline Compression & 2 & 1 & Conflict 
& EM, F1, Supporting Fact Accuracy
& Multi-hop QA with paragraph-level sources and sentence-level supporting facts. 
& 2018 \\

\textbf{TriviaQA} \newline \citep{joshi-etal-2017-triviaqa} 
& text & Retrieval, \newline Compression & $\geq$6 & 1 & Conflict 
& EM, F1
& Trivia-style QA with noisy web sources; used for source disagreement studies. 
& 2017 \\

\textbf{WebQuestionsSP} \newline \citep{yih2016value} 
& text & Indexing, \newline Retrieval, \newline Compression & >1* & 1 & Reasoning 
& F1, Accuracy
& Enhanced WebQuestions with structured reasoning chains. 
& 2016 \\

\textbf{Flickr30K} \newline \citep{young2014image} 
& text + image & Retrieval, \newline Compression & 2 & 2 & Retrieval 
& Similarity
& Image-caption pairs widely used for cross-modal retrieval and alignment tasks. 
& 2014 \\
\hline
\end{tabular}
}
\caption{Datasets used for evaluating \textbf{multi-source memory}. “Mo” denotes data modality. “Ops” indicates operations. “Src\#” = number of information sources per instance; “Mod\#” = number of modalities; “Task” = retrieval, fusion, reasoning, or conflict resolution.}
\label{tab:multi_source_datasets}
\end{table*}

\begin{table*}[htbp]
\centering
\arrayrulecolor{black}
\resizebox{\textwidth}{!}{
\begin{tabular}{
    >{\raggedright\arraybackslash}m{2cm}
    >{\raggedright\arraybackslash}m{2.5cm}
    >{\raggedright\arraybackslash}m{0.8cm}
    >{\raggedright\arraybackslash}m{0.8cm}
    >{\raggedright\arraybackslash}m{0.8cm}
    >{\raggedright\arraybackslash}m{2cm}
    >{\raggedright\arraybackslash}m{1.5cm}
    >{\raggedright\arraybackslash}m{2.5cm}
    >{\raggedright\arraybackslash}m{2.5cm}
    >{\raggedright\arraybackslash}m{4cm}
    >{\raggedright\arraybackslash}m{1cm}
}
\toprule
 \textbf{Method} & \textbf{Type} & \textbf{TF} & \textbf{RE} & \textbf{DS} & \textbf{Input} & \textbf{Output} & \textbf{LMs} & \textbf{Ops} & \textbf{Features} & \textbf{Year} \\
\toprule
\textbf{EWE}\newline\cite{chen2024improving} & Memory Grounded Generation & ✓ & ✓ & ✓ & Context & Response & Llama-3.1-70B, 8B & Updating, Retrieval & Explicit working memory, online fact-checking feedback, factual long-form generation & 2025 \\

\textbf{ICAL}\newline\cite{sarch2024vlm} & Generation & ✓ & ✓ & ✓ & Examples + Task Instruction & Trajectory + Thoughts & GPT4V, Qwen2VL & Updating & Trajectory abstraction memory, multi-modal, iterative reasoning correction & 2025 \\

\textbf{MEMORAG}\newline\cite{qian2024memorag} & Memory Grounded Generation & ✓ & ✓ & ✓ & Context + Query & Response & Mistral7B-Instruct, Phi-3-mini-128K-instruct, GPT-4o & Retrieval, Compression & Global memory retrieval, KV memory compression, Feedback-guided generation & 2024 \\

\textbf{ReadAgent}\newline\cite{pmlr-v235-lee24c} & Generation & ✓ & ✓ & ✓ & Context + Query & Retrieved Passages/Summary & PaLM 2 & Updating, Retrieval & Episodic gist memory, dynamic memory retrieval, extended context window & 2024 \\

\textbf{FLOW-RAG}\newline\cite{wang2024machine} & Updating & ✓ & ✓ & ✓ & Knowledge Base + Query & Response & GPT4o, Gemini, llama2-7B-chat & Forgetting & RAG-based unlearning & 2024 \\

\textbf{HippoRAG}\newline\cite{gutierrez2024hipporag} & Retrieval & ✓ & ✓ & ✓ & Context + Query & Response & ColBERTv2, GPT-3.5-turbo, Llama-3.1-8B, 70B & Indexing & Hippocampal-inspired retrieval, multi-hop QA, Knowledge graph integration & 2024 \\

\textbf{IterCQR}\newline\cite{jang-etal-2024-itercqr} & Retrieval & ✓ & ✓ & ✓ & Dialogue History + Query & Retrieved Results & Transformer++ & Retrieval & Iterative query reformulation, context-aware query rewriting & 2024 \\

\textbf{MemoryBank}\newline\cite{zhong2024memorybank} & Consolidation & ✓ & ✓ & ✓ & Retrieved \& Context + Query & Response & ChatGLM-6B, BELLE-7B, gpt-3.5-turbo & Consolidation, Updating, Forgetting, Retrieval & Fine-tuning, RAG, Ebbinghaus Forgetting & 2024 \\

\textbf{FLARE}\newline\cite{jiang-etal-2023-active} & Retrieval & ✓ & ✓ & ✓ & Database + Query & Response & WebGPT, WebCPM & Retrieval & Active retrieval during generation, forward-looking query prediction & 2023 \\

\textbf{MemoChat}\newline\cite{lu2023memochat} & Consolidation & ✓ & ✓ & ✓ & Dialogue History + Query & Response & GPT4, ChatGPT, Vicuna-7B, 13B, 33B, T5 & Consolidation, Retrieval & Structured memos, memory-driven dialogue, memorization–retrieval–response cycle & 2023 \\

\textbf{NLI-Transfer}\newline\cite{bae2022keep} & Updating & ✓ & ✓ & ✓ & Memory + Dialogue History & Response & T5 & Consolidation, Updating, Retrieval & Session-level memory tracking, evolving dialogue system & 2022 \\
\bottomrule
\end{tabular}
}
\caption{Overview of methods for \textbf{long-term memory in memory management and utilization}. ``TF'' (Training Free) denotes whether the method operates without additional gradient-based updates; ``RE'' (Retrieval Module) denotes whether the method uses retrieval; ``DS'' (Dialogue System) denotes whether the method is designed for dialogue tasks.}
\label{tab:longterm-mag}
\end{table*}

\begin{table*}[t!]
\centering
\arrayrulecolor{black}
\resizebox{\textwidth}{!}{
\begin{tabular}{
    >{\raggedright\arraybackslash}m{2cm}
    >{\raggedright\arraybackslash}m{2cm}
    >{\raggedright\arraybackslash}m{1cm}
    >{\raggedright\arraybackslash}m{1cm}
    >{\raggedright\arraybackslash}m{2cm}
    >{\raggedright\arraybackslash}m{1.5cm}
    >{\raggedright\arraybackslash}m{2.5cm}
    >{\raggedright\arraybackslash}m{2cm}
    >{\raggedright\arraybackslash}m{4cm}
    >{\raggedright\arraybackslash}m{1cm}
}
\toprule
 \textbf{Method} & \textbf{Type} & \textbf{TF} & \textbf{RE} & \textbf{Input} & \textbf{Output} & \textbf{LMs} & \textbf{Ops} & \textbf{Features}  & \textbf{Year} \\
\toprule

\textbf{LD-Agent}\newline\cite{li2024hello} & Augmentation & \usym{2713} & \usym{2713} & Retrieved \& Context + Query & Response & ChatGLM, BlenderBot, ChatGPT & Consolidation, Updating, Retrieval & long-term dialogue modeling, event \& persona memory, mudular agent architecture & 2025 \\

\textbf{SiliconFriend}\newline\cite{zhong2024memorybank} & Augmentation & \usym{2717} & \usym{2713} & Retrieved \& Context + Query & Response & ChatGLM-6B, \newline BELLE-7B, \newline gpt-3.5-turbo & Consolidation, Updating, Forgetting, Retrieval & fine-tuning, \newline RAG, Ebbinghaus Forgetting & 2024 \\

\textbf{MALP}\newline\cite{zhang-etal-2024-llm-based} & Adaption & \usym{2717} & \usym{2713} & Retrieved \& Context + Query & Response & GPT3.5, LLaMA-7B, LLaMA-13B & Consolidation, Retrieval & memory coordination, computational bionic memory mechanism, patient profile, self-chat & 2024 \\

\textbf{PERPCS}\newline\cite{tan2024personalized} & Adaption & \usym{2717} & \usym{2717} & User History & / & Llama-2-7B & Consolidation & modular PEFT sharing, collaborative personalization, user history assembly & 2024 \\

\textbf{LAPDOG}\newline\cite{huang-etal-2023-learning} & Augmentation & \usym{2713} & \usym{2713} & Retrieved \& Context + Query & Response & T5 & Consolidation, Updating, Retrieval & Story-based persona retrieval, joint retriever-generator training & 2024 \\ 

\textbf{RECAP}\newline\cite{liu-etal-2023-recap} & Augmentation & \usym{2717} & \usym{2713} & Retrieved \& Context + Query & Response & Transformers & Retrieval & hierarchical transformer retriever, context-aware prefix encoder & 2023 \\

\textbf{CLV}\newline\cite{tang-etal-2023-enhancing-personsiri2025alized} & Adaption & \usym{2717} & \usym{2717} & Persona + Query & Response & GPT-2 & Consolidation & contrastive learning, \newline clustered dense persona, \newline dialogue generation & 2023 \\

\textbf{PERKGQA}\newline\cite{dutt2022perkgqa} & Augmentation & \usym{2713} & \usym{2713} & Retrieved \& Knowledge Graph + Query & Response & RoBERTa & Retrieval & long-term dialogue modeling, event \& persona memory, mudular agent architecture & 2022 \\

\bottomrule
\end{tabular}
}
\caption{Overview of methods for \textbf{long-term memory in personalization}. "TF" (Training Free) denotes whether the method operates without additional gradient-based updates. 
"RE" (Retrieval Module) denotes whether the method needs Retrieval.}
\label{tab:longterm-persona}
\end{table*}

\begin{table*}[t!]
\centering
\arrayrulecolor{black}
\resizebox{\textwidth}{!}{
\begin{tabular}{
    >{\raggedright\arraybackslash}m{2.5cm}
    >{\raggedright\arraybackslash}m{2cm}
    >{\raggedright\arraybackslash}m{1cm}
    >{\raggedright\arraybackslash}m{1cm}
    >{\raggedright\arraybackslash}m{2cm}
    >{\raggedright\arraybackslash}m{4cm}
    >{\raggedright\arraybackslash}m{5cm}
    >{\raggedright\arraybackslash}m{1cm}
}
\toprule
\textbf{Method} & \textbf{Type} & \textbf{TF} & \textbf{FC} & \textbf{Operations} & \textbf{LMs} & \textbf{Features} & \textbf{Year} \\
\toprule

\textbf{FoldGRPO} \newline \citep{sun2025scalinglonghorizonllmagent} & Context Compression & \usym{2717} & \usym{2717} & Compression & Seed-OSS-36B & RL framework training agents to actively fold detailed histories into concise summaries & 2025 \\

\textbf{RECOMP} \newline \citep{xu2024recomp} 
& Context Compression 
& \usym{2717}  & \usym{2717}
& Compression 
& GPT-2, GPT2-XL, GPT-J, Flan-UL2 
& Hard prompt compression with extractive and abstractive strategies 
& 2024 \\

\textbf{LongLLMLingua} \newline \citep{jiang-etal-2024-longllmlingua} 
& Context Compression 
& \usym{2713}  & \usym{2717} 
& Compression 
& GPT-3.5-Turbo-06136, LongChat-13B-16k 
& Hard prompt compression for efficient input representation 
& 2024 \\

\textbf{LLMLingua-2} \newline \citep{pan-etal-2024-llmlingua} 
& Context Compression 
& \usym{2717}  & \usym{2717} 
& Compression 
& XLM-RoBERTa-Large, Multilingual-BERT 
& Hard prompt compression with data distillation for multilingual contexts 
& 2024 \\

\textbf{QGC} \newline \citep{cao-etal-2024-retaining} 
& Context Compression 
& \usym{2717}  & \usym{2717} 
& Compression 
& LongChat-13B-16K, LLaMA-2-7B 
& Query-guided dynamic context compression 
& 2024 \\

\textbf{xRAG} \newline \citep{cheng2024xrag} 
& Context Compression 
& \usym{2717}  & \usym{2717} 
& Compression 
& Mistral-7B, Mixtral-8x7B 
& Soft prompt compression for parametric integration 
& 2024 \\ 

\textbf{AutoCompressor} \newline \citep{chevalier-etal-2023-adapting} 
& Context Compression 
& \usym{2717}  & \usym{2713} 
& Compression 
& OPT-1.3B, OPT-2.7B, LLaMA-2-7B 
& Soft prompt compression for general-purpose efficiency 
& 2023 \\

 \textbf{StreamingLLM} \newline \citep{xiao2024efficient} 
 & KV Cache Eviction 
 & \usym{2713} 
 & \usym{2717} 
 & Compression 
 & Llama-2, MPT, PyThia, Falcon 
 & Static KV cache eviction, Attention sink in the initial tokens 
 & 2024 \\

 \textbf{FastGen} \newline \citep{ge2024model} 
 & KV Cache Eviction 
 & \usym{2713} 
 & \usym{2717} 
 & Compression 
 & Llama-1 7B/13B/30B/65B 
 & Adaptive profiling-based KV cache eviction 
 & 2024 \\

 \textbf{SnapKV} \newline \citep{NEURIPS2024_28ab4182} 
 & KV Cache Eviction 
 & \usym{2713} 
 & \usym{2717} 
 & Compression 
 & LWM-Text-Chat-1M, LongChat-7b-v1.5-32k, Mistral-7B-Instruct-v0.2, Mixtral-8x7B-Instruct-v0.1 
 & Head-wise KV cache eviction, Attention head behavior 
 & 2024 \\

 \textbf{LESS} \newline \citep{pmlr-v235-dong24f} 
 & KV Cache Storing Optimization 
 & \usym{2717} 
 & \usym{2713} 
 & Compression 
 & Llama-2 13B, Falcon 7B 
 & Low-rank KV cache storage enabling querying of all tokens 
 & 2024 \\

 \textbf{KIVI} \newline \citep{pmlr-v235-liu24bz} 
 & KV Cache Storing Optimization 
 & \usym{2713} 
 & \usym{2713} 
 & Compression 
 & Llama-2 7B/13B, Llama-3 8B, Falcon 7B, Mistral-7B 
 & Asymmetrical KV cache quantization 
 & 2024 \\

 \textbf{KVQuant} \newline \citep{NEURIPS2024_028fcbcf} 
 & KV Cache Storing Optimization 
 & \usym{2713} 
 & \usym{2713} 
 & Compression 
 & LLaMA-7B/13B/30B/65B, Llama-2-7B/13B/70B, Llama-3-8B/70B, Mistral-7B 
 & KV cache quantization 
 & 2024 \\

 \textbf{H$_2$O} \newline \citep{zhang2023ho} 
 & KV Cache Eviction 
 & \usym{2713} 
 & \usym{2717} 
 & Compression 
 & OPT, Llama-1, GPT-NeoX 
 & Dynamic KV cache eviction, Retain Heavy Hitter tokens 
 & 2023 \\

 \textbf{Scissorhands} \newline \citep{liu2023scissorhands} 
 & KV Cache Eviction 
 & \usym{2713} 
 & \usym{2717} 
 & Compression 
 & OPT 6.7B, 13B, 30B, 66B 
 & Dynamic KV cache eviction, Persistence of importance hypothesis 
 & 2023 \\
 
 \textbf{ShadowKV} \newline \citep{sun2025shadowkv} 
 & KV Cache Storing Optimization
 & \usym{2713} 
 & \usym{2713} & Compression & Llama-3.1-8B, Llama-3-8B-1M, GLM-4-9B-1M, Yi-9B-200K, Phi-3-Mini-128K, and Qwen2-7B-128K & Reduces GPU memory usage by keeping compressed shadow versions of keys on the GPU while offloading full data to the CPU & 2025\\

 \textbf{FlexGen} \newline \citep{10.5555/3618408.3619696} 
 & KV Cache Storing Optimization 
 & \usym{2713} 
 & \usym{2713} 
 & Compression 
 & OPT 6.7B to 175B 
 & KV cache quantization and offloading 
 & 2023 \\

\bottomrule
\end{tabular}
}
\caption{Overview of methods for \textbf{long-context memory compression}. ``TF'' (Training Free) indicates no additional gradient-based updates. ``FC'' (Full Context) indicates that the method preserves the access to all context tokens.}
\label{tab:Cha-longcontexteffectiveness}
\end{table*}

\begin{table*}[t!]
\centering
\arrayrulecolor{black}
\resizebox{\textwidth}{!}{
\begin{tabular}{
    >{\raggedright\arraybackslash}m{2cm}
    >{\raggedright\arraybackslash}m{2cm}
    >{\raggedright\arraybackslash}m{1cm}
    >{\raggedright\arraybackslash}m{1cm}
    >{\raggedright\arraybackslash}m{1.5cm}
    >{\raggedright\arraybackslash}m{4cm}
    >{\raggedright\arraybackslash}m{5cm}
    >{\raggedright\arraybackslash}m{1cm}
}
\toprule
\textbf{Method} & \textbf{Type} & \textbf{TF} & \textbf{FC} & \textbf{Operations} & \textbf{LMs} & \textbf{Features} & \textbf{Year} \\
\toprule
\textbf{Sparse RAG} \newline \citep{zhu2025accelerating} 
& Context Selection 
& \usym{2717} 
& \usym{2717} 
& Retrieval 
& Gemini 
& Sparse context selection, reduces the number of documents involved during decoding 
& 2025 \\

\textbf{GraphReader} \newline \citep{li-etal-2024-graphreader} 
& Context Selection 
& \usym{2713} 
& \usym{2717} 
& Retrieval 
& GPT-4-128k 
& Graph-based agent; structures long context into a graph 
& 2024 \\

\textbf{Ziya-Reader} \newline \citep{he-etal-2024-never} 
& Context Selection 
& \usym{2717} 
& \usym{2713} 
& Retrieval 
& Ziya2-13B-Base (LLaMA-2-13B) 
& Supervised fine-tuning; position-agnostic multi-step QA 
& 2024 \\

\textbf{FILM} \newline \citep{NEURIPS2024_71c3451f} 
& Context Selection 
& \usym{2717} 
& \usym{2713} 
& Retrieval 
& FILM-7B (Mistral 7B) 
& Data-driven approach; addresses the “lost in the middle” problem 
& 2024 \\

 \textbf{TokenSelect} \newline \citep{wu2025tokenselectefficientlongcontextinference} 
 & KV Cache Selection 
 & \usym{2713} 
 & \usym{2713} 
 & Retrieval 
 & Qwen2 7B, Llama-3 8B, Yi-1.5-6B 
 & Dynamic token-level KV cache selection 
 & 2025 \\

 \textbf{QUEST} \newline \citep{pmlr-v235-tang24l} 
 & KV Cache Selection 
 & \usym{2713} 
 & \usym{2713} 
 & Retrieval 
 & LongChat-7B-v1.5-32K, Yarn-Llama2-7B-128K 
 & Query-aware KV cache selection 
 & 2024 \\
 
 \textbf{Memorizing Transformers} \newline \cite{wu2022memorizing} 
 & KV Cache Selection 
 & \usym{2717} 
 & \usym{2713} 
 & Retrieval 
 & Transformers 
 & External KV cache memory for enhanced recall 
 & 2022 \\
\bottomrule
\end{tabular}
}
\caption{Overview of methods for \textbf{long-context memory retrieval}. ``TF'' (Training Free) indicates no additional gradient-based updates. ``FC'' (Full Context) indicates that the method preserves the access to all context tokens.}
\label{tab:Cha-longcontexteffectiveness}
\end{table*}

\definecolor{Para}{HTML}{D1ECF9}

\begin{table*}[t!]
\centering
\arrayrulecolor{black}
\resizebox{\textwidth}{!}{
\begin{tabular}{
>{\raggedright\arraybackslash}m{2cm}
>{\raggedright\arraybackslash}m{2.5cm}
>{\raggedright\arraybackslash}m{0.8cm}
>{\raggedright\arraybackslash}m{0.8cm}
>{\raggedright\arraybackslash}m{0.8cm}
>{\raggedright\arraybackslash}m{0.8cm}
>{\raggedright\arraybackslash}m{2cm}
>{\raggedright\arraybackslash}m{6cm}
>{\raggedright\arraybackslash}m{1cm}
}
\toprule
 \textbf{Method} 
& \textbf{Type} 
& \textbf{PR} 
& \textbf{TF} 
& \textbf{BES} 
& \textbf{SEO} 
& \textbf{LMs} 
& \textbf{Main Advancement}
& \textbf{Year} \\
\toprule

\textbf{AlphaEdit}\newline\cite{fang2025alphaedit}
& locating-then-editing
& \usym{2717} 
& \usym{2713} 
& \usym{2713} 
& \usym{2713} 
& gpt2-xl-1.5b,\newline gpt-j-6b,\newline llama3-8b
& Protect preserved knowledge by \textbf{projecting perturbation onto the null space} and adding a \textbf{regularization term} for \textbf{sequential} editing.
& 2024 \\

\textbf{MEMAT}\newline\cite{mela2024mass}
& locating-then-editing
& \usym{2717} 
& \usym{2713} 
& \usym{2713} 
& \usym{2717}
& aguila-7b
& Extension of MEMIT with \textbf{attention head corrections} for \textbf{cross-lingual} editing.
& 2024 \\

\textbf{DEM}\newline\cite{huang2024commonsense}
& locating-then-editing
& \usym{2717} 
& \usym{2713} 
& \usym{2713}
& \usym{2717}
& gpt-j-6b,\newline llama2-7b
& Uses a \textbf{dynamic aware module} to select editing layers, targeting commonsense knowledge editing in free text.
& 2024 \\

\textbf{DAFNET}\newline\cite{zhang2024dafnet}
& meta learning
& \usym{2717} 
& \usym{2717} 
& \usym{2717} 
& \usym{2713}
& gpt-j-6b, \newline llama2-7b
& Supports \textbf{sequential editing} via \textbf{Intra-editing Attention Flow} (within facts) and \textbf{Inter-editing Attention Flow} (across facts).
& 2024 \\

\textbf{Larimar} \newline\cite{Larimar2024}
& additional parameters
& \usym{2713}
& \usym{2713}
& \usym{2713}
& \usym{2713}
& gpt2-xl, gpt-j-6b
& Introduces a \textbf{decoupled latent memory module} that conditions the LLM decoder at test time without parameter updates.
& 2024 \\

\textbf{MEMORYLLM} \newline\cite{wang2024memoryllm}
& additional parameters
& \usym{2713}
& \usym{2717}
& \usym{2713}
& \usym{2713}
& llama2-7b
& Introduces a \textbf{fixed-size memory pool} that is incrementally and selectively updated in a frozen LLM.
& 2024 \\

\textbf{WISE} \newline\cite{wang2024wise}
& additional parameters
& \usym{2713}
& \usym{2717}
& \usym{2713}
& \usym{2713}
& llama2-7b,\newline mistral-7b,\newline gpt-j-6b
& Supports sequential editing through \textbf{Side Memory Design} and \textbf{Knowledge Sharding and Merging}.
& 2024 \\

\textbf{PMET} \newline\cite{li2024pmet}
& locating-then-editing
& \usym{2717} 
& \usym{2713} 
& \usym{2713} 
& \usym{2717} 
& gpt-j-6b,\newline gpt-neox-20b
& \textbf{Jointly optimizes attention heads and FFN}, updating only FFN weights.
& 2023 \\

\textbf{IKE}\newline\cite{zheng2023can}
& prompt
& \usym{2713}
& \usym{2713}
& -
& -
& gpt-j-6b,\newline gpt2-xl-1.5b,\newline gpt-neo,\newline gpt-neox,\newline opt-175b
& First method to use \textbf{in-context learning (ICL)} for LLM editing.
& 2023 \\

\textbf{MeLLo} \newline\cite{zhong2023mquake}
& prompt
& \usym{2713}
& \usym{2713}
& -
& -
& vicuna-7b,\newline gpt-j-6b
& Combines \textbf{Question Decomposition + Self Check} for editing.
& 2023 \\

\textbf{MALMEN}\newline\cite{tan2024massive}
& meta learning
& \usym{2717}
& \usym{2717}
& \usym{2713}
& \usym{2717}
& bert-base,\newline gpt-2,\newline t5-xl,\newline gpt-j-6b
& Uses least squares to merge edits reliably and decouples networks to save memory, supporting \textbf{massive batch} editing.
& 2023 \\

\textbf{MEMIT} \newline\cite{meng2023massediting}
& locating-then-editing
& \usym{2717} 
& \usym{2713} 
& \usym{2713} 
& \usym{2717} 
& gpt-j-6b,\newline gpt-neox-20b
& Optimizes a relaxed least-squares objective, enabling a \textbf{closed-form solution} for \textbf{massive batch} editing.
& 2022 \\

\textbf{ROME}\newline\cite{meng2022locating}
& locating-then-editing
& \usym{2717} 
& \usym{2713} 
& \usym{2717} 
& \usym{2717} 
& gpt2-xl-1.5b
& A \textbf{classic locate-then-edit} method performing a \textbf{rank-one update} on a single MLP layer.
& 2022 \\

\textbf{CaliNET}\newline\cite{dong2022calibrating}
& additional parameters
& \usym{2713}
& \usym{2717}
& \usym{2713}
& \usym{2717}
& t5-base,\newline t5-large
& Adds \textbf{FFN-like calibration layers} to modify outputs efficiently.
& 2022 \\

\textbf{SERAC}\newline\cite{mitchell2022memory}
& additional parameters
& \usym{2713}
& \usym{2717}
& \usym{2713}
& \usym{2713}
& t5-large,\newline bert-base,\newline blenderbot-90m
& Combines \textbf{Scope Classifier + Counterfactual Model} to support sequential or simultaneous edits with consistent results.
& 2022 \\

\textbf{GRACE}\newline\cite{mitchell2022memory}
& additional parameters
& \usym{2713}
& \usym{2717}
& \usym{2717}
& \usym{2713}
& t5-small,\newline bert-base,\newline gpt2-xl-1.5b
& Supports sequential editing using a \textbf{codebook with a deferral mechanism}.
& 2022 \\

\textbf{MEND}\newline\cite{mitchell2022fast}
& meta learning
& \usym{2717}
& \usym{2717}
& \usym{2713}
& \usym{2717}
& gpt-neo,\newline gpt-j-6b,\newline t5-xl,\newline t5-xxl,\newline bert-base,\newline bart-base
& Decomposes gradient updates into a rank-one outer product for scalable, fast editing.
& 2021 \\

\textbf{KE}\newline\cite{de2021editing}
& meta learning
& \usym{2717}
& \usym{2717}
& \usym{2713}
& \usym{2717}
& bert-base,\newline bart-base
& First to use a \textbf{hypernetwork} to project sentence embeddings into a rank-1 gradient mask.
& 2021 \\

\bottomrule
\end{tabular}
}
\caption{Overview of methods for \textbf{parametric memory optimization in editing}. "PR" (Parametric Reserving) denotes whether model weights remain untouched. "TF" (Training-Free) indicates editing without iterative optimization. "BES" (Batch Editing Support) highlights batch editing capability. "SEO" (Sequential Editing Optimization) shows mechanisms tailored for sequential edits. "LMs" lists the language models used for experiments.}
\label{tab:methods-editing}
\end{table*}

\begin{table*}[t!]
\centering
\arrayrulecolor{black}
\resizebox{\textwidth}{!}{
\begin{tabular}{
>{\raggedright\arraybackslash}m{3cm}
>{\raggedright\arraybackslash}m{2.5cm}
>{\raggedright\arraybackslash}m{1cm}
>{\raggedright\arraybackslash}m{1cm}
>{\raggedright\arraybackslash}m{1cm}
>{\raggedright\arraybackslash}m{1cm}
>{\raggedright\arraybackslash}m{3cm}
>{\raggedright\arraybackslash}m{6cm}
>{\raggedright\arraybackslash}m{1cm}
}
\toprule
 \textbf{Method} 
& \textbf{Type} 
& \textbf{PR} 
& \textbf{TF} 
& \textbf{BUS} 
& \textbf{SUO} 
& \textbf{LMs} 
& \textbf{Main Advancement}
& \textbf{Year} \\
\toprule

\textbf{ULD} \newline\cite{ji2024reversing}
& additional parameters
& \usym{2713}
& \usym{2717}
& \usym{2713}
& \usym{2717}
& llama2-chat-7b, mistral-7b-instruct
& Derives an unlearned LLM by computing the \textbf{logit difference} between the target and assistant models.
& 2024 \\

\textbf{ECO} \newline\cite{liu2024large}
& prompt
& \usym{2713}
& \usym{2717}
& \usym{2713}
& \usym{2717}
& 68 LLMs ranging from 0.5B to 236B
& Performs unlearning by \textbf{corrupting prompt embeddings} detected by a classifier, without altering model weights.
& 2024 \\

\textbf{WAGLE} \newline \cite{jia2024wagle}
& locating-then-unlearning
& \usym{2717}
& \usym{2717}
& \usym{2713}
& \usym{2717}
& llama2-7b-chat, zephyr-7b-beta, llama2-7b
& Uses bi-level optimization to compute weight attribution scores for \textbf{selective fine-tuning} to achieve efficient, modular unlearning.
& 2024 \\

\textbf{SOUL}\newline\cite{jia2024soul}
& training objective
& \usym{2717}
& \usym{2717}
& \usym{2713}
& \usym{2713}
& opt-1.3b, llama2-7b
& Leverages a \textbf{second-order optimizer} for more effective LLM unlearning.
& 2024 \\

\textbf{SKU} \newline\cite{liu2024towards}
& training objective
& \usym{2717}
& \usym{2717}
& \usym{2713}
& \usym{2713}
& opt-2.7b, llama2-7b, llama2-13b
& Combines harmful knowledge learning with task vector negation in a two-stage framework for robust unlearning.
& 2024 \\

\textbf{EUL}\newline\cite{chen2023unlearn}
& additional parameters
& \usym{2713}
& \usym{2717}
& \usym{2713}
& \usym{2713}
& t5-base, t5-3b
& Introduces \textbf{unlearning layers} to forget specific data, supporting sequential unlearning through a \textbf{fusion mechanism} to merge multiple layers.
& 2023 \\

\textbf{ICUL} \newline\cite{pawelczyk2024context}
& prompt
& \usym{2713}
& \usym{2713}
& -
& -
& bloom-560m, bloom-1.1b, bloom-3b, llama2-7b
& First method to leverage \textbf{in-context learning (ICL)} for unlearning in language models.
& 2023 \\

\textbf{GA+Mismatch} \newline\cite{yao2024large}
& training objective
& \usym{2717}
& \usym{2717}
& \usym{2713}
& \usym{2717}
& opt-1.3b, opt-2.7b, llama2-7b
& Pioneered LLM unlearning by blending \textbf{forgetting, random mismatch, and KL-based preservation} objectives.
& 2023 \\

\textbf{KGA} \newline\cite{wang2023kga}
& training objective
& \usym{2717}
& \usym{2717}
& \usym{2713}
& \usym{2717}
& bart-base, distil-bert, lstm
& Simulates forgetting by aligning knowledge gaps between retain and forget models via \textbf{distributional divergence minimization}.
& 2023 \\

\textbf{DEPN}\newline\cite{wu2023depn}
& locating-then-unlearning
& \usym{2713}
& \usym{2713}
& \usym{2713}
& \usym{2717}
& bert-base
& Detects and disables privacy-related neurons to reduce sensitive data leakage in language models.
& 2023 \\

\bottomrule
\end{tabular}
}
\caption{Overview of methods for \textbf{parametric memory optimization in unlearning}. "PR" (Parametric Reserving) indicates whether the method avoids direct modification of internal weights. "TF" (Training-Free) shows if the method works without iterative optimization. "BUS" (Batch Unlearning Support) marks support for multiple edits simultaneously. "SUO" (Sequential Unlearning Optimization) indicates sequential unlearning capabilities. "LMs" lists language models used for experiments.}
\label{tab:methods-unlearning}
\end{table*}

\begin{table*}[t!]
\centering
\arrayrulecolor{black}
\resizebox{\textwidth}{!}{
\begin{tabular}{
>{\raggedright\arraybackslash}m{2cm}
>{\raggedright\arraybackslash}m{2cm}
>{\raggedright\arraybackslash}m{1cm}
>{\raggedright\arraybackslash}m{1cm}
>{\raggedright\arraybackslash}m{1cm}
>{\raggedright\arraybackslash}m{1.5cm}
>{\raggedright\arraybackslash}m{2cm}
>{\raggedright\arraybackslash}m{5cm}
>{\raggedright\arraybackslash}m{1cm}
}
\toprule
 \textbf{Method} 
& \textbf{Type} 
& \textbf{TF} 
& \textbf{TB}
& \textbf{TS} 
& \textbf{Domain} 
& \textbf{LMs} 
& \textbf{Main Advancement}
& \textbf{Year} \\
\toprule

\textbf{HippoRAG 2} \newline\cite{gutierrez2025rag}
& 
& \usym{2717}
& \usym{2717}
& Task-Free
& Question Answering
& 
& Employs a training objective that \textbf{minimizes the Kullback-Leibler (KL) divergence} between the predictions of the original model and target model. 
& 2025 \\ 

\textbf{SELF-PARAM} \newline\cite{wangself}
& Regularization-based Learning
& \usym{2713}
& \usym{2713}
& Task-Free
& Question Answering
& Llama-3.3-70B-Instruct
& Enhances Personalized PageRank-based retrieval with deeper passage integration and online LLM usage, achieving superior performance on factual, associative, and sense-making memory tasks. 
& 2025 \\ 

\textbf{MBPA++} \newline\cite{wang2024towards}
& Replay-based
& \usym{2717}
& \usym{2717}
& CIL
& None
& REPLAY, MBPA
& Maintains a small, randomly selected subset (as low as 1\%) of past examples in memory to achieve performance comparable to larger memory sizes.
& 2025 \\ 

\textbf{LSCS} \newline\cite{wang2024towards}
& Interactive Learning
& \usym{2717}
& \usym{2717}
& CIL
& Abstracting/ Merging/ Retrieval
& /
& Integrates \textbf{multiple storage mechanisms} to achieve abstraction, experience merging, and long-term retention with accurate recall.
& 2025 \\ 

\textbf{TaSL} \newline\cite{feng2024tasl}
& Regularization-based Learning
& \usym{2717}
& \usym{2717}
& TIL
& Dialogue System
& T5, Llama-7B
& Parameter-level task skill localization and consolidation enabling knowledge transfer \textbf{without memory replay}.
& 2024 \\ 

\textbf{EMP} \newline\cite{liu2022incremental}
& Replay-based
& \usym{2717}
& \usym{2717}
& CLI
& Event Detection
& BERT-ED, KCN
& Designs \textbf{continuous prompts} associated with each event type.
& 2023 \\ 

\textbf{UDIL} \newline\cite{shi2023unified}
& Interactive Learning
& \usym{2717}
& \usym{2713}
& DLI
& Event Detection
& oEWC, SI, LwF, A-GEM, CLS-ER, ESM, etc.
& Introduces \textbf{adaptive coefficients} optimized during training to achieve tighter generalization error bounds and improved performance across domains.
& 2023 \\ 

\textbf{DSI++} \newline\cite{mehta2022dsi++}
& Replay-based
& \usym{2717}
& \usym{2713}
& TIL
& Information Retrieval
& T5
& Enables \textbf{continual document indexing} while retaining query performance on old and new data.
& 2022 \\ 

\textbf{MRDC} \newline\cite{wang2022memory}
& Replay-based
& \usym{2717}
& \usym{2713}
& CIL
& Object Detection
& LUCIR, PODNet
& Enhances \textbf{memory replay by compressing data}, balancing sample quality and quantity for continual learning.
& 2022 \\ 

\bottomrule
\end{tabular}
}
\caption{Overview of methods for \textbf{parametric memory modification in continual learning}. "TB" denotes whether task boundaries exist. "TS" denotes task settings including TIL (Task Incremental Learning), CIL (Class Incremental Learning), DIL (Domain Incremental Learning), and Task-Free.}
\label{tab:methods-continual_learning}
\end{table*}

\begin{table*}[t!]
\centering
\arrayrulecolor{black}
\resizebox{\textwidth}{!}{
\begin{tabular}{
    >{\raggedright\arraybackslash}m{1.5319cm}
    >{\raggedright\arraybackslash}m{1.5319cm}
    >{\raggedright\arraybackslash}m{0.3830cm}
    >{\raggedright\arraybackslash}m{1.5319cm}
    >{\raggedright\arraybackslash}m{2.2979cm}
    >{\raggedright\arraybackslash}m{1.5319cm}
    >{\raggedright\arraybackslash}m{1.1489cm}
    >{\raggedright\arraybackslash}m{1.5319cm}
    >{\raggedright\arraybackslash}m{1.9149cm}
    >{\raggedright\arraybackslash}m{3.0638cm}
    >{\raggedright\arraybackslash}m{0.76596cm}
}
\toprule
 \textbf{Method} & \textbf{Type} & \textbf{TF} & \textbf{STs} & \textbf{SNs} & \textbf{Input} & \textbf{Output} & \textbf{LMs} & \textbf{Ops} & \textbf{Features}  & \textbf{Year} \\
\toprule

\textbf{GoG} \newline\cite{xu2024generate} & reasoning & \usym{2713} & KG + text & WebQSP, CWQ & KG + prompt + query & answer & GPT-3.5, GPT-4, Qwen-1.5-72B-Chat, LLaMA3-70B-Instruct & Retrieval, Compression & Integrate internal and external knowledge & 2024 \\

\textbf{RKC-LLM} \newline\cite{wang2023resolving} & conflict & \usym{2713} & model + text & prompt + context & entities & answer & ChatGPT & Compression & Conflict span localization, instruction-guided conflict handling & 2024 \\

\textbf{BGC-KC} \newline\cite{tan-etal-2024-blinded} & conflict & \usym{2713} & model + text & AIG, AIR & documents + query & answer & GPT-4, GPT-3.5, Llama2-13b, Llama2-7b & Retrieval, Compression & Attribution tracing framework, evaluate LLM bias & 2024 \\

\textbf{Sem-CoT} \newline\cite{su2023semi} & reasoning & \usym{2717} & Knowledge Graph + text + Model & Wikidata, 2Wiki, MuSiQue, TKB & CoT prompt + query & answer & LLaMA2-7B, 13B, 70B, 65B & Retrieval, Compression & Semi-structured prompting for multi-source input fusion & 2023 \\

\textbf{CoK} \newline\cite{li2024chain} & reasoning & \usym{2717} & Database + Tables + Text & Wikidata, Wikipedia, Wikitables, Flashcard, UpToDate, ScienceQA, CK-12 & CoT prompt + query & answer & GPT-3.5-turbo & Retrieval, Compression & Heterogeneous knowledge integration, dynamic knowledge retrieval, adaptive query generation across formats & 2023 \\

\textbf{DIVKNOWQA} \newline\cite{zhao2024divknowqa} & reasoning & \usym{2717} & Knowledge Base + text & Wikidata, DIVKNOWQA & CoT prompt + query & answer & GPT-3.5-turbo & Retrieval, Compression & Two-hop reasoning, symbolic query generation for structured data & 2023 \\

\textbf{StructRAG} \newline\cite{li2024structrag} & reasoning & \usym{2717} & KG + Table + text & Loong, Podcast Transcripts & documents + query & answer & Qwen2-7B, 72B & Retrieval, Compression & Cognitive-inspired structurization, dynamic structure selection & 2023 \\

\bottomrule
\end{tabular}
}
\caption{Overview of methods for \textbf{multi-source memory in cross-textual integration}. "TF" (Training Free) denotes whether the method operates without additional gradient-based updates. 
"STs" denotes the source types. "SNs" denotes the source dataset names.}
\label{tab:Cha-cross_text_multi_source}
\end{table*}

\begin{table*}[t!]
\centering
\arrayrulecolor{black}
\resizebox{\textwidth}{!}{
\begin{tabular}{
    >{\raggedright\arraybackslash}m{1.8cm}
    >{\raggedright\arraybackslash}m{1.6cm}
    >{\raggedright\arraybackslash}m{1.6cm}
    >{\raggedright\arraybackslash}m{1.8368cm}
    >{\raggedright\arraybackslash}m{3.4082cm}
    >{\raggedright\arraybackslash}m{2.2041cm}
    >{\raggedright\arraybackslash}m{4cm}
    >{\raggedright\arraybackslash}m{1.5cm}
}
\toprule
 \textbf{Memory Tool} & \textbf{Level} & \textbf{Taxonomy} & \textbf{Operation} & \textbf{Function} & \textbf{Input/Output} & \textbf{Example Use} & \textbf{Source Type} \\
\toprule

\textbf{FAISS} \newline \citep{douze2024faiss} & Components & Contextual-Unstructured & Consolidation, Indexing, Retrieval & Library for fast storage, indexing, and retrieval of high-dimensional vectors & Vectors / Index, relevance scores & Vector database — index a large set of text embeddings and retrieve the most relevant documents for a query in a retrieval-augmented generation (RAG) system. & open \\

\textbf{Neo4j} \newline \citep{neo4j2012} & Components & Contextual-Structured & Consolidation, Indexing, Updating, Retrieval & Native graph database supporting ACID transactions and Cypher query language & Nodes and relationships with properties / Query results via Cypher & Graph database — model and retrieve complex relational data for tasks like fraud detection and recommendation engines. & conditional open \\

\textbf{BM25} \citep{robertson1995okapi} & Components & Contextual-Unstructured & Retrieval & A probabilistic ranking function for information retrieval to estimate the relevance of documents to a given query. & Text queries / Ranked list of documents & Enhancing search engine results and document retrieval systems. & open \\

\textbf{Contriever} \citep{izacard2021unsupervised} & Components & Contextual-Unstructured & Retrieval & An unsupervised dense retriever trained with contrastive learning, capable of retrieving semantically similar documents across languages. & Query text / List of similar documents & High-recall retrieval tasks in multilingual question-answering systems. & open \\

\textbf{Embedding Models} (e.g., OpenAI embeddings \citep{openai_embeddings_2025}) & Components & Contextual & Consolidation, Retrieval & Techniques to convert text, images, or audio into dense vector representations capturing semantic meaning. & Raw data / Vector embeddings & Text similarity computation, recommendation systems, and clustering tasks. & open \\

\bottomrule
\end{tabular}
}
\caption{\textbf{Component-Level} Tools for Memory Management and Utilization.}
\label{tab:tools_components}
\end{table*}

\begin{table*}[t!]
\centering
\arrayrulecolor{black}
\resizebox{\textwidth}{!}{
\begin{tabular}{
    >{\raggedright\arraybackslash}m{1.4694cm}
    >{\raggedright\arraybackslash}m{1.4694cm}
    >{\raggedright\arraybackslash}m{2cm}
    >{\raggedright\arraybackslash}m{1.8368cm}
    >{\raggedright\arraybackslash}m{4.4082cm}
    >{\raggedright\arraybackslash}m{2.2041cm}
    >{\raggedright\arraybackslash}m{3.6735cm}
    >{\raggedright\arraybackslash}m{1cm}
}
\toprule
 \textbf{Memory Tool} & \textbf{Level} & \textbf{Taxonomy} & \textbf{Operation} & \textbf{Function} & \textbf{Input/Output} & \textbf{Example Use} & \textbf{Source Type} \\
\toprule
\textbf{Graphiti} \citep{he2025graphiti} & framework & Contextual-Structured & Consolidation, Indexing, Updating, Retrieval & Framework for building and querying temporally-aware knowledge graphs tailored for AI agents in dynamic environments.  & Multi-source data / Queryable knowledge graph & Constructing real-time knowledge graphs to enhance AI agent memory. & open \\

\textbf{LLamaIndex} \citep{llamaindex} & framework & Contextual & Consolidation, Indexing, Retrieval  & A flexible framework for building knowledge assistants using LLMs connected to enterprise data. &  Text / Context-augmented responses & Developing knowledge assistants that process complex data formats. & open \\

\textbf{LangChain} \citep{langchain} & framework & Contextual & Consolidation, Indexing, Updating, Forgetting, Retrieval & Provides a framework for building context-aware, reasoning applications by connecting LLMs with external data sources. & Input prompts / Multi-step reasoning outputs & Creating complex LLM applications like question-answering systems and chatbots. & open \\

\textbf{LangGraph} \citep{langgraph2025} & framework & Contextual-Structured & Consolidation, Indexing, Updating, Forgetting, Retrieval & Constructs controllable agent architectures supporting long-term memory and human-in-the-loop multi-agent systems. & Graph state / State updates & Building complex task workflows with multiple AI agents. & open \\

\textbf{EasyEdit} \citep{wang-etal-2024-easyedit} & framework & Parametric & Updating & An easy-to-use knowledge editing framework for LLMs, enabling efficient behavior modification within specific domains.  & Edit instructions / Updated model behavior & Modifying LLM knowledge in specific domains, such as updating factual information. & open \\

\textbf{CrewAI} \citep{duan2024exploration} & framework & Contextual  & Consolidation, Indexing, Retrieval & A platform for building and deploying multi-agent systems, supporting automated workflows using any LLM and cloud platform. & Multi-agent tasks / Collaborative results & Automating workflows across agents like project management and content generation. & open \\

\textbf{Letta} \citep{packer2023memgpt} & framework & Contextual-Unstructured & Consolidation, Retrieval & Constructs stateful agents with long-term memory, advanced reasoning, and custom tools within a visual environment. &  User interactions / Improved Response & Developing AI agents that learn and improve over time. & open \\

\bottomrule
\end{tabular}
}
\caption{\textbf{Framework-Level} Tools for Memory Management and Utilization.}
\label{tab:tools-framework}
\end{table*}

\begin{table*}[t!]
\centering
\arrayrulecolor{black}
\resizebox{\textwidth}{!}{
\begin{tabular}{
    >{\raggedright\arraybackslash}m{1.4694cm}
    >{\raggedright\arraybackslash}m{1.4694cm}
    >{\raggedright\arraybackslash}m{1.4694cm}
    >{\raggedright\arraybackslash}m{1.8368cm}
    >{\raggedright\arraybackslash}m{4.4082cm}
    >{\raggedright\arraybackslash}m{2.2041cm}
    >{\raggedright\arraybackslash}m{3.6735cm}
    >{\raggedright\arraybackslash}m{1cm}
}
\toprule
 \textbf{Memory Tool} & \textbf{Level} & \textbf{Taxonomy} & \textbf{Operation} & \textbf{Function} & \textbf{Input/Output} & \textbf{Example Use} & \textbf{Source Type} \\
\toprule

\textbf{MemOS} \citep{li2025memos} 
& Application Layer 
& Contextual-Structured 
& Consolidation, Updating, Retrieval 
& An operating-system-like architecture that manages hierarchical memory (working, short-term, long-term) to optimize memory-augmented generation. 
& Agent queries, complex contexts / Hierarchical memory blocks 
& Managing complex memory resources for agents handling multi-step reasoning and long-horizon tasks. 
& open \\

\textbf{O-Mem} \citep{wang2025omem} 
& Application Layer 
& Contextual-Unstructured 
& Consolidation, Updating, Retrieval 
& An omni-memory system that enables agents to self-evolve and maintain long-horizon consistency through recursive memory consolidation. 
& Long-term interaction logs / Evolved memory state 
& Creating self-evolving personal AI assistants that adapt to user growth over time. 
& open \\

\textbf{Mem0} \citep{mem0} & Application Layer & Contextual-Unstructured & Consolidation, Indexing, Updating, Retrieval & Provides a smart memory layer for LLMs, enabling direct addition, updating, and searching of memories in models. & User interactions / Personalized responses & Enhancing AI systems with persistent context for customer support and personalized recommendations. & open \\

\textbf{Zep} \citep{rasmussen2025zep} & Application Layer & Contextual-Structured & Consolidation, Indexing, Updating, Retrieval & Integrates chat messages into a knowledge graph, offering accurate and relevant user information. & Chat logs, business data / Knowledge graph query results & Augmenting AI agents with knowledge through continuous learning from user interactions. & open \\

\textbf{Memary} \citep{memary2025} & Application Layer & Contextual & Consolidation, Indexing, Updating, Retrieval & An open memory layer that emulates human memory to help AI agents manage and utilize information effectively. & Agent tasks / Memory management and utilization & Building AI agents with human-like memory characteristics. & open \\

\textbf{Memobase} \citep{memobase2025} & Application Layer & Contextual & Consolidation, Indexing, Updating, Retrieval & A user profile-based long-term memory system designed to provide personalized experiences in generative AI applications. & User interactions / Personalized responses & Implementing virtual assistants, educational tools, and personalized AI companions. & open \\

\bottomrule
\end{tabular}
}
\caption{\textbf{Application Layer-Level} Tools for Memory Management and Utilization.}
\label{tab:tools-application_layer}
\end{table*}

\begin{table*}[t!]
\centering
\arrayrulecolor{black}
\resizebox{\textwidth}{!}{
\begin{tabular}{
    >{\raggedright\arraybackslash}m{1.8cm}
    >{\raggedright\arraybackslash}m{1.4694cm}
    >{\raggedright\arraybackslash}m{1.4694cm}
    >{\raggedright\arraybackslash}m{1.8368cm}
    >{\raggedright\arraybackslash}m{4.4082cm}
    >{\raggedright\arraybackslash}m{2.2041cm}
    >{\raggedright\arraybackslash}m{3.6735cm}
    >{\raggedright\arraybackslash}m{1cm}
}
\toprule
 \textbf{Memory Tool} & \textbf{Level} & \textbf{Taxonomy} & \textbf{Operation} & \textbf{Function} & \textbf{Input/Output} & \textbf{Example Use} & \textbf{Source Type} \\
\toprule

\textbf{Me.bot} \citep{mebot2025} & Product & Contextual & Consolidation, Indexing, Updating, Retrieval & AI-powered personal assistant that organizes notes, tasks, and memories, providing emotional support and productivity tools. & User inputs (text, voice) / Organized notes, reminders, summaries & Personal productivity enhancement, emotional support, idea organization. & closed \\

\textbf{ima.copilot} \citep{imacopilot2025} & Product & Contextual & Consolidation, Indexing, Updating, Retrieval & Intelligent workstation powered by Tencent's Mix Huang model, building a personal knowledge base for learning and work scenarios. & User queries / Customized responses, knowledge retrieval & Enhancing learning efficiency, work productivity, knowledge management. & closed \\

\textbf{Coze} \citep{coze2024} & Product & Contextual & Consolidation & Enabling multi-agent collaboration across various platforms. & User-defined workflows / Response & Deployed chatbots, AI agents & closed \\

\textbf{Grok} \citep{grok2023} & Product & Contextual & Retrieval, Compression & AI assistant developed by xAI, designed to provide truthful, useful, and curious responses, with real-time data access and image generation. & Query / Informative answers, generated images & Answering questions, generating images, providing insights. & closed \\

\textbf{ChatGPT} \citep{openai2023chatgpt} & Product & Contextual & Consolidation, Retrieval & Conversational AI developed by OpenAI, capable of understanding and generating human-like text based on prompts. & User prompts / Generated text responses & Answering questions, generating images, providing insights. & closed \\

\textbf{Replika} \citep{replika2025} & Product & Contextual & Consolidation, Updating, Retrieval & AI companion maintaining longitudinal interaction history for emotional continuity. & Text input / Emotionally responsive dialogues & Affective support, mental wellness, simulated companionship. & closed \\

\textbf{Amazon Recommender} \citep{linden2003amazon} & Product & Contextual & Consolidation, Retrieval, Indexing & Personalized recommendation engine using behavioral memory traces. & User behavior logs / Ranked product recommendations & E-commerce personalization, customer profiling, targeted marketing. & closed \\

\textbf{GitHub Copilot} \citep{githubcopilot} & Product & Contextual & Retrieval, Compression & Code assistant that provides suggestions based on coding history and file context. & Code editor context / Code completions, snippets & Programming aid, autocomplete, contextual understanding. & closed \\

\textbf{CodeBuddy} \citep{codebuddy2025} & Product & Contextual & Retrieval, Compression & AI code assistant. & Code and edits / Personalized coding suggestions & Habit-aware code generation, interactive development support. & closed \\

\textbf{Doubao} \citep{doubao2025} 
& Product 
& Contextual 
& Consolidation, Retrieval 
& High-efficiency multimodal AI assistant capable of handling long-context interactions and diverse everyday tasks. 
& Text, voice, image inputs / Answers, creative content 
& Daily conversation, writing assistance, coding support, and role-playing. 
& closed \\

\textbf{Siri} \citep{siri2025} 
& Product 
& Contextual 
& Consolidation, Retrieval 
& Intelligent voice assistant utilizing on-device personal semantic memory for context-aware and cross-application actions. 
& Voice commands / Action execution, personal information retrieval 
& Device control, retrieving personal context (e.g., flight schedules), and cross-app tasks. 
& closed \\

\textbf{Xiaoyi} \citep{xiaoyi2025} 
& Product 
& Contextual 
& Consolidation, Retrieval 
& Smart assistant integrated into HarmonyOS, leveraging ecosystem-level memory for proactive services and document understanding. 
& Voice, text, documents / Summaries, suggestions, IoT control 
& Document summarization, smart home control, and personalized travel planning. 
& closed \\

\bottomrule
\end{tabular}
}
\caption{\textbf{Product-Level} Tools for Memory Utilization.}
\label{tab:tools_products}
\end{table*}









\end{document}